%% file: arxiv_main.tex
\documentclass[10pt, letterpaper]{article}
\usepackage{arxiv}

\input{math_commands.tex}

\usepackage{url}
\usepackage{bm}
\usepackage{amsmath}
\usepackage{amsfonts}
\usepackage{amssymb}
\usepackage{mathtools}
\usepackage{algorithm}
\usepackage{algorithmic}
\usepackage{amsthm}
\usepackage{comment}
\usepackage{caption}
\usepackage{subcaption}
\usepackage{booktabs}

\title{Functional Critics for Off-Policy Actor-Critic:\\ 
From Stability to Exploration}

\author{
  Qinxun Bai$^{*\;\dagger\;1}$
  \quad
  Yuxuan Han$^{*\;2}$
  \quad
  Yuxiao Wen$^{*\;2}$
  \quad
  Wei Xu$^{3}$
  \quad
  Zhengyuan Zhou$^{2}$
}

\date{}

\begin{document}

\maketitle

\renewcommand{\thefootnote}{\fnsymbol{footnote}}

\footnotetext[1]{Equal contribution.}

\footnotetext[2]{Correspondence. }

\renewcommand{\thefootnote}{\arabic{footnote}} 

\footnotetext[1]{Isara Laboratories. Work partially performed while at Horizon Robotics. \texttt{jerry@isara.io}}
\footnotetext[2]{Stern School of Business, New York University. \texttt{\{yh6061, yuxiaowen, zzhou\}@stern.nyu.edu}}
\footnotetext[3]{Horizon Robotics. \texttt{wei.xu@horizon.auto}}

\begin{abstract}
Actor-critic methods are central to modern reinforcement learning, yet off-policy actor-critic theory remains far removed from practical algorithms, due in part to the interaction between the deadly triad and the continually changing target policy. We revisit functional critics, i.e., value functions that explicitly condition on the policy, and 
show that this representation provides a shared foundation for off-policy
training stability and policy-space exploration.
We first develop a convergent off-policy actor-critic framework under linear
function approximation that accommodates target-based TD, partial coverage,
and evolving behavior policies.
Our analysis combines a policy-evaluation guarantee under partial coverage with policy-gradient generalization of functional critics to yield a dual trust-coverage framework that controls policy-gradient error under evolving behavior policies while enabling greater reuse of off-policy data. 
We further develop a replay-buffer analysis under i.i.d. sampling,
yielding a principled replay-sampling design for minimizing
the replay-sample complexity required to reach a prescribed evaluation
accuracy.
We next show that functional critics provide a natural representation for model-free posterior-sampling-style exploration. 
Standard critic ensembles do not naturally represent joint uncertainty over
return functions across policy space, whereas functional-critic ensembles
represent policy-conditioned return functions directly.
This motivates a particle-based, model-free analogue of
posterior-sampling-style exploration without explicit model learning.
Finally, we instantiate these ideas in a deep functional actor-critic
algorithm that, despite omitting several standard off-policy AC heuristics,
achieves strong performance across the full Dog and Humanoid task series of
the DeepMind Control Suite.
\end{abstract}

\section{Introduction}
\label{sec:intro}

Off-policy actor-critic (AC) methods combine policy optimization with the ability to reuse experience collected by behavior policies, making them particularly attractive for continuous and high-dimensional control
\citep{degris2012off,silver2014deterministic,lillicrap2015continuous,haarnoja2018soft}.
Despite their empirical success, however, the theory of off-policy AC remains far removed from practical algorithms.
Two central challenges are \textbf{training stability} and \textbf{exploration}: the former concerns how to reliably improve the policy from off-policy experience, while the latter concerns how to collect informative experience in the first place.

A \textbf{functional critic}, or policy-conditioned value function
\citep{harb2020policy,tang2022inputting,faccio2020parameter},
models a value functional $Q(s,a,\pi)$ by explicitly taking the policy as input.
Functional critics were originally motivated by the ``moving target''
problem in AC: as the actor changes, a conventional critic must repeatedly
relearn a new value function.
Yet their broader theoretical role remains unclear, particularly for
exploration.
In this work, we show that functional critics provide a shared foundation
for off-policy training stability and exploration.

\paragraph{Training stability.}
Off-policy AC combines two sources of difficulty.
Even for a fixed target policy, temporal-difference learning with function approximation and off-policy data can diverge due to the ``deadly triad''
\citep{baird1995residual,tsitsiklis1996analysis,sutton2018reinforcement}.
At the same time, actor updates continuously change the policy being evaluated.
Existing convergence theory therefore relies on algorithms and assumptions that differ substantially from common practice: gradient or emphatic TD
\citep{sutton2009fast,sutton2016emphatic,zhang2020provably}
rather than target-based TD; fixed behavior policies with strong coverage assumptions rather than behavior policies that track the actor; and samples from a stationary behavior distribution rather than a replay buffer.

We formalize functional critics under linear function approximation and
establish a convergent off-policy AC framework that accommodates three
practical ingredients:
\textbf{target-based TD, partial coverage, and evolving behavior policies}.
We separately develop an i.i.d.\ replay-buffer analysis that connects replay
sampling to policy-evaluation accuracy and replay-sample complexity.
Several technical innovations are central to the analysis.
First, we establish a policy-evaluation guarantee under partial coverage, requiring only coverage of feature directions relevant to the target policy rather than full state-action coverage.
Second, functional critics provide a policy-gradient generalization property that controls gradient error as the actor moves away from the reference policy at which the critic was evaluated.
Together, they yield a dual trust-coverage framework that controls evaluation error arising from off-policy experience and gradient-generalization error arising from target-policy evolution, establishing convergence in policy-gradient norm while permitting greater reuse of off-policy data.
Third, under i.i.d.\ replay sampling, we decompose the policy-evaluation
error into finite-buffer statistical and TD-optimization components,
exposing a trade-off between \emph{sample diversity} and
\emph{feature coverage}.
This yields a principled replay-sampling and evaluation-budget design for
minimizing the replay-sample complexity required to reach a prescribed
evaluation accuracy.

\paragraph{Exploration through the functional lens.}
Functional critics also reveal a connection to efficient exploration.
Posterior Sampling Reinforcement Learning (PSRL) provides a principled exploration framework, but explicitly maintaining a posterior over environment models and repeatedly solving sampled MDPs is computationally prohibitive in large-scale problems.
Existing model-free approximations instead represent uncertainty through ensembles or randomized priors
\citep{osband2016deep,osband2018randomized,osband2019deep}.
We illustrate that ordinary critic ensembles do not naturally represent
joint uncertainty over return functions across policy space.
Functional-critic ensembles instead represent policy-conditioned return
functions directly, motivating a particle-based, model-free analogue of
posterior-sampling-style exploration.

\paragraph{From theory to practice.}
We instantiate these ideas in an AC algorithm using a neural functional-critic architecture.
Our implementation deliberately omits several standard off-policy AC
heuristics, including minimum twin-target values, exploration-noise schedules,
and entropy regularization.
It achieves strong performance across the full Dog and Humanoid task series
of the DeepMind Control Suite \citep{tassa2018deepmind}.


\paragraph{Contributions.}
Our contributions are threefold:
\begin{enumerate}[leftmargin=1.5em, labelsep=0.5em, itemsep=0.5ex, label=\textbf{\arabic*.}]
   
    \item \textbf{Bridging off-policy AC theory and practice.}
    We establish an off-policy AC convergence guarantee under linear function
    approximation that jointly accommodates target-based TD, partial coverage,
    and evolving behavior policies.
    Our analysis combines a partial-coverage policy-evaluation guarantee with
    policy-gradient generalization of functional critics to obtain the dual
    trust-coverage framework.
    We additionally develop an i.i.d.\ replay-buffer analysis that yields a
    principled replay-sampling and evaluation-budget design.    
    
    \item \textbf{Policy-space exploration through functional critics.}
    We reveal a structural connection between functional critics and
    posterior-sampling-style exploration.
    Standard critic ensembles do not naturally represent joint uncertainty over
    return functions across policy space, whereas functional-critic ensembles
    represent policy-conditioned return functions directly.
    This motivates a particle-based, model-free analogue of posterior-sampling
    exploration without explicit model learning.

    \item \textbf{A neural functional actor-critic implementation.}
    We translate the functional-critic representation into a practical
    actor-critic algorithm with a tailored neural architecture.
    The implementation deliberately omits several heuristics commonly used by
    modern off-policy AC methods, yet achieves strong performance across the full
    Dog and Humanoid task series of the DeepMind Control Suite using a single
    hyperparameter configuration.
    These results demonstrate the practical viability of neural functional
    critics beyond the linear theoretical setting.

\end{enumerate}

\section{Preliminaries}
\label{sec-prob_def}

\paragraph{MDP and Off-Policy Actor-Critic.}
We consider an infinite-horizon discounted Markov decision process (MDP)
$(\mathcal{S},\mathcal{A},p,r,\gamma)$ with finite state and action spaces,
transition kernel $p$, reward $r:\mathcal{S}\times\mathcal{A}\rightarrow[0,1]$,
and discount $\gamma\in[0,1)$.
For a policy $\pi$, its value and action-value functions are
\begin{align*}
V_\pi(s)
&:=\E_\pi\left[\sum_{t=0}^{\infty}\gamma^t r(s_t,a_t)\mid s_0=s\right],\\
Q_\pi(s,a)
&:=r(s,a)+\gamma\E_{s'\sim p(\cdot|s,a)}[V_\pi(s')].
\end{align*}
We parameterize the policy by $\theta\in\Theta\subseteq\mathbb{R}^d$ and
write $\pi_\theta$, $V_\theta$, and $Q_\theta$ accordingly.
Given an initial-state distribution $\rho$, the objective is
$J_\rho(\theta):=\E_{s\sim\rho}[V_\theta(s)]$.

In off-policy RL, experience from a \emph{behavior policy} $\mu$ is reused
to optimize $\pi_\theta$.
Actor-critic (AC) alternates critic evaluation and actor updates, while
off-policy critic learning with function approximation can suffer from
deadly-triad instability
\citep{sutton2009fast,sutton2016emphatic,zhang2021breaking}.


\paragraph{Off-Policy Gradients with Functional Critics.}
Because the target-policy state distribution cannot be sampled directly, the
standard policy gradient cannot be directly estimated from behavior data.
Existing off-policy formulations instead optimize a surrogate
$J_{d^\mu}$ with a policy-dependent emphasis correction
\citep{imani2018off}.
We instead differentiate $J_\rho(\theta)$ directly:
\begin{equation}
\label{eq-policy-grad-full}
\nabla_\theta J_\rho(\theta)
=
\sum_s\rho(s)\sum_a
\left[
\nabla_\theta\pi_\theta(a|s)Q_\theta(s,a)
+
\pi_\theta(a|s)\nabla_\theta^{\mathrm{pol}} Q_\theta(s,a)
\right].
\end{equation}
The second term in \eqref{eq-policy-grad-full} is difficult to evaluate with a conventional critic because $Q_\theta$ depends implicitly on the policy parameters.
A functional critic instead models this dependence explicitly as
$\hat Q(\pi_\theta,s,a;\xi)$.
Following \citet{faccio2020parameter}, this yields the direct gradient estimate
\begin{equation}
\label{eq-our-gradient-formula}
\nabla_\theta J_\rho(\pi_\theta;\xi)
=
\E_{s\sim\rho}\left[
\sum_a
\left(
\hat Q(\pi_\theta,s,a;\xi)\nabla_\theta\pi_\theta(a|s)
+
\pi_\theta(a|s)\nabla_\theta^{\mathrm{pol}}\hat Q(\pi_\theta,s,a;\xi)
\right)
\right].
\end{equation}
Here, $\nabla_\theta^{\mathrm{pol}}$ denotes differentiation with respect
to the policy parameters through the policy dependence of the value function,
while holding the state--action arguments fixed. For a functional critic,
this corresponds to differentiation through its explicit policy input.


\paragraph{Posterior Sampling for Reinforcement Learning (PSRL).}
PSRL extends Thompson sampling to MDPs.
It maintains a posterior over the environment, samples an MDP at the
beginning of each episode, computes its optimal policy, executes that policy,
and updates the posterior with the collected experience.
Although PSRL provides a principled exploration mechanism \citep{osband2013more}, maintaining a
posterior over transition and reward models and repeatedly
solving sampled MDPs is generally computationally prohibitive in high-dimensional continuous spaces.

\section{Methodology}
\label{sec-method}

We now develop the functional actor-critic framework and its theoretical foundations.
Section~\ref{subsec: meta-algorithm} introduces the general meta-algorithm.
Section~\ref{sec-linear-functional} establishes convergence under linear
function approximation, including target-based evaluation with partial
coverage and the dual trust-coverage mechanism for evolving behavior policies.
Section~\ref{sec:replay-budget} separately studies replay-buffer sampling and
critic-budget allocation under an i.i.d.\ sampling model.


\subsection{Functional Actor-Critic Meta-Algorithm}
\label{subsec: meta-algorithm}

Our framework maintains an actor $\pi_\theta$ and a functional critic
$\hat Q(\pi_\theta,s,a;\xi)$ that explicitly takes the policy as input.
At each iteration, off-policy experience is used to update the critic toward
the Bellman equation of the current actor, while the actor is updated using
the direct functional gradient in \eqref{eq-our-gradient-formula}.
Because the policy is an explicit input, a single critic can generalize across
evolving actors rather than relearning an independent value function after
each policy update.

Algorithm~\ref{alg-meta-algorithm} summarizes this template; subsequent
sections instantiate the functional policy evaluator and its update schedule.

\begin{algorithm}[t]
\caption{Functional Actor-Critic Meta-Algorithm}
\label{alg-meta-algorithm}
\begin{algorithmic}[1]
\STATE \textbf{Input:} $\theta_0,\xi_0,T,\{\eta_t\}_{t=1}^T$
\FOR{$t=1,\ldots,T$}
    \STATE Collect/sample off-policy data $\mathcal D_t$
    \STATE $\xi_t \gets
    \textsc{FunctionalPolicyEvaluator}
    (\xi_{t-1},\pi_{\theta_{t-1}},\mathcal D_t)$
    \STATE $\theta_t \gets \theta_{t-1}
    + \eta_t
    \nabla_\theta J_{\rho}(\pi_\theta;\xi_t)
    |_{\theta=\theta_{t-1}}$
\ENDFOR
\STATE \textbf{Return} $\theta_T$
\end{algorithmic}
\end{algorithm}

\subsection{Convergence Analysis under Linear Function Approximation}
\label{sec-linear-functional}

We now instantiate the functional actor-critic framework under linear function approximation and establish convergence for target-based off-policy AC with partial coverage and evolving behavior policies.
We present the main results and defer complete proofs to Appendix~\ref{appendix:sec-linear-functional}.
\subsubsection{Linear Function Approximation Setup}
We analyze our framework under the following structural and regularity
assumptions.

\begin{assumption}[Strong Mixing]\label{assumption-mixing}
For every $\pi\in\Pi$, the induced Markov chain is irreducible and aperiodic,
with unique stationary distribution $d^\pi$.
\end{assumption}

\begin{assumption}[Linear Functional Representation]
\label{assumption-linear}
For the policy class $\Pi$, there exist a feature map
$\phi:\Pi\times\mathcal S\times\mathcal A\to\mathbb B^d$
and $\xi^\star\in\mathbb R^d$, $\|\xi^\star\|_2\le L$, such that
\[
Q^\pi(s,a)=\langle\phi^\pi(s,a),\xi^\star\rangle,
\qquad
\forall(\pi,s,a)\in\Pi\times\mathcal S\times\mathcal A .
\]
\end{assumption}

\begin{assumption}[Bellman Completeness]\label{assumption-closeness}
For any $\xi\in\RR^d$, there exists $\xi'\in\RR^d$ such that
\[
\langle\phi^\pi(s,a),\xi'\rangle
=
\mathcal T^\pi
\left(\left\lceil\langle\phi^\pi(\cdot,\cdot),\xi\rangle\right\rceil\right)(s,a)
+\mathcal E_{\rm approx},
\quad
\forall(\pi,s,a)\in\Pi\times\mathcal S\times\mathcal A,
\]
where $\lceil\cdot\rceil$ projects onto $[0,(1-\gamma)^{-1}]$, 
and $\cE_{\text{approx}}$ is the inherent approximation error.
\end{assumption}

Assumption~\ref{assumption-mixing} is a standard ergodicity condition for
off-policy stochastic approximation
\citep{zhang2020provably,zhang2021breaking,chen2023target}.
Assumptions~\ref{assumption-linear}--\ref{assumption-closeness} impose
functional realizability and approximate Bellman completeness across policies;
further discussion appears in Appendix~\ref{appendix-assumption-linear}.


\subsubsection{Target-Based Evaluation with Partial Coverage}\label{sec:trust_coverage}

We evaluate a target policy $\pi$ from data collected by a behavior policy
$\mu$ using target-based TD, with the complete procedure given in
Algorithm~\ref{alg:target-based-evaluation}.
Unlike existing guarantees that typically require full coverage by $\mu$, we require
coverage only of feature directions relevant to the target policy.



Define the on-policy and off-policy feature covariance as follows,
\begin{equation}\label{eq: A-def}
\mathbf A_\pi
:=
\E_{d^\pi}\left[
\phi^\pi(s,a)\phi^\pi(s,a)^\top
\right],
\qquad
\widetilde{\mathbf A}_\pi
:=
\E_{d^\mu}\left[
\phi^\pi(s,a)\phi^\pi(s,a)^\top
\right].
\end{equation}

\begin{definition}[Evaluation Trust Metric]
\label{def:eval_metric}
For target policy $\pi$ and behavior policy $\mu$, define
\begin{equation}
\label{eq:eval_metric}
\mathcal C(\pi,\mu)
:=
\E_{d^\pi}
\left[
\|\phi^\pi(s,a)\|_{\widetilde{\mathbf A}_\pi^{-1}}^2
\right],
\end{equation}
where
$\|\mathbf u\|_{\mathbf A^{-1}}
:=
\lim_{\lambda\to0_+}
\|\mathbf u\|_{(\mathbf A+\lambda\mathbf I)^{-1}}$.
\end{definition}

Finite $\mathcal C(\pi,\mu)$ requires coverage only of feature directions
relevant under $\pi$, rather than every state--action pair; The key technical step is to establish Bellman contraction in the
$\mathbf A_\pi$-weighted $L^2$ norm.


\begin{theorem}[Evaluation Guarantee with Partial Coverage]
\label{thm-evaluation}
Let Assumptions~\ref{assumption-mixing}--\ref{assumption-closeness} hold.
With suitable $\alpha$ and $K\asymp\epsilon^{-2}$, the target-based evaluator
satisfies
\begin{equation}
\label{eq:Q-hat-convergence}
\E\left[
\|\hat Q_T-Q^\pi\|_{d^\pi}^2
\right]
\lesssim
\gamma^T
+
\mathcal C(\pi,\mu)
\left(\epsilon^2+\mathcal E_{\rm approx}^2\right)
+
\mathcal E_{\rm approx}^2 .
\end{equation}
\end{theorem}
To our knowledge, Theorem~\ref{thm-evaluation} is the first target-based
TD policy-evaluation guarantee that replaces pointwise full coverage with a
feature-based coverage condition, allowing convergence whenever
$\mathcal C(\pi,\mu)<\infty$.
The corresponding parameter-space guarantee is
\begin{equation}
\label{eq:weight-convergence}
\E\left[
\|\hat\xi_T-\xi^\star\|_{\mathbf A_\pi}^2
\right]
\lesssim
\gamma^T
+
\mathcal C(\pi,\mu)
\left(\epsilon^2+\mathcal E_{\rm approx}^2\right)
+
\mathcal E_{\rm approx}^2,
\end{equation}
which will serve as the evaluation guarantee for the full actor-critic
analysis.

\subsubsection{Off-Policy AC with Dual Trust-Coverage: Algorithm and Convergence}


We now extend the partial-coverage evaluation guarantee to the full
actor-critic setting.
At epoch $k$, a reference policy $\bar{\pi}^{(k)}$ anchors the functional
critic, with
$\mathbf A_k:=\mathbf A_{\bar{\pi}^{(k)}}$.
Two sources of error must be controlled:
(i) the error in evaluating $\bar{\pi}^{(k)}$ from off-policy experience,
and (ii) the gradient-generalization error as the current actor
$\pi_t^{(k)}$ evolves away from $\bar{\pi}^{(k)}$.

The first is controlled by the evaluation trust metric
$\mathcal C(\bar{\pi}^{(k)},\mu)$.
Whenever $\mathcal C(\bar{\pi}^{(k)},\mu)>C_{\max}$, the behavior policy is
refreshed as $\mu\gets\bar{\pi}^{(k)}$.
After evaluating the reference policy to obtain $\xi^{(k)}$, gradient
compatibility controls the actor update at the reference policy, while a
second, \emph{gradient-generalization trust metric} $\Delta_{k,t}$ governs
subsequent reuse of the critic as the actor evolves away from the reference.
Algorithm~\ref{alg-linear-AC} summarizes this dual trust-coverage mechanism.





\begin{algorithm}[t]
\caption{Functional Actor--Critic with Dual Trust-Coverage}
\label{alg-linear-AC}
\begin{algorithmic}[1]
\STATE \textbf{Input:} $\phi,\epsilon,\eta,C_{\max}$
\STATE \textbf{Initialize:}
$\bar{\pi}^{(0)}$, $\mu\gets\bar{\pi}^{(0)}$
\FOR{$k=0,1,\ldots$}
    \IF{$\mathcal C(\bar{\pi}^{(k)},\mu)>C_{\max}$}
        \STATE $\mu\gets\bar{\pi}^{(k)}$
    \ENDIF
    \STATE
    $\xi^{(k)}
    \gets
    \textsc{FunctionalPolicyEvaluator}
    (\mu,\bar{\pi}^{(k)},\phi,\epsilon)$
    \STATE $\pi_0^{(k)}\gets\bar{\pi}^{(k)}$
    \FOR{$t=1,2,\ldots$}
        \STATE
        $\pi_t^{(k)}
        \gets
        \pi_{t-1}^{(k)}
        +\eta\hat\nabla J(\pi_{t-1}^{(k)},\xi^{(k)})$
        \IF{$\Delta_{k,t}>2$}
            \STATE $\bar{\pi}^{(k+1)}\gets\pi_t^{(k)}$; \textbf{break}
        \ENDIF
    \ENDFOR
\ENDFOR
\end{algorithmic}
\end{algorithm}

By Theorem~\ref{thm-evaluation}, whenever
$\mathcal C(\bar{\pi}^{(k)},\mu)\leq C_{\max}$,
the anchored critic satisfies
\begin{equation}
\label{eq:xi-k-error}
\sqrt{
\E\left[
\|\xi^{(k)}-\xi^\star\|_{\mathbf A_k}^2
\right]}
\lesssim
\bar\epsilon
:=
\sqrt{C_{\max}}
\left(\epsilon+\mathcal E_{\rm approx}\right).
\end{equation}

To transfer~\eqref{eq:xi-k-error} to the gradient evaluation error, we introduce the functional-gradient operator,
\begin{equation}
\label{eq:gradient-error-operator}
\mathbf G_{\pi_\theta}
:=\mathbb E_{s\sim\rho}\!\left[
\sum_a\left(
\nabla_\theta\pi_\theta(a\mid s)\phi^{\pi_\theta}(s,a)^\top
+\pi_\theta(a\mid s)D\phi^{\pi_\theta}(s,a)
\right)\right],
\end{equation}
under which the gradient error can be represented as
$\hat\nabla J_\rho(\pi;\xi)-\nabla J_\rho(\pi)
=\mathbf G_\pi(\xi-\xi^\star)$. 
We impose the following compatibility condition,

\begin{assumption}[Gradient Compatibility]
\label{assumption:anchor-gradient}
There exists a finite constant $\kappa_{\mathrm g}>0$ such that for every
reference policy $\pi$,
\begin{equation}
\label{eq:anchor-gradient-compatibility}
\mathbf G_\pi^\top\mathbf G_\pi
\preceq\kappa_{\mathrm g}^2\mathbf A_\pi.
\end{equation}
\end{assumption}
Assumption~\ref{assumption:anchor-gradient} ensures that critic error
controlled in the $\mathbf A_\pi$ seminorm also controls the corresponding
policy-gradient error, up to a factor of $\kappa_{\rm g}$.
Together with~\eqref{eq:xi-k-error}, it therefore yields gradient error
$\mathcal O(\kappa_{\rm g}\bar\epsilon)$ at the reference policy.
Away from the reference policy, critic estimation error can be amplified
through both the functional features $\phi$ and their policy derivatives
$D\phi$.
We quantify this amplification by the
\emph{gradient-generalization trust metric}
\begin{equation}
\label{eq:delta}
\Delta_{k,t}
:=
\max\{\mathcal C_1^{(k,t)},\mathcal C_2^{(k,t)}\},
\end{equation}
where $\mathcal C_1^{(k,t)}$ and $\mathcal C_2^{(k,t)}$ measure,
respectively, amplification through $\phi$ and $D\phi$; their explicit definitions are given in
Appendix~\ref{appendix:dual-trust-coverage}.
The resulting gradient error satisfies
\begin{equation}
\label{eq:gradient-generalization-bound}
\E\left[
\|\nabla J_\rho(\pi_t^{(k)})
-\hat\nabla J_\rho(\pi_t^{(k)},\xi^{(k)})\|_\infty
\right]
\lesssim
\Delta_{k,t}\bar\epsilon .
\end{equation}
The evaluation criterion $\mathcal C$ controls reference-policy evaluation. The first actor update is controlled by Assumption~\ref{assumption:anchor-gradient}, while
$\Delta_{k,t}$ controls subsequent reuse of the anchored critic as the actor evolves.



Under the standard smoothness assumption that $\phi^\pi$ and $D\phi^\pi$
are Lipschitz in the policy parameters, we obtain the following convergence
guarantee.

\begin{theorem}[Convergence of Functional Actor-Critic]
\label{thm-main-convergence}
Let Assumptions~\ref{assumption-mixing}--\ref{assumption-closeness},
Assumption~\ref{assumption:anchor-gradient}, and the smoothness condition
above hold. Let $c_{\mathrm g}:=\max\{\kappa_{\mathrm g}^2,16d\}$.
For a suitable stepsize $\eta$, the policies generated by
Algorithm~\ref{alg-linear-AC} over $M$ actor updates satisfy 
\begin{equation}
\label{eq:global-convergence}
\frac{1}{M}
\E\left[
\sum_{k}\sum_{t=0}^{T_k-1}
\|\nabla J_\rho(\pi_t^{(k)})\|^2
\right]
\leq
\frac{2(J_\rho^\star-J_\rho(\pi^{(0)}))}{\eta M}
+
\tilde{\mathcal{O}}(c_{\mathrm g}\bar\epsilon^2).
\end{equation}
\end{theorem}
The dual trust-coverage mechanism therefore exposes two complementary reuse
horizons: the first criterion governs how long behavior data can be reused,
while the second governs how many actor updates can reuse the same functional
critic.

\subsection{Replay Sampling and Critic Budgets}
\label{sec:replay-budget}

The analysis above, following recent off-policy AC theory, already allows
Markovian samples generated by a behavior policy.
Yet it still shares a major abstraction with existing theory: policy
evaluation draws samples directly from the behavior process, whereas replay-buffer
training is a standard component of practical off-policy AC.
Bridging this gap is more than an implementation detail.
While replay sampling is often handled heuristically in practice, a
principled treatment should connect the sampling design directly to
convergence and the replay-sample complexity required to reach a prescribed
evaluation accuracy. In this section, we propose a optimization problem for buffer sampling distribution design. For simplicity, throughout this section, we assume that samples in the buffer are independent, $\mathcal{E}_{\mathrm{approx}}=0$, and $\bA_{\pi}\succ 0$.

There are two competing considerations.
On the one hand, broadly reusing both old and new experience increases
\emph{sample diversity} and reduces finite-data statistical error.
On the other hand, when the behavior policy tracks the target policy, newer
experience tends to provide better \emph{feature coverage} of the current
target and can therefore accelerate TD evaluation.
We show that these two effects appear as distinct terms in the policy
evaluation error, yielding a principled diversity--coverage trade-off.

\paragraph{Replay distributions and evaluation error.}
Fix a target policy $\pi$ and a replay buffer
$\mathcal B=\{(s_i,a_i,r_i,s_i')\}_{i=1}^{n}$.
A replay distribution is a vector
$\mathbf w=(w_1,\ldots,w_n)\in\Delta^n$ specifying the probability of
sampling each buffer entry.
Define
\begin{equation}
\label{eq:replay-design}
\boldsymbol\Sigma_{\mathbf w}
:=
\sum\nolimits_{i=1}^{n}
w_i\phi^\pi(s_i,a_i)\phi^\pi(s_i,a_i)^\top,
\qquad
\mathbf M_{\mathbf w}
:=
\sum\nolimits_{i=1}^{n}
w_i^2\phi^\pi(s_i,a_i)\phi^\pi(s_i,a_i)^\top ,
\end{equation}
and the two replay-design criteria
\begin{align}
\mathcal V_{\mathcal B}(\pi,\mathbf w)
&:=
\mathrm{tr}\big(
\mathbf A_\pi
\boldsymbol\Sigma_{\mathbf w}^{-1}
\mathbf M_{\mathbf w}
\boldsymbol\Sigma_{\mathbf w}^{-1}
\big),
\quad \mathcal C_{\mathcal B}(\pi,\mathbf w)
:=
\mathrm{tr}\big(
\mathbf A_\pi
\boldsymbol\Sigma_{\mathbf w}^{-1}
\big).
\label{eq:replay-coverage}
\end{align}
Here $\mathcal V_{\mathcal B}$ captures the finite-buffer statistical cost of
the sampling design, whereas $\mathcal C_{\mathcal B}$ measures how well the
replayed features cover directions relevant to the target policy.
The squared weights in $\mathbf M_{\mathbf w}$ arise because the statistical
variance of a weighted observation scales with $w_i^2$.

Recall that Algorithm~\ref{alg:target-based-evaluation} performs $K$
inner TD updates for each of $T$ target-network iterations.
Under replay sampling, each inner TD update draws one transition from
$\mathcal B$ according to $\mathbf w$; hence a complete policy-evaluation
call uses $TK$ replay samples in total.

We can then establish the following replay-based policy-evaluation guarantee.

\begin{lemma}[Replay evaluation guarantee]
\label{prop:replay-evaluation}
Suppose Assumptions~\ref{assumption-mixing}--\ref{assumption-closeness}
hold.
There exist constants $\alpha>0$, $c_{\mathrm{stat}}>0$, and
$c_{\mathrm{TD}}>0$ such that, for any integers $K,T\geq1$, the output of
Algorithm~\ref{alg:target-based-evaluation} using replay distribution
$\mathbf w$ and step size $\alpha$ satisfies, up to logarithm factors,
\begin{equation}
\label{eq:replay-evaluation-guarantee}
\mathbb E\left[
 \|\hat\xi_T-\xi^\star\|_{\mathbf A_\pi}^2
 \right]
\leq
\frac{c_{\mathrm{stat}}
\mathcal V_{\mathcal B}(\pi,\mathbf w)}{(1-\gamma)^3}
+
\frac{c_{\mathrm{TD}}
\mathcal C_{\mathcal B}(\pi,\mathbf w)}{K(1-\gamma)^4}
+
\frac{3\gamma^{2T}}{(1-\gamma)^2}.
\end{equation}
\end{lemma}

\paragraph{Statistical and optimization errors.}
For a fixed replay distribution $\mathbf w$, let
$\xi^{\mathrm{LS}}(\mathbf w)$ denote the corresponding weighted least-squared
solution.
The policy-evaluation error admits the decomposition
\begin{equation}
\label{eq:replay-error-decomposition}
\mathbb E\big[
\|\hat\xi_T-\xi^\star\|_{\mathbf A_\pi}^2
\big]
\leq
2\mathbb E\big[
\|\xi^{\mathrm{LS}}(\mathbf w)-\xi^\star\|_{\mathbf A_\pi}^2
\big]
+
2\mathbb E\big[
\|\hat\xi_T-\xi^{\mathrm{LS}}(\mathbf w)\|_{\mathbf A_\pi}^2
\big].
\end{equation}
The first term is the finite-buffer statistical error under $\mathbf{w}$ and
cannot be reduced by additional TD updates; the second is the TD-to-LS
optimization gap.
Lemma~\ref{prop:replay-evaluation} controls these terms through
$\mathcal V_{\mathcal B}(\pi,\mathbf w)$ and
$\mathcal C_{\mathcal B}(\pi,\mathbf w)/K$, respectively, with
$\gamma^{2T}$ accounting for the residual target-network error.
Thus, $\mathcal V_{\mathcal B}$ sets the accuracy floor, while
$\mathcal C_{\mathcal B}/K$ governs the cost of approaching it.

\paragraph{Diversity--coverage trade-off.}
These two error components favor different replay distributions.
Let $\mathbf u=(1/n,\ldots,1/n)$ denote uniform replay.
The statistical criterion satisfies $\mathcal V_{\mathcal B}(\pi,\mathbf w)
\geq
\mathcal V_{\mathcal B}(\pi,\mathbf u),$
so uniform replay minimizes the statistical error by making the broadest use
of the finite buffer.
This favors \emph{sample diversity}: experience collected under both older
and newer behavior policies contributes to reducing the finite-data
statistical error.

In contrast, the TD optimization term favors replay distributions with small
$\mathcal C_{\mathcal B}(\pi,\mathbf w)$, i.e., distributions whose sampled
features provide strong coverage of the current target policy.
When the behavior policy tracks the target policy, this can favor more recent
experience.
Hence replay sampling faces a diversity--coverage trade-off:
uniform sampling provides the best statistical limit, whereas a more
coverage-focused distribution can require fewer TD updates to approach its
corresponding limit.
A concrete numerical example illustrating this trade-off is shown in
Fig.~\ref{fig:buffer-trade-off} in the Appendix~\ref{appendix:fixed-replay-budget}.

\paragraph{Replay design and sample complexity.}
The decomposition above suggests choosing the replay distribution together
with the evaluation budget.
For a prescribed evaluation accuracy $\epsilon_{\rm c}$ and fixed number
$T$ of target-network iterations, we consider
\begin{equation}
\label{eq:replay-budget-program}
\min_{\mathbf w\in\Delta^n,\;K\in\mathbb N} K
\qquad\text{s.t.}\quad
\frac{c_{\mathrm{stat}}\mathcal V_{\mathcal B}(\pi,\mathbf w)}
{(1-\gamma)^3}
+
\frac{c_{\mathrm{TD}}\mathcal C_{\mathcal B}(\pi,\mathbf w)}
{K(1-\gamma)^4}
+
\frac{3\gamma^{2T}}{(1-\gamma)^2}
\leq \epsilon_{\rm c}^2 .
\end{equation}
Since each TD update consumes one replay sample, minimizing $K$ for fixed
$T$ is equivalent to minimizing the total replay-sample complexity $TK$.
For a fixed replay budget $K$, the corresponding sampling-design problem
and its interpretation are provided in
Appendix~\ref{appendix:fixed-replay-budget}.

\begin{remark}[From policy evaluation to policy optimization]
Lemma~\ref{prop:replay-evaluation} and
\eqref{eq:replay-budget-program} optimize replay under the
$\mathbf A_\pi$-weighted policy-evaluation metric. Such guarantee can be transferred to gradient estimation as well. More precisely, for each reference policy $\pi$, by~\eqref{eq:gradient-error-operator} and Assumption~\ref{assumption:anchor-gradient}, we have
$\mathbb E\big[\|\hat\nabla J_\rho(\pi,\hat\xi_T)-\nabla J_\rho(\pi)\|_2^2\big]
\leq\kappa_{\mathrm g}^2\mathbb E\big[\|\hat\xi_T-\xi^\star\|_{\mathbf A_\pi}^2\big]$.
Thus the same replay program controls gradient accuracy at the reference policy.
\end{remark}

Empirical results for the resulting replay-sampling strategy are reported in
Appendix~\ref{appendix:replay-experiments}.

\section{Functional Critics for Posterior-Sampling-Style Exploration}
\label{sec-psrl}

\begin{figure}[t]
    \centering
    \includegraphics[width=\textwidth]{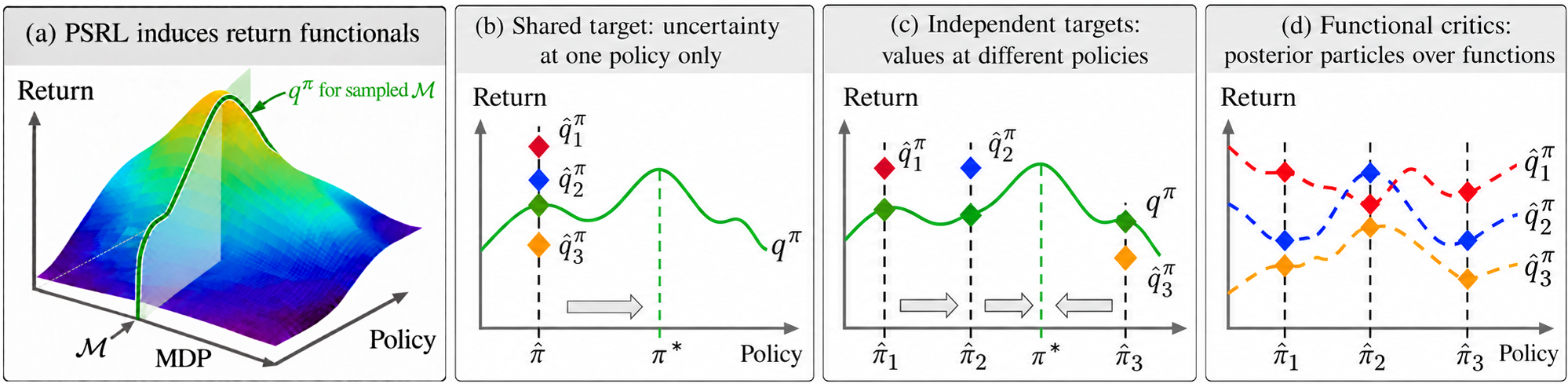}
    \caption{\small
    From PSRL to functional-critic ensembles.
    (a) An MDP sample induces a return functional over policy space.
    (b) With a shared target policy, a standard critic ensemble represents
    uncertainty only at that policy.
    (c) With independent target policies, ensemble members estimate different
    policy-dependent quantities rather than samples from a common posterior
    over return functions.
    (d) Functional-critic ensembles represent particles over entire
    policy-conditioned return functions.}
    \label{fig:psrl}
\end{figure}

As reviewed in Section~\ref{sec-prob_def}, PSRL maintains a posterior over MDPs and, for each rollout, executes the
optimal policy of a sampled MDP.
Because return depends jointly on the MDP and the policy, each sampled MDP
defines a return functional over the entire policy space (Fig.~\ref{fig:psrl} (a)).
This observation motivates a model-free posterior-sampling-style construction
that represents uncertainty jointly across policy space, rather than only at
individual policies.

Model-free exploration methods often use ensembles of standard critics
\citep{osband2016deep,osband2018randomized,osband2019deep}.
Fig.~\ref{fig:psrl} (b)--(c) illustrates why such ensembles do not naturally
represent uncertainty jointly over return functions across policy space.
With a shared target critic, all critics estimate the value of the same
policy and therefore characterize uncertainty only at that policy
(Fig.~\ref{fig:psrl} (b)).
With independent target critics, ensemble members instead estimate
different policy-dependent quantities, which are not samples of a common
return functional over policy space
(Fig.~\ref{fig:psrl} (c)).

Functional critics provide a direct representation of the relevant object.
Each functional critic models an entire policy-conditioned return
$\hat Q(\pi,s,a)$, so an ensemble represents a collection of candidate return
functionals over policy space (Fig.~\ref{fig:psrl} (d)).
Sampling one such functional and improving an actor against it therefore
motivates a model-free analogue of posterior-sampling-style exploration,
without explicitly maintaining or solving sampled MDP models.

We instantiate this idea with ensembles of actors and functional critics.
During actor optimization, critics are sampled from the ensemble and actors
are updated using the corresponding functional gradients.
The complete procedure is given in
Algorithm~\ref{alg-ensemble-algorithm} in
Appendix~\ref{appendix:ensemble-psrl}.

\section{Practical Implementation}\label{sec-experiment}

We instantiate the functional actor-critic framework with neural networks and
the ensemble construction of Section~\ref{sec-psrl}.
The main experiments evaluate the resulting neural functional actor-critic system; 
heuristic neural instantiations of evaluation-trust behavior reuse and replay-sampling mechanisms, together with targeted empirical evaluations, are reported in the 
Appendix~\ref{appendix:tr-experiments} and~\ref{appendix:replay-experiments}.

\paragraph{Neural Functional Critics.}
We maintain an ensemble of functional critics
$\{\hat Q^{(i)}\}_{i=1}^n$.
Each critic consists of an actor encoder $E_{\rm act}^{(i)}$, a state-action
encoder $E_{\rm sa}^{(i)}$, and a joint encoder $E_{\rm joint}^{(i)}$:
\begin{equation}
\label{eq:func_q_nn}
\hat Q^{(i)}(\pi_\theta,s,a)
=
E_{\rm joint}^{(i)}
\left(
E_{\rm act}^{(i)}(\pi_\theta),
E_{\rm sa}^{(i)}(s,a)
\right).
\end{equation}
Target critics use delayed copies of the state-action and joint encoders,
while sharing the actor encoder.
Each critic is trained by target-based TD on actor-transition pairs drawn
from the product of the current actor ensemble and replay data.

To represent a policy efficiently, $E_{\rm act}$ evaluates the actor at a
small set of trainable states $\{\zeta_j\}$ and processes the resulting
actions and hidden activations with a transformer encoder.
The evaluation states are learned end-to-end together with the critic.

\paragraph{Actor Ensemble and Exploration.}
We use an ensemble of deterministic actors
$\{\pi_{\theta^{(i)}}\}_{i=1}^n$, avoiding additional hyperparameters
associated with stochastic actors.
Each functional critic is trained across the actor ensemble, while every
actor update independently samples a critic from which to obtain the
functional policy gradient.
For deterministic actors, gradients propagate through both the action and
explicit policy inputs of the functional critic; the resulting
gradient is given in Appendix~\ref{appendix:det-functional-gradient}.


\paragraph{Omitted Standard Heuristics.}
Our implementation intentionally omits several heuristics widely used in
modern off-policy AC practice: clipped/minimum double-Q targets, exploration
noise, target-policy smoothing, and entropy regularization
\citep{fujimoto2018addressing,haarnoja2018soft}.
Further architectural details and the complete training procedure are provided in
Appendix~\ref{appendix:deep-algorithm}.

\subsection{Empirical Results}
\label{sec:exp_results}
\begin{figure}[tb]
    \centering
    \vspace{-0.1in}
    \begin{subfigure}[t]{0.24\textwidth}
        \centering
        \includegraphics[width=\linewidth]{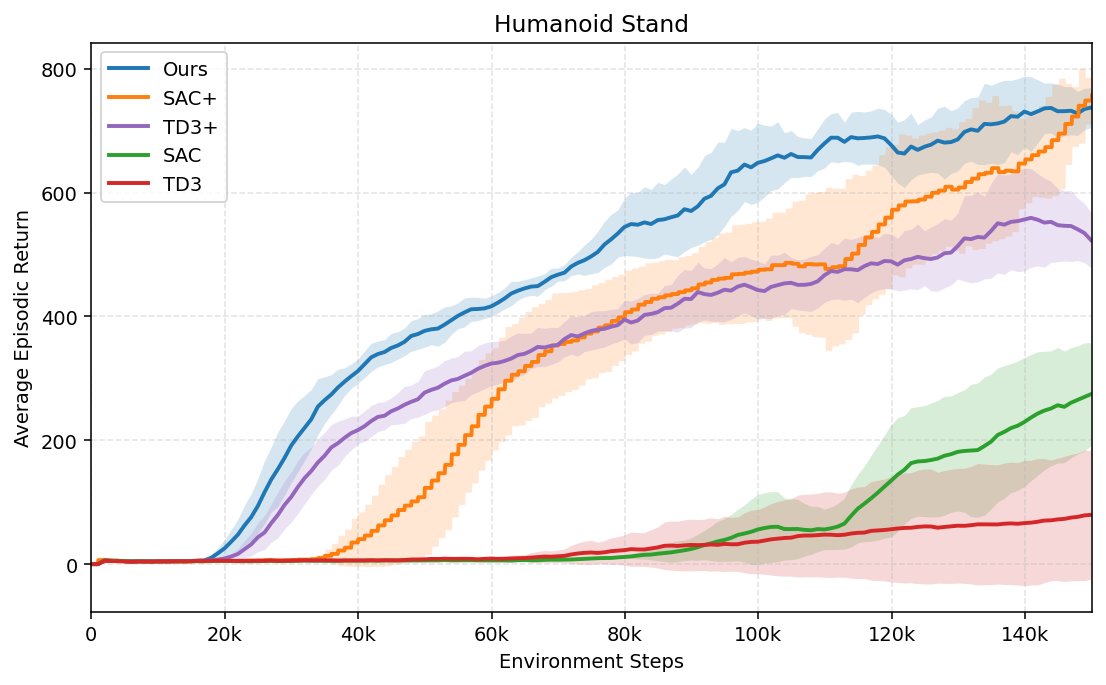}
        \caption{Humanoid\_stand}
        \label{fig:hum_stand}
    \end{subfigure}
    \hfill
    \begin{subfigure}[t]{0.24\textwidth}
        \centering
        \includegraphics[width=\linewidth]{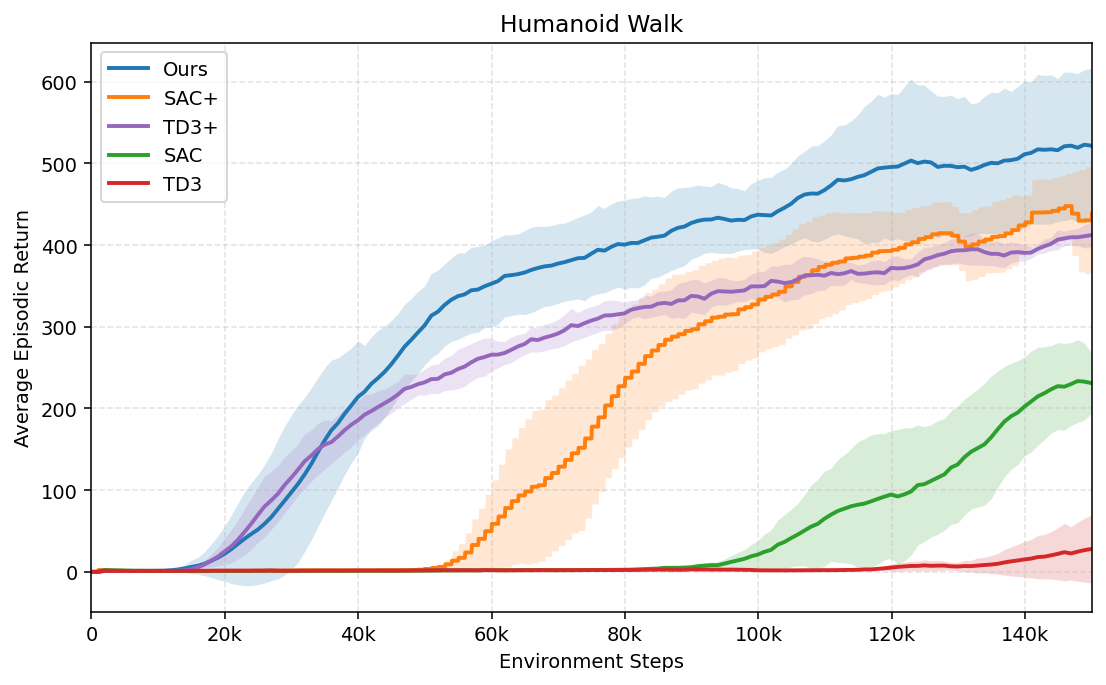}
        \caption{Humanoid\_walk}
        \label{fig:hum_walk}
    \end{subfigure}
    \hfill
    \begin{subfigure}[t]{0.24\textwidth}
        \centering
        \includegraphics[width=\linewidth]{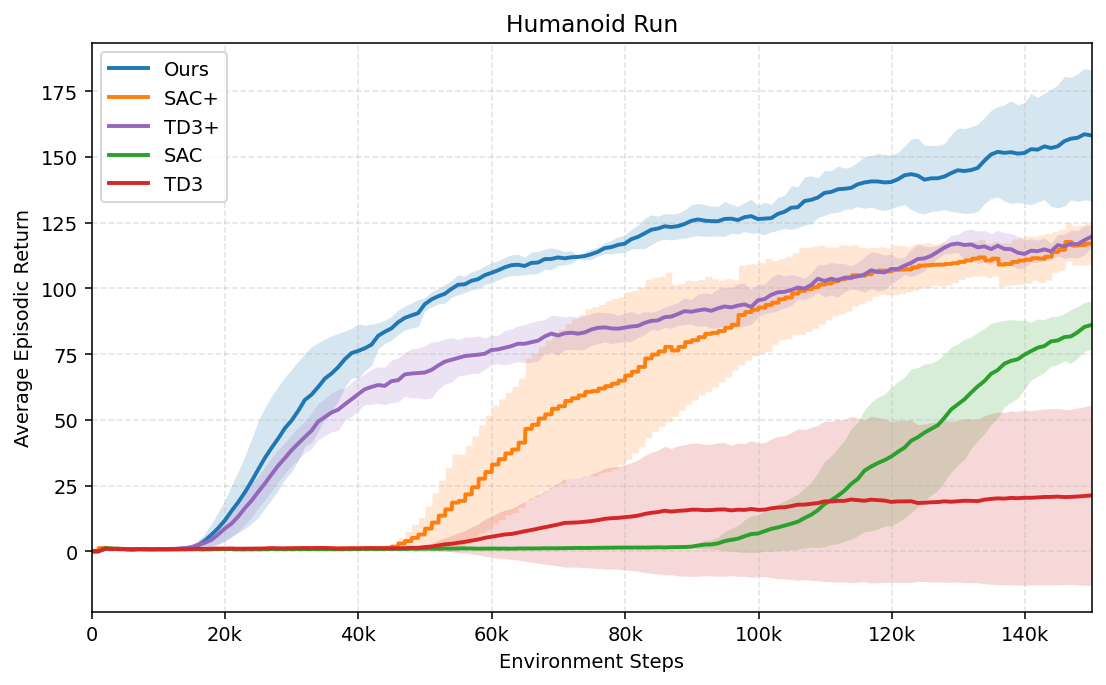}
        \caption{Humanoid\_run}
        \label{fig:hum_run}
    \end{subfigure}
    \hfill
    \begin{subfigure}[t]{0.24\textwidth}
        \centering
        \includegraphics[width=\linewidth]{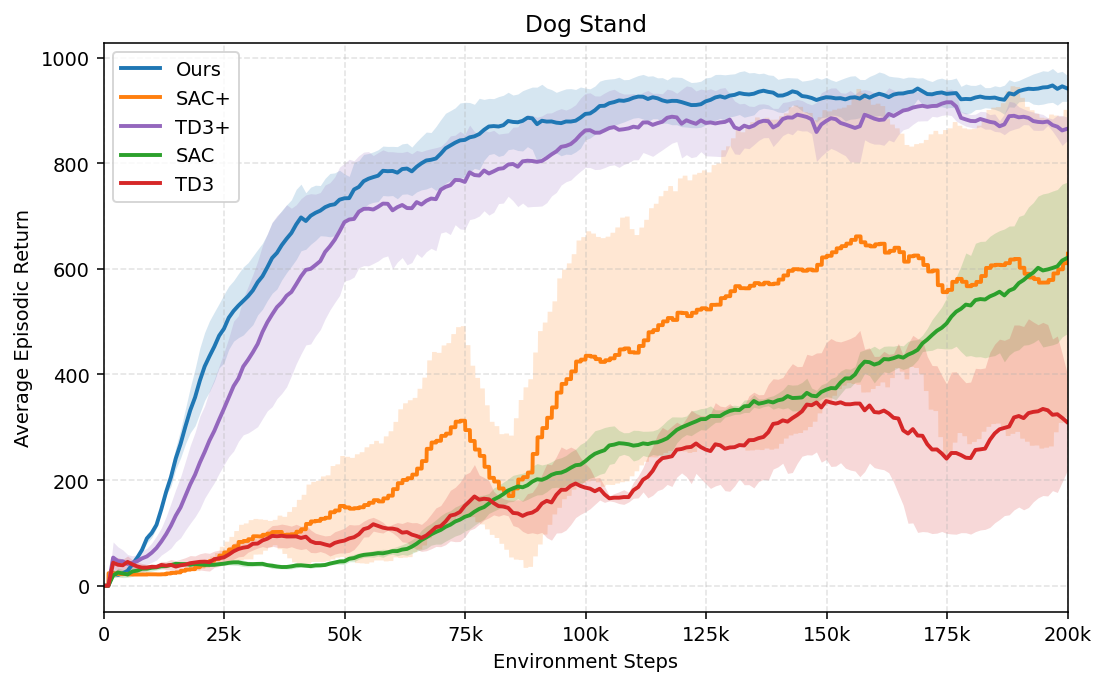}
        \caption{Dog\_stand}
        \label{fig:dog_stand}
    \end{subfigure}

    \vspace{0.5em}

    \begin{subfigure}[t]{0.24\textwidth}
        \centering
        \includegraphics[width=\linewidth]{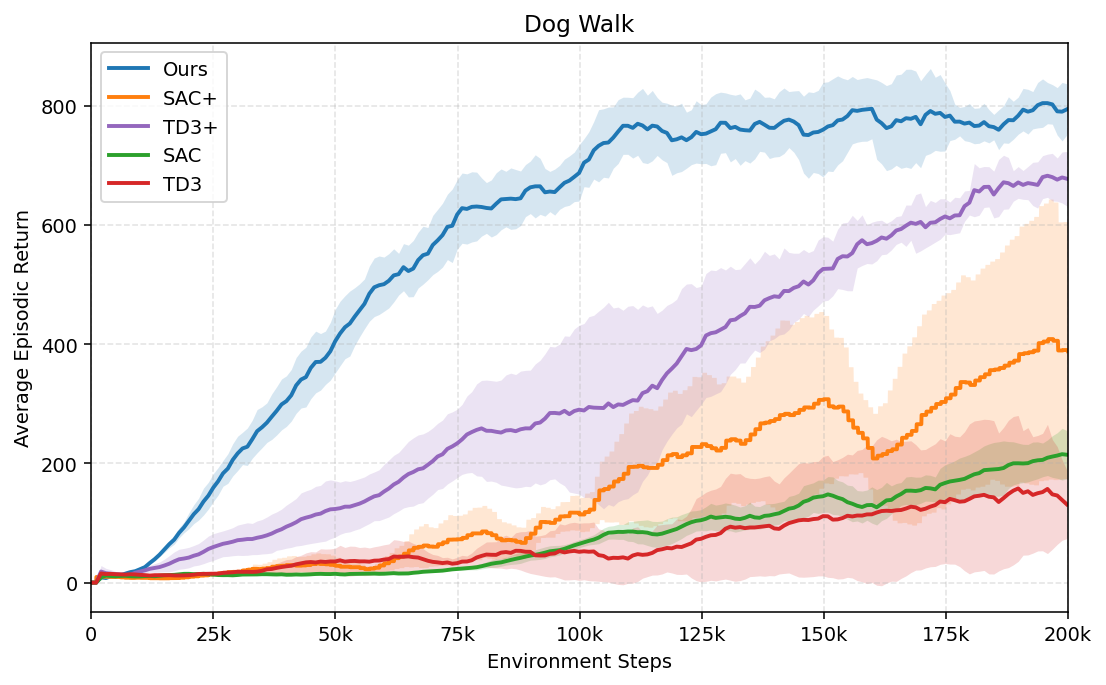}
        \caption{Dog\_walk}
        \label{fig:dog_walk}
    \end{subfigure}
    \hfill
    \begin{subfigure}[t]{0.24\textwidth}
        \centering
        \includegraphics[width=\linewidth]{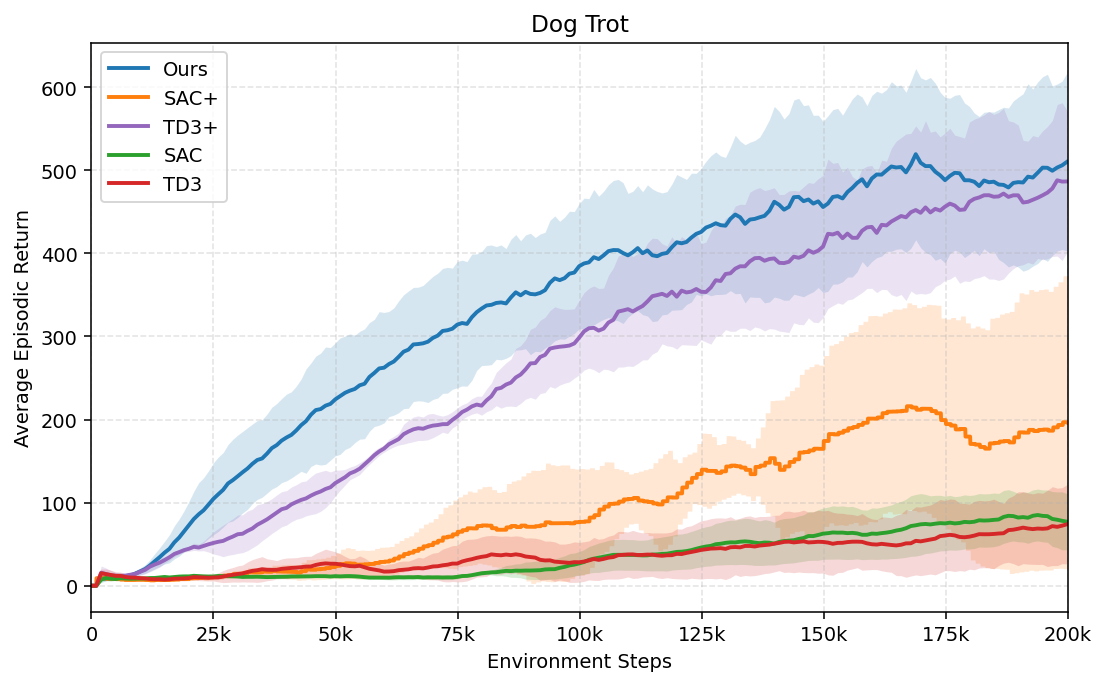}
        \caption{Dog\_trot}
        \label{fig:dog_trot}
    \end{subfigure}
    \hfill
    \begin{subfigure}[t]{0.24\textwidth}
        \centering
        \includegraphics[width=\linewidth]{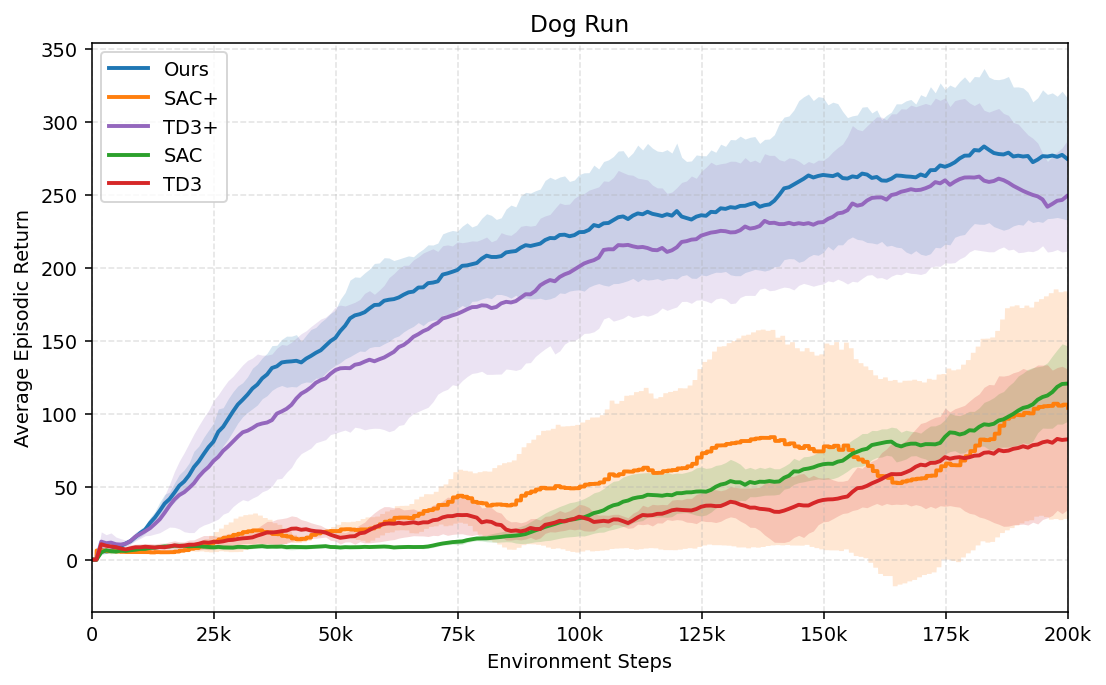}
        \caption{Dog\_run}
        \label{fig:dog_run}
    \end{subfigure}
    \hfill
    \begin{subfigure}[t]{0.24\textwidth}
        \centering
        \includegraphics[width=\linewidth]{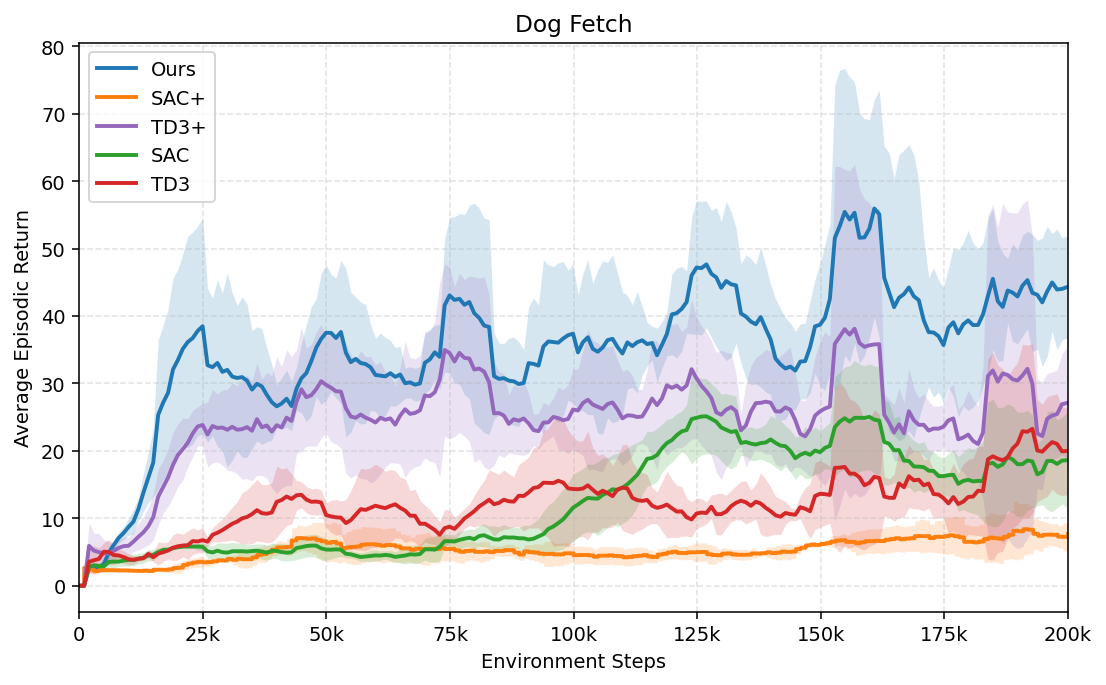}
        \caption{Dog\_fetch}
        \label{fig:dog_fetch}
    \end{subfigure}

    \caption{Average episodic return on Dog and Humanoid tasks from DM Control.
    Curves show the mean over four seeds and shaded regions indicate one standard
    deviation; all experiments use four parallel environments.}
    \label{fig:main_experiment}
    \vspace{-0.1in}
\end{figure}

\paragraph{Baselines and Setup.}
We evaluate on the Dog and Humanoid task series from the DeepMind Control
Suite \citep{tunyasuvunakool2020dmc}.
These two series feature relatively high-dimensional state and action spaces
and distinct morphologies, providing challenging and complementary testbeds;
we evaluate the full series rather than selecting individual tasks.

We compare against SAC and TD3
\citep{haarnoja2018soft,fujimoto2018addressing}, together with strengthened
variants SAC+ and TD3+.
Recent high-performance RL systems increasingly rely on aggressive critic
updating, often expressed through a high update-to-data (UTD) ratio, together
with critic ensembles and layer normalization
\citep{chen2021redq,li2023efficient,ball2023efficient}, a recipe underlying
several recent strong online and offline RL methods
\citep{zhou2024efficient,xiao2025efficient,nauman2024bigger,seo2025fasttd3}.
We incorporate this recipe into SAC and TD3 using the extensively tested ALF
codebase \citep{Xu2021ALF} and carefully tune the resulting SAC+ and TD3+
baselines.
As shown in Fig.~\ref{fig:main_experiment}, both substantially outperform
their vanilla counterparts, confirming that the modern training recipe itself
materially strengthens the baselines and providing substantially stronger
comparators for our method.

After independent tuning, both SAC+ and TD3+ use 10 critic updates per actor
update. We use the same 10:1 critic-to-actor update ratio for our method in
all main experiments.
Our method, SAC+, and TD3+ use 10 critics, while the vanilla SAC and TD3
retain their original two-critic architectures.
Functional-critic updates additionally reuse each sampled transition across
the actor ensemble; the comparisons therefore match environment interaction,
critic ensemble size, and nominal update frequency, but not total critic-side
computation.
We match network backbones across methods where their architectures permit,
and our method uses a single hyperparameter configuration across all eight
tasks.
Full experimental protocols, baseline configurations, architectures, and
hyperparameters are provided in Appendix~\ref{appendix:hp}.

Prior policy-conditioned value-function methods are discussed in
Appendix~\ref{sec:related-pcvf}.
Their published implementations and evaluations operate in substantially
different training regimes, and a subsequent benchmark found the major
variants noncompetitive with TD3 \citep{griesbach2023improving}.
We therefore focus the main comparison on carefully tuned modern SAC+ and
TD3+ baselines.

\paragraph{Results.}
Figure~\ref{fig:main_experiment} shows that our method matches or exceeds the
strongest baseline in mean return across all evaluated Dog and Humanoid tasks.
It exhibits stable learning across independent runs and favorable
environment-sample efficiency, despite omitting the standard off-policy AC
heuristics described above.
These results demonstrate the practical viability of the proposed neural
functional actor-critic system on challenging continuous-control tasks.
Targeted experiments in
Appendices~\ref{appendix:replay-experiments} and
\ref{appendix:tr-experiments} further illustrate how the theoretically
motivated replay-sampling and trust-coverage mechanisms can be instantiated
with learned neural features.

\section{Related Work}
\label{sec:related}

\subsection{Policy-Conditioned and Parameter-Based Value Functions}
\label{sec:related-pcvf}

Making the value function an explicit function of the policy has been explored
in several forms, differing both in how the policy is represented and in how
the policy-conditioned value function is used for actor optimization.

\paragraph{Policy-conditioned value functions.}
Policy Evaluation Networks (PVNs) \citep{harb2020policy} represent a policy
through its behavior on a set of probing states and learn a value function over
a pre-collected set of policies. Once trained, the differentiable policy input
can be used to optimize a new policy by backpropagating through the learned
value function. The original experiments focused on relatively small policy
networks and simple control problems rather than online off-policy
actor-critic learning.

Parameter-Based Value Functions \citep{faccio2020parameter} instead use the
policy parameters themselves as inputs to the value function.
The state-value variants PSSVF and PSVF update the policy directly through the
policy dependence of the learned value function, without the standard
action-gradient pathway used by deterministic actor-critic methods.
Importantly, the $Q$-function variant PAVF is different: its actor update
contains both a DPG-style action-gradient term and a derivative through the
explicit policy input. PAVF is therefore the closest formal antecedent of the
two-path functional gradient used in our deterministic implementation; our
distinction from PAVF is not the presence of this second gradient term.
A subsequent work \citep{faccio2022general} replaces raw policy parameters
with behavior on a small set of learned probing states, further reducing the
dependence of the representation on the particular policy-network
parameterization.

\paragraph{Policy embeddings for value estimation.}
PeVFA \citep{tang2022inputting} uses policy conditioning primarily to improve value
estimation. It learns policy representations from policy-network parameters or
state--action behavior using auxiliary representation-learning objectives.
Its practical PPO-PeVFA instantiation retains the standard PPO actor update:
the policy-conditioned value function is used for value estimation rather than
introducing the additional policy-input derivative used by PAVF and our
method. PPO-PeVFA is evaluated in an on-policy setting, with historical
policies and their associated trajectories used to train the value function,
and studies generalization primarily along a policy-improvement trajectory.
Its ablations also illustrate an important practical challenge for
policy-conditioned value functions: high-dimensional raw parameter
representations can degrade performance, motivating more structured policy
representations.

\paragraph{Relation to our method.}
Our implementation differs from prior policy-conditioned value-function
methods along several distinct design axes. Table~\ref{tab:related-axes}
summarizes the main differences.

\begin{table}[t]
\caption{Comparison with prior policy-conditioned value-function methods.
The listed design differences should not all be attributed to policy
conditioning itself; see the discussion below.}
\label{tab:related-axes}
\centering
\footnotesize
\setlength{\tabcolsep}{4pt}
\renewcommand{\arraystretch}{1.15}
\begin{tabular}{@{}lp{2.5cm}p{2.4cm}p{2.6cm}p{2.5cm}@{}}
\toprule
& \textbf{Policy representation}
& \textbf{Critic training}
& \textbf{Policy distribution}
& \textbf{Actor update} \\
\midrule
PSSVF / PSVF
& raw parameters
& MC / TD
& parameter perturbations
& policy-input derivative only \\
PVN
& probe-state actions
& MC on fixed policy set
& pre-collected policies
& policy-input derivative only \\
PAVF
& raw parameters
& TD
& parameter perturbations
& DPG $+$ policy term \\
PeVFA
& learned OPR / SPR
& MC, on-policy
& single improvement path
& standard PPO \\
\midrule
Ours
& probe-state actions $+$ hidden activations; end-to-end transformer
& off-policy target-based TD $+$ replay
& concurrent actor ensemble
& DPG $+$ policy term \\
\bottomrule
\end{tabular}
\end{table}

First, our policy representation is behaviorally grounded: the actor encoder
evaluates the policy on learned probing states and processes the resulting
actions and hidden activations with a transformer. The representation is
trained end-to-end through the critic TD objective rather than through a
separate representation-learning loss.

Second, the functional critic is trained by off-policy target-based TD on
replay data and across the current actor ensemble. Consequently, a replayed
transition can provide Bellman supervision for multiple policies without
requiring a separately collected trajectory for each policy.

Third, the critic is trained on a concurrently evolving ensemble of actors,
rather than only on parameter perturbations or policies lying along a single
policy-improvement trajectory.

Finally, our deterministic actor update retains the standard DPG
action-gradient pathway while also differentiating through the explicit policy
input of the functional critic, as given in
Eq.~\eqref{eq:det-functional-gradient}. This gradient structure itself is not
unique to our work: PAVF also contains both pathways. 
Our contribution instead lies in combining this gradient structure with a
scalable neural policy representation, modern off-policy actor-critic
training, and a theoretical framework that analyzes functional critics under
partial coverage and evolving behavior policies.

More broadly, our work connects the policy-conditioned value-function
literature to modern off-policy actor-critic design and analysis.
Prior policy-conditioned approaches established that value functions can
generalize across policies, but were largely developed in substantially
different training regimes, often using Monte Carlo evaluation, parameter
perturbations, or comparatively small-scale actor-critic implementations.
Our method instead embeds policy conditioning into a modern off-policy
actor-critic system based on replay, target-based TD, high-rate critic
optimization, and critic ensembles.

At the same time, our theoretical analysis studies the functional-critic
representation under practical off-policy conditions that are largely absent
from the earlier policy-conditioned literature, including target-based TD,
partial feature coverage, and evolving behavior policies.
Conversely, existing off-policy actor-critic theory typically analyzes
ordinary state-action critics rather than value functions that explicitly
generalize across policies.
Thus, an important distinction of our work is to bring these two lines
together: policy-conditioned value-function modeling on the one hand, and
modern off-policy actor-critic training and convergence analysis on the other.

These distinctions should not be interpreted as an attribution of the
empirical gains to policy conditioning alone. Off-policy TD and replay are
standard actor-critic ingredients, while the concurrent policy distribution is
a property of the actor ensemble. Likewise, the standard action-gradient
pathway is inherited from conventional deterministic actor-critic methods.
The explicit policy conditioning and its learned representation are the
functional-critic-specific ingredients. Our main experiments evaluate the
resulting integrated system and do not isolate the contribution of each of
these design choices individually.

\paragraph{Prior empirical comparisons.}
A subsequent benchmark \citep{griesbach2023improving} found several earlier
policy-conditioned value-function methods noncompetitive with standard
algorithms such as TD3 \citep{fujimoto2018addressing}. That benchmark predates
the critic-ensemble, high-update-ratio, and layer-normalization recipe used by
our modern baselines, so we do not interpret it as evidence that policy
conditioning is intrinsically unsuitable for modern off-policy learning.
For this reason, the main experiments compare against carefully tuned SAC+
and TD3+ baselines, while the methodological relationship to earlier
policy-conditioned approaches is documented here.

\subsection{Off-Policy Actor-Critic Theory and Coverage Conditions}
\label{sec:related-theory}

Convergence analyses for off-policy actor-critic have largely relied on
gradient or emphatic TD
\citep{sutton2009fast,sutton2016emphatic,imani2018off,zhang2020provably},
or on target networks and truncation to stabilize TD with function
approximation
\citep{carvalho2020new,zhang2021breaking,chen2023target};
target-based actor-critic under linear approximation was further analyzed by
\citet{barakat2022analysis}.
These results typically assume a fixed behavior policy together with strong
coverage conditions.
Feature-based coverage quantities are also standard in offline RL with linear
structure
\citep{duan2020minimax,min2021variance,xiong2022nearly,yin2022near},
primarily in least-squares or pessimism-based analyses.

These analyses generally study ordinary state--action value approximators,
rather than a single critic that explicitly conditions on and generalizes
across evolving policies.
Our analysis brings functional critics into this off-policy setting,
establishing a target-based TD guarantee under target-relevant feature
coverage and incorporating it into an actor-critic scheme with evolving
behavior policies.

\subsection{Modern Off-Policy Actor-Critic Practice}
\label{sec:related-modern}

Modern continuous-control systems often combine large critic update-to-data
ratios, critic ensembles, and normalization techniques
\citep{chen2021redq,li2023efficient,ball2023efficient},
a recipe used by several recent high-performing RL methods
\citep{nauman2024bigger,zhou2024efficient,xiao2025efficient,seo2025fasttd3}.
The earlier policy-conditioned value-function methods discussed in
Section~\ref{sec:related-pcvf} were developed in substantially different
training regimes and do not establish how policy conditioning behaves within
this modern off-policy recipe.

Our implementation brings functional critics into this regime, adopting these
ingredients where applicable, while our SAC+ and TD3+ baselines apply the same
modern training recipe to SAC \citep{haarnoja2018soft} and TD3
\citep{fujimoto2018addressing} within the same ALF codebase
\citep{Xu2021ALF}.
As reflected by the large gap between these strengthened baselines and their
vanilla counterparts in Section~\ref{sec:exp_results}, this training recipe
itself accounts for a substantial part of absolute performance and should not
be attributed to the functional-critic representation.

\subsection{Posterior Sampling and Model-Free Exploration}
\label{sec:related-psrl}

Posterior Sampling for Reinforcement Learning (PSRL)
\citep{osband2013more} samples an environment hypothesis and acts according
to the corresponding policy, but explicitly maintaining and solving sampled
models is difficult in high-dimensional continuous-control problems.
Model-free approximations therefore often use bootstrapped critic ensembles
or randomized priors to induce temporally coherent exploration
\citep{osband2016deep,osband2018randomized,osband2019deep}.
Our construction differs in representing entire policy-conditioned return
functions, rather than critic uncertainty only at individual policies,
motivating the posterior-sampling-style construction of
Section~\ref{sec-psrl}.
We do not interpret the learned critic ensemble as an exact Bayesian
posterior: in particular, our critics do not use bootstrap masks or randomized
priors, and critic samples are used during actor optimization rather than as
episode-level sampled environment hypotheses.

\section{Conclusion}

We studied functional critics as a shared framework for off-policy
actor-critic learning. Under linear functional approximation, we establish
convergence with target-based TD, partial feature coverage, and evolving
behavior policies, and develop a replay-sampling analysis that exposes a
diversity--coverage trade-off. Functional-critic ensembles further motivate
model-free posterior-sampling-style exploration over policy-conditioned
return functions. Our neural functional actor-critic system performs strongly
on challenging continuous-control tasks while omitting several standard
off-policy AC heuristics; targeted Appendix experiments further illustrate
neural instantiations of the evaluation-trust and replay-sampling mechanisms.
Our theory remains limited to linear functional approximation, and the
exploration construction is posterior-sampling-inspired rather than an exact
Bayesian posterior approximation. Extending the guarantees to nonlinear
critics and developing calibrated functional uncertainty are important
directions for future work.
\vspace{0.1in}
\noindent Moving forward, several promising directions remain:
\begin{itemize}
    \item \textbf{Trust-Criteria Implementation:} 
    The trust coverage criteria in our theoretical Algorithm~\ref{alg-linear-AC} utilize the linear functional representation and its associated covariance matrices. While our minimalist implementation demonstrates robustness without them, developing a sample-based, neural network-compatible estimator for $\mathcal{C}(\pi, \mu)$ and $\Delta_{k,t}$ could further stabilize deep RL training and improve sample efficiency in highly non-stationary environments.    
    \item \textbf{Adaptive Behavior Control:} 
    Our analysis suggests that behavior policies should be optimized to maximize the condition of the feature covariance $\tilde{\mathbf{A}}_k$. Actively adapting $\mu$ to minimize the evaluation trust metric $\mathcal{C}(\pi, \mu)$ presents a promising path toward superior sample efficiency in off-policy setting.  
    \item \textbf{Trust-Region Optimization:} 
    Integrating the gradient generalization metric $\Delta_{k,t}$ directly into the actor's objective as a functional constraint—mirroring the success of KL-divergence constraints in TRPO and PPO—could provide a more robust, data-aware mechanism for policy improvement. This would allow the actor to automatically slow its update rate when entering feature regimes where the functional critic's predictions are less reliable.
\end{itemize}

\bibliographystyle{ims}
\bibliography{iclr2027_conference}

\newpage
\appendix

\input{iclr2027_appendix}

\end{document}

%% file: math_commands.tex
\usepackage{amsmath,amsfonts,bm}

\newcommand{\cA}{\mathcal{A}}
\newcommand{\cC}{\mathcal{C}}
\newcommand{\cE}{\mathcal{E}}

\newcommand{\cS}{\mathcal{S}}

\newcommand{\cH}{\mathcal{H}}
\newcommand{\cY}{\mathcal{Y}}
\newcommand{\bA}{\mathbf{A}}
\newcommand{\bD}{\mathbf{D}}
\newcommand{\bF}{\mathbf{F}}
\newcommand{\bI}{\mathbf{I}}
\newcommand{\bP}{\mathbf{P}}
\newcommand{\bPi}{\boldsymbol{\Pi}}
\newcommand{\bPhi}{\boldsymbol{\Phi}}
\newcommand{\bQ}{\mathbf{Q}}
\newcommand{\bU}{\mathbf{U}}
\newcommand{\br}{\mathbf{r}}
\newcommand{\bu}{\mathbf{u}}
\newcommand{\bv}{\mathbf{v}}
\newcommand{\bx}{\mathbf{x}}
\newcommand{\by}{\mathbf{y}}
\newcommand{\EE}{\mathbb{E}}
\newcommand{\RR}{\mathbb{R}}

\def\eqref#1{equation~\ref{#1}}

\def\1{\bm{1}}

\DeclareMathAlphabet{\mathsfit}{\encodingdefault}{\sfdefault}{m}{sl}
\SetMathAlphabet{\mathsfit}{bold}{\encodingdefault}{\sfdefault}{bx}{n}

\newcommand{\E}{\mathbb{E}}



%% file: iclr2027_appendix.tex

\section{Deep Functional Actor--Critic Implementation}
\label{appendix:deep-algorithm}

This section provides additional architectural and algorithmic details for the
neural functional actor--critic implementation used in
Section~\ref{sec-experiment}.

\subsection{Neural Functional Critic Architecture}

We maintain an ensemble of $n$ functional critics
$\{\hat Q^{(i)}\}_{i=1}^n$.
Each critic contains three components: an actor encoder
$E_{\rm act}^{(i)}$, a state--action encoder $E_{\rm sa}^{(i)}$, and a joint
encoder $E_{\rm joint}^{(i)}$. Writing
$\Xi^{(i)}
=
\{\xi_{\rm act}^{(i)},\xi_{\rm sa}^{(i)},\xi_{\rm joint}^{(i)}\}$,
the critic takes the form
\begin{equation}
\hat Q^{(i)}(\pi_\theta,s,a;\Xi^{(i)})
=
E_{\rm joint}^{(i)}
\left(
E_{\rm act}^{(i)}(\pi_\theta;\xi_{\rm act}^{(i)}),
E_{\rm sa}^{(i)}(s,a;\xi_{\rm sa}^{(i)});
\xi_{\rm joint}^{(i)}
\right).
\label{eq:appendix-func-q}
\end{equation}

For each functional critic, we maintain a target critic with delayed copies of
the state--action and joint encoders, while sharing the actor encoder:
\begin{equation}
\hat Q_{\rm tar}^{(i)}(\pi_\theta,s,a)
=
E_{\rm joint}^{(i)}
\left(
E_{\rm act}^{(i)}(\pi_\theta;\xi_{\rm act}^{(i)}),
E_{\rm sa}^{(i)}(s,a;\xi_{\rm sa}^{'(i)});
\xi_{\rm joint}^{'(i)}
\right).
\label{eq:appendix-target-q}
\end{equation}
Thus, policy representations evolve jointly with the primary critic, whereas
the state--action and joint components of the target critic are updated
slowly.

Given a policy $\pi$, transition $(s,a,r,s')$, and target critic $i$, the TD
target is
\begin{equation}
y^{(i)}(\pi;s,a,r,s')
=
r
+
\gamma
\hat Q_{\rm tar}^{(i)}
\left(
\pi,s',\pi(s')
\right).
\label{eq:appendix-td-target}
\end{equation}

A key distinction from a standard critic ensemble is that each functional
critic is trained across the \emph{entire} current actor ensemble
\begin{equation}
\mathcal A_t
=
\{\pi_{\theta_t^{(j)}}\}_{j=1}^n.
\end{equation}
For a transition minibatch $B_C$, critic $i$ is therefore trained on the
product set $\mathcal A_t\times B_C$ using
\begin{equation}
\mathcal L_Q^{(i)}
=
\frac{1}{|\mathcal A_t||B_C|}
\sum_{\pi\in\mathcal A_t}
\sum_{\tau\in B_C}
\left[
y^{(i)}(\pi;\tau)
-
\hat Q^{(i)}(\pi,s_\tau,a_\tau)
\right]^2 .
\label{eq:appendix-func-q-loss}
\end{equation}
This cross-policy training is what allows every critic particle to provide
value and gradient information for every actor in the ensemble.

\subsection{Deterministic Functional Policy Gradient}
\label{appendix:det-functional-gradient}

The functional policy gradient in
Eq.~\eqref{eq-our-gradient-formula} is written for stochastic policies.
Our continuous-control experiments instead use deterministic actors
$a=\pi_\theta(s)$.
For a functional critic
$\hat Q(\pi_\theta,s,a;\xi)$, we define the corresponding deterministic
surrogate objective as
\begin{equation}
\label{eq:det-functional-objective}
J_\rho(\pi_\theta;\xi)
:=
\E_{s\sim\rho}
\left[
\hat Q(\pi_\theta,s,\pi_\theta(s);\xi)
\right].
\end{equation}
The policy parameters therefore affect the objective through two distinct
computational paths: through the action $a=\pi_\theta(s)$ and through the
explicit policy input of the functional critic. Applying the chain rule gives
\begin{equation}
\label{eq:det-functional-gradient}
\nabla_\theta J_\rho(\pi_\theta;\xi)
=
\E_{s\sim\rho}\!\left[
\left.
\nabla_\theta\pi_\theta(s)
\nabla_a
\hat Q(\pi_\theta,s,a;\xi)
\right|_{a=\pi_\theta(s)}
+
\left.
\nabla_\theta^{\mathrm{pol}}
\hat Q(\pi_\theta,s,a;\xi)
\right|_{a=\pi_\theta(s)}
\right].
\end{equation}
Here,
$\nabla_\theta^{\mathrm{pol}}\hat Q(\pi_\theta,s,a;\xi)$
denotes differentiation through the explicit policy input of the functional
critic while holding its state and action inputs fixed.
Thus, the first term in Eq.~\eqref{eq:det-functional-gradient} is the standard
deterministic policy-gradient term, whereas the second term captures the
additional dependence of the predicted return on the policy represented by
the functional critic.

In our implementation, both terms in
Eq.~\eqref{eq:det-functional-gradient} are backpropagated to the actor.
Specifically, the first term propagates through the action input
$a=\pi_\theta(s)$, while the second propagates through the actor encoder
$E_{\rm act}(\pi_\theta)$ used as the explicit policy input to the critic.
For the policy representation described in Section~\ref{sec-experiment},
this second pathway differentiates through the actor outputs and hidden
activations evaluated at the trainable evaluation states
$\{\zeta_j\}$.
During the actor update, the critic parameters, evaluation states, and the
state--action inputs of this second pathway are held fixed.

For an actor update using critic $\hat Q^{(i)}$, we therefore optimize
Eq.~\eqref{eq:det-functional-objective} with $\hat Q=\hat Q^{(i)}$ and
backpropagate the full gradient in
Eq.~\eqref{eq:det-functional-gradient}, rather than detaching the explicit
policy input.

\subsection{Actor Encoder}

The actor encoder must represent a high-dimensional policy compactly while
remaining differentiable with respect to the actor parameters.
Motivated by evaluation functionals, we introduce a set of trainable
evaluation states
\begin{equation}
\mathcal Z
=
\{\zeta_k\}_{k=1}^{N_\zeta},
\end{equation}
which serve as learned probes of the input policy.
For each $\zeta_k$, the actor is evaluated and the resulting action together
with selected hidden activations is used to construct a policy token.
The resulting sequence of policy tokens is processed by a transformer encoder
to obtain $E_{\rm act}(\pi_\theta)$.

In our current implementation, the policy tokens use activations from the
final two layers of the actor network, providing a lightweight representation
of the policy while avoiding direct encoding of all actor parameters.
The evaluation states $\{\zeta_k\}$ are themselves trainable and are optimized
end-to-end through the functional-critic TD loss in
Eq.~\eqref{eq:appendix-func-q-loss}.

This construction may be viewed as learning a set of evaluation functionals
$\delta_{\zeta_k}$ over the policy space, where
$\delta_{\zeta_k}(\pi)=\pi(\zeta_k)$, augmented with internal actor
representations. It provides the functional critic with a policy-dependent
representation without requiring an explicit vectorization of the full actor
network.

\subsection{Ensemble Functional Actor-Critic for Posterior-Sampling Exploration}
\label{appendix:ensemble-psrl}

We use an ensemble of deterministic actors
$\mathcal A_t=\{\pi_{\theta_t^{(i)}}\}_{i=1}^n$.
The actor ensemble plays two roles.
First, it supplies the policy training set on which each functional critic is
fit.
Second, it provides multiple candidate policies that can be improved under
different sampled critic particles.

Importantly, actors and functional critics are \emph{not} permanently paired.
At every actor update, actor $i$ independently samples an index
\begin{equation}
c_t^{(i)}
\sim
\mathrm{Unif}(\{1,\ldots,n\}),
\end{equation}
and is updated using the functional policy gradient induced by
$\hat Q^{(c_t^{(i)})}$:
\begin{equation}
\theta_{t+1}^{(i)}
=
\theta_t^{(i)}
+
\eta_t
\nabla_\theta
J_\rho
\left(
\pi_\theta;
\xi^{(c_t^{(i)})}
\right)
\bigg|_{\theta=\theta_t^{(i)}}.
\label{eq:appendix-sampled-critic-update}
\end{equation}

Because every functional critic is defined over the policy space, any critic
particle can be used to improve any actor.
Resampling a critic for every actor update implements the particle-based
posterior-sampling mechanism described in Section~\ref{sec-psrl}.

Algorithm~\ref{alg-ensemble-algorithm} instantiates the
posterior-sampling-style construction described in
Section~\ref{sec-psrl}.
We maintain $n$ actors and $n$ functional critics.
Each critic is trained across the current actor ensemble, while actor updates
independently sample critics from the ensemble and use their corresponding
functional policy gradients.

\begin{algorithm}[th]
\caption{Ensemble Functional Actor--Critic}
\label{alg-ensemble-algorithm}
\begin{algorithmic}[1]
\STATE \textbf{Input:}
$\{\theta_0^{(i)}\}_{i=1}^n$,
$\{\xi_0^{(j)}\}_{j=1}^n$,
$T$, $\{\eta_t\}_{t=1}^T$

\FOR{$t=1,\ldots,T$}
    \STATE Collect/sample off-policy data $\mathcal D_t$
    \STATE
    $\mathcal A_{t-1}
    \gets
    \{\pi_{\theta_{t-1}^{(i)}}\}_{i=1}^n$

    \COMMENT{Update functional-critic ensemble}
    \FOR{$j=1,\ldots,n$}
        \STATE
        $\xi_t^{(j)}
        \gets
        \textsc{FunctionalPolicyEvaluator}
        (\xi_{t-1}^{(j)},\mathcal A_{t-1},\mathcal D_t)$
    \ENDFOR

    \COMMENT{Update actors with independently sampled critics}
    \FOR{$i=1,\ldots,n$}
        \STATE Sample $c_t^{(i)} \sim \mathrm{Unif}(\{1,\ldots,n\})$
        \STATE
        $\theta_t^{(i)}
        \gets
        \theta_{t-1}^{(i)}
        +
        \eta_t
        \nabla_\theta
        J_\rho(\pi_\theta;\xi_t^{(c_t^{(i)})})
        \big|_{\theta=\theta_{t-1}^{(i)}}$
    \ENDFOR
\ENDFOR

\STATE \textbf{Return} $\{\theta_T^{(i)}\}_{i=1}^n$
\end{algorithmic}
\end{algorithm}
For the deterministic actors used in our experiments, the actor update in
Algorithm~\ref{alg-ensemble-algorithm} uses the full deterministic functional
policy gradient in Eq.~\eqref{eq:det-functional-gradient}, including both the
action-gradient pathway and the derivative through the explicit policy input.

\subsection{Training Algorithm}

Algorithm~\ref{alg-nn} summarizes the complete neural functional
actor--critic training procedure.
Each functional critic is trained across the current actor ensemble, while
each actor independently samples a functional critic at every actor update.

\begin{algorithm}[h!]
\caption{Neural Ensemble Functional Actor--Critic}
\label{alg-nn}
\begin{algorithmic}[1]
\STATE \textbf{Input:}
ensemble size $n$,
critic-to-actor update ratio $G$,
actor learning rate $\eta$,
target-update coefficient $\rho$

\STATE Initialize actor parameters
$\{\theta_0^{(i)}\}_{i=1}^n$
\STATE Initialize functional critics
$\{\Xi_0^{(i)}\}_{i=1}^n$
\STATE Initialize target critic parameters
$\{(\xi_{\rm sa}^{'(i)},\xi_{\rm joint}^{'(i)})\}_{i=1}^n$
\STATE Initialize replay buffer $\mathcal R$

\FOR{$t=0,1,\ldots$}

    \IF{starting a new episode}
        \STATE Sample rollout actor
        $r\sim\mathrm{Unif}(\{1,\ldots,n\})$
    \ENDIF

    \STATE Execute
    $a_t=\pi_{\theta_t^{(r)}}(s_t)$
    and store
    $(s_t,a_t,r_t,s_{t+1})$ in $\mathcal R$

    \STATE
    $\mathcal A_t
    \gets
    \{\pi_{\theta_t^{(i)}}\}_{i=1}^n$

    \COMMENT{Update functional-critic ensemble}
    \FOR{$g=1,\ldots,G$}
        \STATE Sample transition batch $B_C\sim\mathcal R$

        \FOR{$j=1,\ldots,n$}
            \STATE Compute TD targets
            Eq.~\ref{eq:appendix-td-target}
            on $\mathcal A_t\times B_C$

            \STATE Update $\Xi^{(j)}$ by minimizing
            Eq.~\ref{eq:appendix-func-q-loss}

            \STATE Soft-update target critic:
            \[
            \xi_{\rm sa}^{'(j)}
            \leftarrow
            \rho\xi_{\rm sa}^{'(j)}
            +(1-\rho)\xi_{\rm sa}^{(j)}
            \]
            \[
            \xi_{\rm joint}^{'(j)}
            \leftarrow
            \rho\xi_{\rm joint}^{'(j)}
            +(1-\rho)\xi_{\rm joint}^{(j)}
            \]
        \ENDFOR
    \ENDFOR

    \STATE Sample actor-update batch $B_A\sim\mathcal R$

    \COMMENT{Update actors with independently sampled critics}
    \FOR{$i=1,\ldots,n$}
        \STATE Sample
        $c^{(i)}
        \sim
        \mathrm{Unif}(\{1,\ldots,n\})$

        \STATE Evaluate the deterministic functional policy gradient
        in Eq.~\ref{eq:det-functional-gradient}
        using $\hat Q^{(c^{(i)})}$ and $B_A$

        \STATE
        $\theta_{t+1}^{(i)}
        \gets
        \theta_t^{(i)}
        +
        \eta
        \hat\nabla_\theta J_\rho
        (\pi_{\theta_t^{(i)}};\hat Q^{(c^{(i)})})$
    \ENDFOR

\ENDFOR
\end{algorithmic}
\end{algorithm}

In all main experiments, we use $n=10$ functional critics and a
critic-to-actor update ratio of $G=10$.
Additional hyperparameters are reported in
Table~\ref{tab:hyperparams}.

\subsection{Omitted Off-Policy AC Heuristics}

The minimalist implementation deliberately excludes several techniques widely
used in modern off-policy actor--critic implementations.

For critic learning, we do not use clipped double-Q or minimum-target value
estimation \citep{fujimoto2018addressing,haarnoja2018soft}.
For actor learning and data collection, the actors are deterministic and we
do not inject exploration noise.
We further omit target-policy smoothing
\citep{fujimoto2018addressing} and entropy regularization or automatic entropy
tuning \citep{haarnoja2018soft}.

These choices are not intended to argue that the corresponding heuristics are
generally unnecessary. Rather, they isolate a deliberately simple
implementation that remains close to the structural ingredients analyzed in
the paper: policy-conditioned critic generalization and functional-ensemble
exploration.

\section{Experimental Details}
\label{appendix:hp}

\subsection{Benchmarks and Evaluation Protocol}

We evaluate the complete Dog and Humanoid task series used in our benchmark
from the DeepMind Control Suite
\citep{tunyasuvunakool2020dmc}:
Humanoid Stand, Humanoid Walk, Humanoid Run, Dog Stand, Dog Walk, Dog Trot,
Dog Run, and Dog Fetch.
These task families provide two distinct morphologies with relatively
high-dimensional state and action spaces.

For each task, experiments are run with four parallel environments and four
independent random seeds.
Figure~\ref{fig:main_experiment} reports mean episodic return as a function of
environment steps, with shaded regions denoting one standard deviation.
We evaluate the full set of tasks in these two series rather than selecting
individual tasks.

\subsection{Baseline Construction}

We compare against the original SAC and TD3 algorithms
\citep{haarnoja2018soft,fujimoto2018addressing}, as well as strengthened
variants SAC+ and TD3+ implemented using the extensively tested ALF codebase
\citep{Xu2021ALF}.

The vanilla SAC and TD3 baselines retain their original two-critic
architectures and critic-to-actor update ratios of 1:1 and 2:1,
respectively.
SAC+ and TD3+ use 10 critics and, after independent tuning, a 10:1
critic-to-actor update ratio.
Our method likewise uses 10 functional critics and a 10:1 ratio in all main
experiments.

Our method also uses 10 functional critics and the same 10:1 critic-to-actor
update ratio in all main experiments.
Thus, the main comparison against SAC+ and TD3+ matches both the critic
ensemble size and the frequency of critic updates, while retaining
method-specific architectural components.
SAC+ retains the stochastic Gaussian actor and entropy regularization of SAC,
without introducing additional exploration mechanisms beyond those in the
original algorithm.
TD3+ retains the deterministic actor of TD3, together with its additive
rollout exploration noise and target-policy smoothing noise. 
Our method uses the deterministic functional-actor ensemble described in
Appendix~\ref{appendix:deep-algorithm}.

\subsection{Prior Policy-Conditioned Value Functions}

We did not include earlier policy-conditioned value-function (PCVF) methods as
primary baselines.
A comprehensive benchmark by \citet{griesbach2023improving} found the major
PCVF variants to be noncompetitive with standard RL algorithms such as TD3,
consistent with our own preliminary attempts.
Our experiments therefore focus on the substantially stronger SAC+ and TD3+
baselines; we refer readers to \citet{griesbach2023improving} for a detailed
comparison among earlier PCVF approaches.

\begin{table}[!h]
\centering
\caption{Hyperparameters of our functional actor--critic implementation,
shared across all reported tasks.}
\label{tab:hyperparams}
\begin{tabular}{l c}
\toprule
\textbf{Hyperparameter} & \textbf{Value} \\
\midrule
optimizer & Adam \\
learning rate & $3 \times 10^{-4}$ \\
batch size & 256 \\
discount factor ($\gamma$) & 0.99 \\
target update rate ($\tau$) & 0.005 \\
critic ensemble size & 10 \\
critic-to-actor update ratio & 10 \\
actor network hidden layers & $(256,256)$ \\
critic state--action encoder hidden layers & $(256,256)$ \\
critic joint encoder hidden layers & $(256,256)$ \\
actor transformer encoder layers & 4 \\
actor transformer encoder heads & 1 \\
trainable evaluation states & 512 \\
actor layers exposed to actor encoder & last two \\
actor encoding dimension & 128 \\
state--action encoding dimension & 64 \\
activation function & ReLU \\
\bottomrule
\end{tabular}
\end{table}

\begin{table}[!h]
\centering
\caption{Key algorithmic differences among the compared methods.}
\label{tab:baseline-config}
\small
\begin{tabular}{lccccc}
\toprule
 & Ours & SAC+ & TD3+ & SAC & TD3 \\
\midrule
Number of critics
    & 10 & 10 & 10 & 2 & 2 \\

Critic-to-actor update ratio
    & 10:1 & 10:1 & 10:1 & 1:1 & 2:1 \\

Layer normalization
    & Yes & Yes & Yes & No & No \\
    
Actor policy
    & Det. & Stoch. & Det. & Stoch. & Det. \\

Additional rollout action noise
    & No & No & Yes & No & Yes \\

Target-policy smoothing
    & No & No & Yes & No & Yes \\

Entropy regularization
    & No & Yes & No & Yes & No \\
\bottomrule
\end{tabular}
\end{table}

\subsection{Hyperparameters and Baseline Configurations}

Hyperparameters of our method used across the experiments reported in
Fig.~\ref{fig:main_experiment} are summarized in
Table~\ref{tab:hyperparams}.
Key algorithmic differences among the compared methods are summarized in
Table~\ref{tab:baseline-config}.

\section{Heuristic Solution to Replay Sampling}
\label{appendix:replay-experiments}

In this section, we develop a heuristic neural instantiation of the replay
sampling mechanism proposed in Section~\ref{sec:replay-budget}.
Starting from the base neural ensemble functional actor--critic in
Algorithm~\ref{alg-nn}, we modify only the critic replay update: instead of
using uniform replay weights, we compute a feature-based sampling distribution
over the replay minibatch and use it to form a weighted TD objective.
Because the linear features required by the theory are not directly available
in the neural implementation, we approximate them using frozen critic
representations and evaluate the resulting modification on Dog\_run and
Humanoid\_run.


\subsection{Feature Covariance with Frozen Neural Features}\label{sec:neural_feature}

While our convergence guarantees for Algorithm \ref{alg-linear-AC} are based on the linear functional representation \textbf{Assumption~\ref{assumption-linear}}, practical implementations typically adopt a nonlinear neural parameterization of the value function, such as Section \ref{sec-experiment}. Note that this does not contradict with \textbf{Assumption~\ref{assumption-linear}}, because the latter only assumes the existence of a linear representation in hindsight, which needs not be the neural value function we use in practice. Nonetheless, this does bring difficulty as we do not know the linear feature representation $\phi^\pi$ and thereby cannot directly approximate the feature covariance with Monte Carlo methods. 

To address this issue, we propose freezing the reference features at each update of the critic network $\hat{Q}^{(i)}(\pi,s,a)$ in the critic ensemble. To retain the linear structure in \textbf{Assumption~\ref{assumption-linear}}, we take the penultimate activation as the linear feature $\hat{\phi}^{(i),\pi}$ and use them to approximate the population covariance $\hat{\bA}_\pi$ by a sample average, where the expectation in \eqref{eq: A-def} over $d^\pi$ is replaced by state-action samples from a small recent buffer. The weighted covariance matrices over the buffer in \eqref{eq:replay-design} also use the frozen reference features above; but to reduce the computational burden, we first uniformly draw a minibatch $n_{\textnormal{mini}} < n$ from the entire size-$n$ buffer $\mathcal{B}$ and then assemble the matrices in \eqref{eq:replay-design} over the minibatch and solve the distribution design $\hat{\mathbf w}$ in \eqref{eq:replay-budget-program} over the minibatch. 

\subsection{Approximately Solving the Nonconvex Program}\label{sec:approximate_nonconvex_replay_program}

It remains to solve the generally nonconvex optimization program in \eqref{eq:replay-budget-program} for the distribution design $\hat{\mathbf w}$ over the drawn minibatch $\mathcal{B}_{\textnormal{mini}}$. For an approximate solution, we parameterize $\hat{\mathbf w}$ with a softmax representation
\[
\hat{\mathbf w}_i = \frac{\min\{p_i,p_{\max}\}}{\sum_{j\in \mathcal{B}_{\textnormal{mini}}}\min\{p_j,p_{\max}\}},\qquad p_i = \mathrm{softmax}(\mathbf z)_i
\]
where $p_{\max}\in(0,1]$ is a hyperparameter ceiling and $\mathbf{z}\in\mathbb{R}^{n_{\textnormal{mini}}}$ is the trainable parameter. We initialize $\mathbf{z}$ to the zero vector so $\hat{\mathbf w}$ is initially uniform, and then we apply $K_{\mathrm{lbfgs}}$ steps of the L-BFGS optimizer to optimize the objective in \eqref{eq:replay-budget-program} over $\mathbf{z}$ and arrive at the final distribution design $\hat{\mathbf w}$. 

\paragraph{Modification to Algorithm~\ref{alg-nn}.}
All components of Algorithm~\ref{alg-nn} are kept unchanged except for the
critic replay update.
For each critic update, we first draw a candidate minibatch
$\mathcal B_{\rm mini}$ from the replay buffer, compute the heuristic replay
weights $\hat{\mathbf w}$ described above, and replace the uniformly averaged
critic TD loss by its $\hat{\mathbf w}$-weighted counterpart.
We refer to this variant as \texttt{Ours\_reweight}.


\begin{figure*}[tb]
    \centering
    \vspace{-0.1in}
    \begin{subfigure}[t]{0.49\textwidth}
        \centering
        \includegraphics[width=\linewidth]{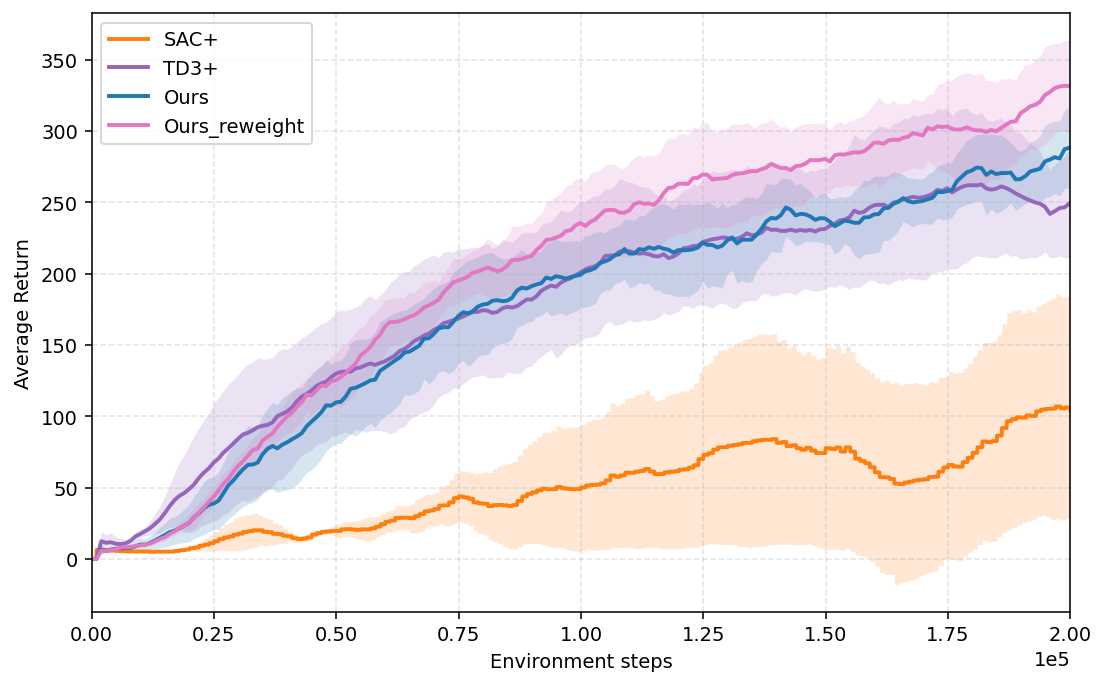}
        \caption{Dog\_run}
        \label{fig:reweight_dog_run}
    \end{subfigure}
    \hfill
    \begin{subfigure}[t]{0.49\textwidth}
        \centering
        \includegraphics[width=\linewidth]{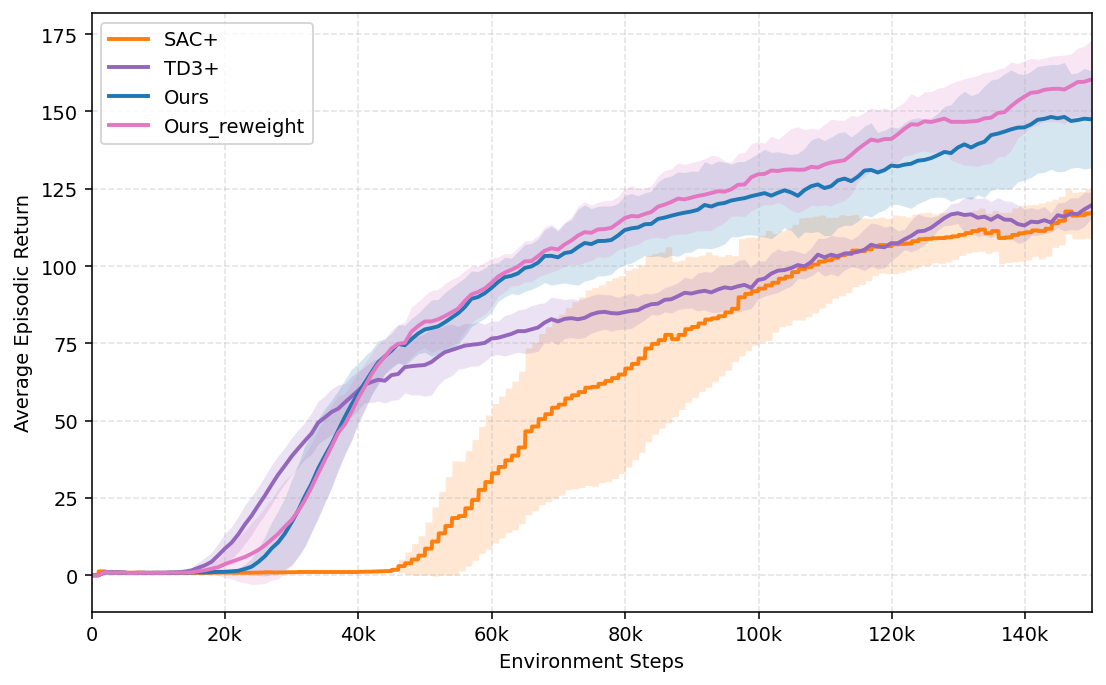}
        \caption{Humanoid\_run}
        \label{fig:reweight_hum_run}
    \end{subfigure}

    \caption{Targeted evaluation of the replay-reweighted variant
    \texttt{Ours\_reweight} on Dog\_run and Humanoid\_run.
    Starting from Algorithm~\ref{alg-nn}, we modify only the critic replay update:
    for each candidate replay minibatch, we compute the heuristic distribution
    $\hat{\mathbf w}$ described above and use it to weight the squared TD loss in
    Eq.~\eqref{eq:appendix-func-q-loss}.
    We use $p_{\max}=0.1$ and $K_{\mathrm{lbfgs}}=1$.
    On Dog\_run, replay reweighting is enabled from the beginning for all four
    runs and improves environment-sample efficiency over the original algorithm.
    On Humanoid\_run, early activation was not beneficial, motivating delayed
    activation after the initial feature-learning phase; because activation timing
    is not identical across all runs, this panel is intended as a diagnostic of
    feature-dependent activation rather than a fixed-protocol four-seed
    comparison.}

    \label{fig:reweighting_experiment}
    \vspace{-0.1in}
\end{figure*}

\paragraph{Delayed Replay Sampling.}
Another practical distinction from the convergence analysis in Section \ref{sec-linear-functional} is that the feature representation $\phi^\pi$ is not known and needs to be learned over time. Recall that we use the penultimate activation of our neural functional critics as the approximate feature representation $\hat{\phi}^{(i),\pi}$. It is therefore possible that a feature-based module like Replay Sampling can be unhelpful or even harmful in the early stage, when the functional critics have not trained sufficiently and the learned features are misrepresented. Indeed, in the Humanoid\_run task in Figure \ref{fig:reweighting_experiment}, we found that enabling Replay Sampling from the beginning is not helpful compared to \texttt{Ours}. Thus in Figure \ref{fig:reweight_hum_run}, we enable Replay Sampling only after 90k environment steps for three of the four independent runs. Figure \ref{fig:reweight_dog_run} enables Replay Sampling from the beginning for all four independent runs. 
These observations suggest that feature-based replay control becomes useful
only after the learned neural representation is sufficiently informative.
We therefore view the Humanoid\_run experiment as a diagnostic of this
feature-maturation effect rather than as a controlled comparison under a
single activation schedule. Automatically determining when to activate the
feature-based replay mechanism remains an open practical question.

\section{Heuristic Neural Instantiation of Trust Coverage}
\label{appendix:tr-experiments}

This section develops a heuristic neural instantiation of the
trust-coverage-based behavior-policy refresh mechanism in
Algorithm~\ref{alg-linear-AC}.
Specifically, we integrate the evaluation-trust criterion
$\mathcal C(\pi,\mu)$ from Algorithm~\ref{alg-linear-AC} into the neural
ensemble functional actor--critic of Algorithm~\ref{alg-nn}.
This experiment instantiates only the evaluation-trust component of
Algorithm~\ref{alg-linear-AC}; the gradient-generalization criterion
$\Delta_{k,t}$ governing reuse of the evaluated critic across actor updates
is not implemented here.
The actor and critic updates otherwise follow Algorithm~\ref{alg-nn}; the
modification concerns only whether a new behavior-policy rollout is collected
or the existing behavior data are reused.

Because Algorithm~\ref{alg-linear-AC} is formulated under linear functional
approximation whereas Algorithm~\ref{alg-nn} uses neural functional critics,
implementing the trust criterion requires neural approximations of the
population feature covariances $\mathbf A_\pi$ and
$\widetilde{\mathbf A}_\pi$, and hence of
$\mathcal C(\pi,\mu)$ in Eq.~\eqref{eq:eval_metric}.

As in Appendix~\ref{sec:neural_feature}, we use the penultimate activations
of frozen functional critics as approximations to the linear feature
representation in Assumption~\ref{assumption-linear}.
For the expectation over the behavior-policy stationary distribution
$d^\mu$ in $\widetilde{\mathbf A}_\pi$, we use the sample average over the
large replay buffer $\mathcal B$.
The expectations over the target-policy stationary distribution $d^\pi$ in
$\mathbf A_\pi$ and Eq.~\eqref{eq:eval_metric} are approximated using a much
smaller recent buffer $\mathcal B_{\rm recent}$.

Our goal here is not a full benchmark of the trust-coverage mechanism, but a
targeted study of whether the learned trust signal can identify periods in
which behavior-policy refreshes may be safely skipped and data reused.

\paragraph{Trust-Guided Behavior Reuse.}
Algorithm~\ref{alg-linear-AC} refreshes the behavior policy when the
evaluation-trust criterion indicates insufficient coverage of the current
reference policy.
We transfer this mechanism to Algorithm~\ref{alg-nn} by evaluating a neural
approximation of $\mathcal C(\pi,\mu)$ before collecting a new rollout.
When the estimated trust metric remains below the prescribed threshold, the
new rollout is skipped and the existing behavior data are reused; otherwise,
a fresh rollout is collected as in the base neural algorithm.
We refer to this variant as \texttt{Ours\_trust}.

Because the trust criterion is computed from learned neural features rather
than the unknown linear features in
Assumption~\ref{assumption-linear}, its reliability can depend on the quality
of the learned representation.
In particular, activating trust-based refresh skipping too early, when the
critic features are still poorly learned, can be unhelpful or even harmful.
We therefore study delayed activation of this mechanism in the experiment
below.

\begin{figure*}[tb]
    \centering
    \vspace{-0.1in}
    \begin{subfigure}[t]{0.49\textwidth}
        \centering
        \includegraphics[width=\linewidth]{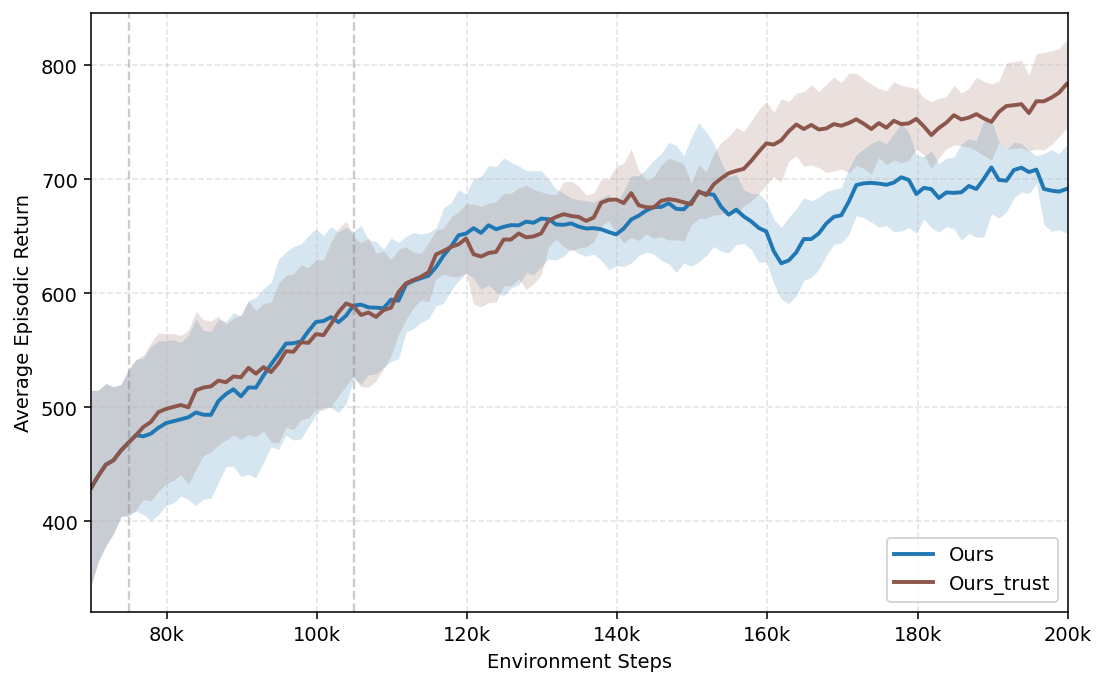}
        \caption{Dog\_walk return curve}
        \label{fig:tr_dog_return}
    \end{subfigure}
    \hfill
    \begin{subfigure}[t]{0.49\textwidth}
        \centering
        \includegraphics[width=\linewidth]{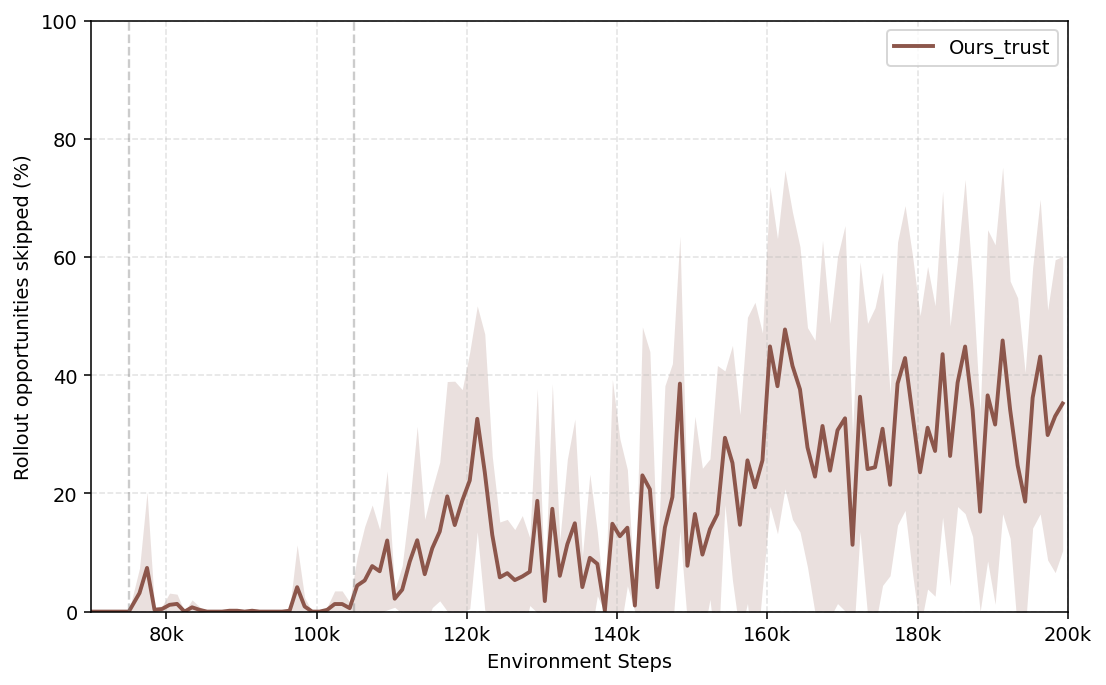}
        \caption{Skip rate of behavior policy refresh based on trust coverage}
        \label{fig:tr_dog_skip}
    \end{subfigure}

    \caption{
    Targeted evaluation of the trust-guided behavior-reuse variant
    \texttt{Ours\_trust} on Dog\_walk.
    The variant combines the neural ensemble functional actor--critic in
    Algorithm~\ref{alg-nn} with the evaluation-trust behavior-refresh criterion
    from Algorithm~\ref{alg-linear-AC}.
    Before collecting a fresh rollout, we compute the neural approximation of
    $\mathcal C(\pi,\mu)$ described above and skip the rollout when the estimate
    remains below the chosen threshold.
    To study delayed activation of the learned feature-based criterion, two runs
    enable refresh skipping at 75k environment steps and two at 105k environment
    steps (dashed lines).
    The threshold for each run is chosen as the $33$-percentile of empirically
    estimated $\mathcal C(\pi,\mu)$ values over random subsets of the replay
    buffer.
    The trust-based variant improves environment-sample efficiency in this
    targeted study while skipping a substantial fraction of behavior-policy
    refreshes.
    The improvement coincides with periods of frequent refresh skipping,
    providing initial evidence that the evaluation-trust criterion can serve as a
    practical signal for data reuse.
    }
    \label{fig:tr_experiment}
    \vspace{-0.1in}
\end{figure*}

\section{Proof for Section~\ref{sec-linear-functional}}
\label{appendix:sec-linear-functional}

\subsection{Discussion of the Linear Functional Approximation Assumptions}
\label{appendix-assumption-linear}

\textbf{Assumption~\ref{assumption-mixing}} is a standard ergodicity condition for
stochastic approximation with Markovian noise
\citep{zhang2020provably,zhang2021breaking,chen2023target}.
Some prior works impose slightly weaker conditions; we assume irreducibility
and aperiodicity for simplicity and to ensure a unique, well-behaved
stationary distribution.

\textbf{Assumption~\ref{assumption-linear}} generalizes the standard linear
realizability assumption by allowing the feature representation
$\phi^\pi(s,a)$ to depend explicitly on the policy.
Thus, rather than representing the value function of a single fixed policy,
the functional features jointly represent the mapping from the policy class
$\Pi$ to the corresponding value functions.
Geometrically, this representation can be interpreted locally through a
first-order expansion over the policy manifold, with policy-dependent
features capturing variations of the value function along its tangent
directions.

We consider the policy space $\Pi$ as a smooth Riemannian manifold. For any fixed state-action pair $(s,a)$, the value function can be understood as a functional map $\mathcal{Q}_{s,a}: \Pi \to \mathbb{R}$. While standard linear function approximation focuses on the representability of a single policy, Assumption~\ref{assumption-linear} leverages the intrinsic local continuity and smoothness of the policy manifold $\Pi$.



For any anchored reference policy $\pi_{0} \in \Pi$, let $T_{\pi_{0}}\Pi$ denote the tangent space at $\pi_{0}$, with the associated exponential map $\exp_{\pi_{0}}(\cdot): T_{\pi_{0}}\Pi \to \Pi$ and its inverse, the logarithmic map $\log_{\pi_{0}}(\cdot): \Pi \to T_{\pi_0}\Pi$. For any policy $\pi$ in a sufficiently small neighborhood of $\pi_{0}$, a first-order Taylor expansion on the manifold yields:
\begin{align*}
Q^{\pi}(s,a) = \mathcal{Q}_{s,a}(\exp_{\pi_{0}}(\log_{\pi_{0}}(\pi))) \approx Q^{\pi_{0}}(s,a) + \langle \nabla_{\pi} \mathcal{Q}_{s,a}(\pi_{0}), \log_{\pi_{0}}(\pi) \rangle_{T_{\pi_{0}}\Pi},
\end{align*}
where $\nabla_{\pi} \mathcal{Q}_{s,a}(\pi_{0})$ represents the functional gradient of the value with respect to the anchor policy.

By identifying a set of basis functions $\Phi(s,a)$ that span the directional derivatives of the value functional, the displacement vector $\log_{\pi_{0}}(\pi) \in T_{\pi_0}\Pi$ interacts linearly with the state-action features. Specifically, we can define the \textbf{functional feature map} as the tensor product $\phi^{\pi}(s,a) := \text{vec}(\Phi(s,a) \otimes \log_{\pi_{0}}(\pi))$. Under this coordinate-free construction, the value function locally satisfies:
\begin{equation}\label{eq-shared-linear-representation}
Q^{\pi}(s,a) \approx \phi^{\pi}(s,a)^\top w_{\pi_{0}}.
\end{equation}

In this light, Assumption~\ref{assumption-linear} holds locally for any $\pi$ close to $\pi_{0}$, with the underlying representation depending on the choice of the anchor policy. Our dual-layer mechanism is precisely designed to respect the boundaries of this local validity: while the first layer ($\mathcal{C}(\pi, \mu)$) ensures the data is sufficient to build the anchored critic, the second layer ($\Delta_{k,t}$) monitors the displacement on the tangent space. Crucially, the \textbf{Gradient Generalization Trust Metric} $\Delta_{k,t}$ serves as a data-aware proxy for the Riemannian distance $d(\pi, \pi_0)$ on the manifold. By enforcing $\Delta_{k,t} \leq 2$, Algorithm~\ref{alg-linear-AC} refreshes the anchor policy exactly when the policy $\pi$ drifts beyond the "trust region" where the manifold's first-order approximation—and thus the functional critic's gradient—remains reliable.

\textbf{Assumption~\ref{assumption-closeness}} similarly extends approximate Bellman
completeness to the functional setting.
Unlike the Linear MDP assumption used in some prior work
\citep{xiong2022nearly,yin2022near,min2021variance}, it does not require the
transition dynamics themselves to admit a linear representation in the
features.

\subsection{Additional Details for Target-Based Evaluation with Partial Coverage}
\label{appendix:partial-coverage-evaluation}

We provide additional details for the target-based policy evaluator analyzed in
Section~\ref{sec-linear-functional}, together with further discussion of the
partial-coverage guarantee.

\paragraph{Target-based policy evaluator.}
For a target policy $\pi$ and behavior policy $\mu$, we use the following
target-network TD procedure.

\begin{algorithm}[t]
\caption{Target-Network Based Policy Evaluation}
\label{alg:target-based-evaluation}
\begin{algorithmic}[1]
\STATE \textbf{Input:} behavior policy $\mu$; target policy $\pi$;
feature map $\phi^\pi$; accuracy level $\epsilon$;
inner iterations $K$; outer iterations $T$
\STATE \textbf{Initialize:} $\hat\xi_0\gets \bm 0$
\FOR{$t=0,1,\ldots,T-1$}
    \STATE Initialize $\xi_{t,0}\gets \bm 0$
    \FOR{$k=0,1,\ldots,K-1$}
        \STATE Sample $(s_k,a_k,r_k,s_k')$ from $\mu$
        \STATE
        $y_k\gets
        r_k+
        \gamma\sum_a\pi(a|s_k')
        \left\lceil
        \phi^\pi(s_k',a)^\top\hat\xi_t
        \right\rceil$
        \STATE
        $\xi_{t,k+1}
        \gets
        \xi_{t,k}
        +
        \alpha\phi^\pi(s_k,a_k)
        \left(
        y_k-\phi^\pi(s_k,a_k)^\top\xi_{t,k}
        \right)$
    \ENDFOR
    \STATE $\hat\xi_{t+1}\gets\xi_{t,K}$
\ENDFOR
\STATE \textbf{Output:} $\hat\xi_T$
\end{algorithmic}
\end{algorithm}

The auxiliary target weights $\hat\xi_t$ are held fixed within each inner
loop and refreshed after $K$ TD updates. Such target-based procedures are
well studied under linear function approximation
\citep{carvalho2020new,chen2023target,zhang2021breaking}.
Our contribution is to establish convergence without requiring pointwise
full coverage by the behavior policy.

\paragraph{Interpretation of feature-based coverage.}
Recall the on-policy and off-policy feature covariances
\[
\mathbf A_\pi
=
\E_{d^\pi}
\left[
\phi^\pi(s,a)\phi^\pi(s,a)^\top
\right],
\qquad
\widetilde{\mathbf A}_\pi
=
\E_{d^\mu}
\left[
\phi^\pi(s,a)\phi^\pi(s,a)^\top
\right].
\]
The evaluation trust metric
$\mathcal C(\pi,\mu)$ in \eqref{eq:eval_metric}
measures whether data collected under $\mu$ cover the feature directions
that matter under the target policy $\pi$.
Unlike pointwise full coverage, this condition allows
$\widetilde{\mathbf A}_\pi$ to be singular, provided the feature directions
required by $\pi$ lie in the span supported by the behavior data.
The inverse norm in \eqref{eq:eval_metric} is therefore interpreted through
the limiting regularization, equivalently the Moore--Penrose pseudoinverse
on the supported subspace.

The key analytical step behind Theorem~\ref{thm-evaluation} is to establish
Bellman contraction in the $L^2$ norm weighted by the on-policy covariance
$\mathbf A_\pi$. This choice of metric allows the analysis to ignore
uncovered directions that are irrelevant to the target policy and is what
permits the weaker feature-based coverage requirement.

\paragraph{Specialization to standard linear function approximation.}
Theorem~\ref{thm-evaluation} only requires
Assumptions~\ref{assumption-linear} and~\ref{assumption-closeness}
for the particular target policy being evaluated.
Consequently, it also applies directly to the standard linear function
approximation setting
\citep{carvalho2020new,chen2023target,zhang2021breaking}.
When $\mathcal C(\pi,\mu)<\infty$, the resulting bound has the same
$\epsilon$-dependence as least-squares policy-evaluation guarantees
\citep{min2021variance,duan2020minimax}.

\paragraph{Weight-space consequence.}
The value-space guarantee in Theorem~\ref{thm-evaluation} immediately implies
a corresponding bound on the critic parameters:
\begin{corollary}[Convergence in Weight Space]
\label{cor:weight_conv}
Under the conditions of Theorem~\ref{thm-evaluation},
\[
\E\!\left[
\|\hat\xi_T-\xi^\star\|_{\mathbf A_\pi}^2
\right]
\lesssim
\gamma^T
+
\mathcal C(\pi,\mu)
\left(
\epsilon^2+\mathcal E_{\rm approx}^2
\right)
+
\mathcal E_{\rm approx}^2 .
\]
\end{corollary}

This parameter-space form is used in the analysis of the full functional
actor-critic algorithm.

\subsection{Additional Details for Dual Trust-Coverage}
\label{appendix:dual-trust-coverage}

We provide additional details for the dual trust-coverage mechanism analyzed
in Section~\ref{sec-linear-functional}, including the derivation of the
gradient-generalization trust metric and several useful special cases.

\paragraph{Notation.}
Within epoch $k$, let
$\pi_t^{(k)}:=\pi_{\theta_t^{(k)}}$ denote the current actor and
$\bar{\pi}^{(k)}$ the reference policy at which the functional critic is
evaluated.
We write
\[
\phi_t^{(k)}(s,a)
:=
\phi^{\pi_t^{(k)}}(s,a),
\qquad
\mathbf A_k
:=
\mathbf A_{\bar{\pi}^{(k)}}.
\]
The covariance $\mathbf A_k$ remains fixed within epoch $k$ and serves as the
metric anchor for both trust criteria.

\paragraph{Reference-policy evaluation.}
By Theorem~\ref{thm-evaluation}, whenever
\[
\mathcal C(\bar{\pi}^{(k)},\mu)\le C_{\max},
\]
the critic parameter $\xi^{(k)}$ obtained by evaluating
$\bar{\pi}^{(k)}$ satisfies
\begin{equation}
\label{eq:appendix-xi-k-error}
\sqrt{
\E\!\left[
\|\xi^{(k)}-\xi^\star\|_{\mathbf A_k}^2
\right]}
\lesssim
\bar\epsilon
:=
\sqrt{C_{\max}}
\left(
\epsilon+\mathcal E_{\rm approx}
\right).
\end{equation}
Thus, the first trust criterion controls the accuracy of the functional
critic at the reference policy using off-policy experience.

\paragraph{Gradient-error decomposition.}
The functional policy-gradient estimate at the current actor
$\pi_t^{(k)}$ reuses the anchored critic $\xi^{(k)}$.
Under linear functional representation, the resulting gradient error
decomposes as
\begin{equation}
\label{eq:appendix-gradient-decomposition}
\begin{aligned}
\hat{\nabla}J_\rho(\pi_t^{(k)},\xi^{(k)})
-\nabla J_\rho(\pi_t^{(k)})
&=
\underbrace{
\E_{s\sim\rho}
\left[
\sum_{a\in\mathcal A}
\left\langle
\phi_t^{(k)}(s,a),
\xi^{(k)}-\xi^\star
\right\rangle
\nabla \pi_t^{(k)}(a|s)
\right]
}_{:=\mathcal G_1^{(k,t)}}
\\
&\quad+
\underbrace{
\E_{s\sim\rho}
\left[
\sum_{a\in\mathcal A}
\pi_t^{(k)}(a|s)
D\phi_t^{(k)}(s,a)
(\xi^{(k)}-\xi^\star)
\right]
}_{:=\mathcal G_2^{(k,t)}} .
\end{aligned}
\end{equation}

The first term captures amplification of critic error through the functional
features themselves, while the second captures amplification through their
policy derivatives.

\paragraph{Gradient estimation guarantee.}
Let $e_k:=\xi^{(k)}-\xi^\star$ and write
$\delta_{k,t}:=\hat\nabla J_\rho(\pi_t^{(k)},\xi^{(k)})
-\nabla J_\rho(\pi_t^{(k)})$.
With the feature defined in~\eqref{eq:gradient-error-operator}, we have
$\delta_{k,t}=\mathbf G_{\pi_t^{(k)}}e_k$. Since $\pi_0^{(k)}=\bar\pi^{(k)}$, Assumption~\ref{assumption:anchor-gradient} gives
$\|\delta_{k,0}\|_2^2
=e_k^\top\mathbf G_{\bar\pi^{(k)}}^\top
  \mathbf G_{\bar\pi^{(k)}}e_k
\leq\kappa_{\mathrm g}^2\|e_k\|_{\mathbf A_k}^2$.
Thus the evaluation guarantee at the start of a fixed epoch transfers to
its first gradient estimate with amplification at most
$\kappa_{\mathrm g}^2$. 

\paragraph{Gradient-generalization trust metric.}
To control these two terms using
\eqref{eq:appendix-xi-k-error}, define
\begin{equation}
\label{eq:appendix-delta}
\Delta_{k,t}
:=
\max\{
\mathcal C_1^{(k,t)},
\mathcal C_2^{(k,t)}
\},
\end{equation}
where
\begin{align}
\mathcal C_1^{(k,t)}
&:=
\left\|
\E_{s\sim\rho}
\left[
\sum_{a\in\mathcal A}
\lvert \nabla \pi_t^{(k)}(a|s)\rvert
\,
\|\phi_t^{(k)}(s,a)\|_{\mathbf A_k^{-1}}
\right]
\right\|_\infty,
\label{eq:appendix-c1}
\\
\mathcal C_2^{(k,t)}
&:=
\E_{s\sim\rho}
\left[
\sum_{a\in\mathcal A}
\pi_t^{(k)}(a|s)
\,
\|D\phi_t^{(k)}(s,a)\|_{\mathbf A_k^{-1},\infty}
\right].
\label{eq:appendix-c2}
\end{align}

By Cauchy--Schwarz,
\[
\|\mathcal G_1^{(k,t)}\|_\infty
\le
\mathcal C_1^{(k,t)}
\|\xi^{(k)}-\xi^\star\|_{\mathbf A_k},
\qquad
\|\mathcal G_2^{(k,t)}\|_\infty
\le
\mathcal C_2^{(k,t)}
\|\xi^{(k)}-\xi^\star\|_{\mathbf A_k}.
\]
Combining these inequalities with
\eqref{eq:appendix-xi-k-error} gives
\begin{equation}
\label{eq:appendix-gradient-generalization}
\E\!\left[
\|
\nabla J_\rho(\pi_t^{(k)})
-
\hat{\nabla}J_\rho(\pi_t^{(k)},\xi^{(k)})
\|_\infty
\right]
\lesssim
\Delta_{k,t}\bar\epsilon .
\end{equation}
Hence, $\Delta_{k,t}$ quantifies how reliably the anchored functional critic
generalizes from $\bar{\pi}^{(k)}$ to the evolving target policy
$\pi_t^{(k)}$.

\paragraph{Smoothness.}
We use the following standard regularity condition to guarantee that the
trust metric remains controlled for sufficiently small policy updates.
\begin{assumption}[Smoothness]
\label{assumption:smoothness}
There exists $L>0$ such that for any
$\theta,\theta'\in\Theta$ and all $(s,a)\in\mathcal S\times\mathcal A$,
\begin{align*}
\|
\phi^{\pi_\theta}(s,a)
-
\phi^{\pi_{\theta'}}(s,a)
\|_2
&\le
L\|\theta-\theta'\|_2,
\quad
\|
D\phi^{\pi_\theta}(s,a)
-
D\phi^{\pi_{\theta'}}(s,a)
\|_2
\le
L\|\theta-\theta'\|_2,\\
|
\pi_\theta(a\lvert s)
-
{\pi_{\theta'}}(a\lvert s)
|
&\le
L\|\theta-\theta'\|_2,
\quad\|
\nabla_\theta \pi_\theta(a\lvert s)
-
\nabla_\theta{\pi_{\theta'}}(a\lvert s)
\|_2
\le
L\|\theta-\theta'\|_2.
\end{align*}
\end{assumption}

This is the standard continuity requirement used to control policy drift in
actor-critic convergence analyses
\citep{zhang2020provably,zhang2021breaking,barakat2022analysis}.

\paragraph{Interpretation of the two trust layers.}
The two criteria control distinct sources of policy-gradient error.
The threshold $C_{\max}$ controls the reference-policy evaluation error by
requiring sufficient feature coverage from the behavior data.
The threshold on $\Delta_{k,t}$ controls the additional
gradient-generalization error incurred as the actor evolves away from the
reference policy.
Thus, the algorithm refreshes the behavior policy only when off-policy
experience no longer provides sufficient coverage, and refreshes the
functional critic only when its gradient estimate no longer generalizes
reliably to the current actor.

This separation is also what enables sample-efficient reuse:
behavior-policy data may be reused across multiple reference-policy
evaluations as long as the first trust criterion holds, while a single
functional critic may be reused across multiple actor updates as long as the
second criterion holds.

\begin{remark}[Similarity-Based Generalization]
\label{remark:similarity}
The benefit of critic reuse depends on the functional critic generalizing
across nearby policies.
In the limiting case where the effective epoch length is one, so that the
critic is refreshed after every actor update, the framework reduces to a
standard two-timescale actor-critic scheme.
Functional generalization therefore extends the classical setting by
permitting multiple policy-improvement steps per critic evaluation whenever
policy similarity can be exploited.
\end{remark}

\begin{remark}[Behavior-Policy Special Cases]
\label{remark:selection}
Several common behavior-policy choices arise as special cases.
Setting
$\mu=\bar{\pi}^{(k)}$
ensures coverage by construction and recovers an on-policy actor-critic
regime, while a fixed behavior policy satisfying conventional full-coverage
conditions yields a standard off-policy regime.
The dual trust-coverage mechanism additionally permits intermediate settings
in which $\mu$ is reused until its feature coverage becomes insufficient.
\end{remark}

\subsection{A Finite-time Bound for Markovian SAA under (Possibly) Degenerate Norm}

In this section, we provide a detailed proof of a finite-time contraction bound for Markovian stochastic approximation under a possibly degenerate design, which serves as a foundation for our proof of Theorem~\ref{thm-evaluation} in the next section. The proof follows a Lyapunov drift argument as in \citet{chen2022finite} while carefully tracking the semi-norm induced by a positive semi-definite matrix. 

\paragraph{Notations.}
For any positive semi-definite matrix $\bA \succeq \mathbf{0}$ and vectors $\bu,\bv$ with the same dimension, we define
$\langle \bu,\bv\rangle_{\bA}:=\bu^\top \bA \bv$ and $\|\bv\|_{\bA}^2:=\bv^\top \bA \bv$.
Let $\lambda_{\max}(\bA)$ denote the largest eigenvalue of $\bA$.
When $\bA$ is singular, let $\lambda_{\min}^{+}(\bA)$ denote the smallest \emph{strictly positive} eigenvalue of $\bA$ (and W.L.O.G. $\lambda_{\min}^+(\bA) = 0$ when $\bA=\mathbf{0}$).
Let $\bPi_{\bA}$ be the orthogonal projection operator onto $\mathrm{Range}(\bA)$ and $\bPi_{\bA}^{\perp}$ the orthogonal projection operator onto $\ker(\bA)$, so that $\bPi_{\bA}+\bPi_{\bA}^\perp=\bI$.
For a finite-state Markov chain $\{\by_k\}_{k\ge0}$ over $\cY$ with transition kernel $P$ and stationary distribution $\nu$, we define the $\beta$-mixing time as
$$
\tau_\beta := \min\big\{k\ge 0:\max_{y\in\cY}\|P^k(y,\cdot)-\nu(\cdot)\|_{\mathrm{TV}}\le \beta\big\}.
$$

\begin{theorem}\label{thm: stochastic-approx-degenerate}
Consider the Markovian stochastic approximation recursion with arbitrary initialization $\bx_0\in\RR^d$ and
\begin{equation}\label{eq: SA-update-appA}
\bx_{k+1}=\bx_k+\alpha\,\bF(\bx_k,\by_k).
\end{equation}
Let $\bar{\bF}(\bx):=\EE_{\by\sim\nu}[\bF(\bx,\by)]$. Suppose that for some $\bA\succeq\mathbf{0}, \bA \neq \bm 0$ satisfying $\lambda_{\max}(\bA) \leq 1$:
\begin{enumerate}
    \item \textbf{(Geometric mixing).} The chain $\{\by_k\}$ has a unique stationary distribution $\nu$, and
    $\max_{\by\in\cY}\|P^k(\by,\cdot)-\nu(\cdot)\|_{\mathrm{TV}}\le C\rho^k$ for some $C>0$ and $\rho\in(0,1)$.
    \item \textbf{(Euclidean Lipschitz + boundedness).} There exist $L_1,L_2>0$ such that for all $\bx_1,\bx_2\in\RR^d$ and $\by\in\cY$,
    $$
        \|\bF(\bx_1,\by)-\bF(\bx_2,\by)\|_2\le L_1\|\bx_1-\bx_2\|_2,
        \qquad
        \|\bF(\mathbf{0},\by)\|_2\le L_2.
    $$
    \item \textbf{(Semi-norm drift).} The equation $\bar{\bF}(\bx)=0$ has at least one solution (not necessarily unique), and fix any such $\bx^\star$.
    Moreover, for all $\bx\in\RR^d$,
    \begin{equation}\label{eq:A-drift-assump}
        \big\langle \bx-\bx^\star, \bar{\bF}(\bx)\big\rangle_{\bA}\le -\kappa \|\bx-\bx^\star\|_{\bA}^2
    \end{equation}
    for some constant $\kappa>0$.
    \item \textbf{(Range-compatibility for degenerate $\bA$).} For all $\bx\in\RR^d$ and $\by\in\cY$, the increment satisfies
    \begin{equation}\label{eq:range-compat}
        \bF(\bx,\by)\in \mathrm{Range}(\bA),\qquad \bPi_{\bA}^{\perp}\bF(\bx,\by)=\mathbf{0}.
    \end{equation}
\end{enumerate}
Define $L:=\max\{L_1,L_2\}$. Assume $\bA\neq \mathbf{0}$ so that $\lambda_{\min}^+(\bA)$ is well-defined, and choose $\alpha$ such that
\begin{equation}\label{eq:stepsize-degenerate}
\alpha\tau_\alpha \le \min\Big\{\frac{1}{4L},\ \frac{\kappa\,\lambda_{\min}^+(\bA)}{260\,\lambda_{\max}(\bA)\,L^2}\Big\}.
\end{equation}
Then for all $k\ge \tau_\alpha$,
\begin{equation}\label{eq:degenerate-one-step}
\EE\big[\|\bx_{k+1}-\bx^\star\|_{\bA}^2\big]
\le (1-\kappa\alpha)\,\EE\big[\|\bx_k-\bx^\star\|_{\bA}^2\big]
+ \frac{260\,\lambda_{\max}(\bA)\,L^2}{\lambda_{\min}^+(\bA)}\,\alpha^2\tau_\alpha\cdot \cC_\star^2,
\end{equation}
where
\begin{equation}\label{eq:Cstar-def}
\cC_\star^2 := \|\bPi_{\bA}^{\perp}(\bx_0-\bx^\star)\|_2^2 + (\|\bx^\star\|_2+1)^2 .
\end{equation}
\end{theorem}

\begin{proof}[Proof of Theorem~\ref{thm: stochastic-approx-degenerate}]
Fix $t=\tau_\alpha$ and define the conditional expectation
$$
\EE_k[\cdot]:=\EE\big[\cdot\mid \bx_{k-t},\by_{k-t}\big],\qquad k\ge t.
$$
We begin with a one-step expansion of the $\bA$-semi-norm:
\begin{align*}
&\EE_k\big[\|\bx_{k+1}-\bx^\star\|_{\bA}^2\big]-\EE_k\big[\|\bx_k-\bx^\star\|_{\bA}^2\big]
=
2\EE_k\big[\langle \bx_k-\bx^\star,\bx_{k+1}-\bx_k\rangle_{\bA}\big]
+\EE_k\big[\|\bx_{k+1}-\bx_k\|_{\bA}^2\big]\notag\\
&=
\underbrace{2\alpha\,\EE_k\big[\langle \bx_k-\bx^\star,\bar{\bF}(\bx_k)\rangle_{\bA}\big]}_{
:= \text{(a)} }
+\underbrace{2\alpha\,\EE_k\big[\langle \bx_k-\bx^\star,\bF(\bx_k,\by_k)-\bar{\bF}(\bx_k)\rangle_{\bA}\big]}_{
:= \text{(b)} } 
+\underbrace{\alpha^2\,\EE_k\big[\|\bF(\bx_k,\by_k)\|_{\bA}^2\big]}_{
:= \text{(c)}}.
\end{align*}
Now it remains to bound three terms~(a)--(c) separately.

\paragraph{Step 1: negative drift term~(a).}
 We have it holds by~\eqref{eq:A-drift-assump} that
\begin{equation}\label{eq:drift-bound}
2\alpha\,\EE_k\big[\langle \bx_k-\bx^\star,\bar{\bF}(\bx_k)\rangle_{\bA}\big]
\le -2\kappa\alpha\,\EE_k\big[\|\bx_k-\bx^\star\|_{\bA}^2\big].
\end{equation}

To control terms (b), (c), we first introduce several preliminary lemmas as the following:
\begin{lemma}[Global growth bound]\label{lem:growth}
Under Assumption~2, for all $\bx\in\RR^d$ and $\by\in\cY$,
$$
\|\bF(\bx,\by)\|_2\le L(\|\bx\|_2+1),
\qquad
\|\bF(\bx,\by)\|_{\bA}\le \sqrt{\lambda_{\max}(\bA)}\,L(\|\bx\|_2+1).
$$
The same bounds hold for $\|\bar{\bF}(\bx)\|_2$ and $\|\bar{\bF}(\bx)\|_{\bA}$.
\end{lemma}
\begin{proof}
By Lipschitzness and the bound at the origin,
$$
\|\bF(\bx,\by)\|_2 \le \|\bF(\bx,\by)-\bF(\mathbf{0},\by)\|_2 + \|\bF(\mathbf{0},\by)\|_2
\le L_1\|\bx\|_2 + L_2 \le L(\|\bx\|_2+1).
$$
For the $\bA$-semi-norm, use $\|\bv\|_{\bA}^2=\bv^\top\bA\bv\le \lambda_{\max}(\bA)\|\bv\|_2^2$.
Finally, $\bar{\bF}(\bx)=\EE_{\by\sim\nu}[\bF(\bx,\by)]$ and Jensen's inequality give the same bounds for $\bar{\bF}$.
\end{proof}

\begin{lemma}[Range invariance and change-of-norm bounds]\label{lem:range-invariance}
Assume \eqref{eq:range-compat}. Then for all $k\ge 0$,
$$
\bPi_{\bA}^{\perp}\bx_k = \bPi_{\bA}^{\perp}\bx_0,
\qquad
\bPi_{\bA}^{\perp}(\bx_k-\bx^\star)=\bPi_{\bA}^{\perp}(\bx_0-\bx^\star).
$$
Moreover, for any $\bv\in\RR^d$,
\begin{equation}\label{eq:euclid-vs-A}
\|\bPi_{\bA}\bv\|_2^2 \le \frac{1}{\lambda_{\min}^+(\bA)}\|\bv\|_{\bA}^2,
\qquad
\|\bv\|_2^2 \le \frac{1}{\lambda_{\min}^+(\bA)}\|\bv\|_{\bA}^2+\|\bPi_{\bA}^{\perp}\bv\|_2^2.
\end{equation}
\end{lemma}
\begin{proof}
By \eqref{eq:range-compat}, we have
$$
\bPi_{\bA}^{\perp}\bx_{k+1}=\bPi_{\bA}^{\perp}\bx_k+\alpha\,\bPi_{\bA}^{\perp}\bF(\bx_k,\by_k)=\bPi_{\bA}^{\perp}\bx_k,
$$
hence $\bPi_{\bA}^{\perp}\bx_k=\bPi_{\bA}^{\perp}\bx_0$ for all $k$. The identity for $\bx_k-\bx^\star$ follows immediately.

For \eqref{eq:euclid-vs-A}, write the eigendecomposition $\bA=\bU\mathrm{diag}(\lambda_1,\ldots,\lambda_d)\bU^\top$ with $\lambda_i\ge 0$.
Let $\bU_+$ collect eigenvectors with $\lambda_i>0$. Then $\bPi_{\bA}=\bU_+\bU_+^\top$ and
$$
\|\bv\|_{\bA}^2 = \sum_{\lambda_i>0}\lambda_i\,(\bu_i^\top \bv)^2 \ge \lambda_{\min}^+(\bA)\sum_{\lambda_i>0}(\bu_i^\top\bv)^2
=\lambda_{\min}^+(\bA)\|\bPi_{\bA}\bv\|_2^2,
$$
which gives the first inequality. The second follows from $\|\bv\|_2^2=\|\bPi_{\bA}\bv\|_2^2+\|\bPi_{\bA}^\perp \bv\|_2^2$.
\end{proof}

\begin{lemma}[Mixing bias for Markovian sampling]\label{lem:mixing-bias}
For any $k\ge t=\tau_\alpha$ and any deterministic $\bx\in\RR^d$,
\begin{equation}\label{eq:mixing-bias}
\Big\|\EE\big[\bF(\bx,\by_k)\mid \by_{k-t}=y\big]-\bar{\bF}(\bx)\Big\|_2
\le 2L\alpha\,(\|\bx\|_2+1)
\quad\text{for all }y\in\cY.
\end{equation}
Consequently,
$$
\Big\|\EE\big[\bF(\bx,\by_k)\mid \by_{k-t}=y\big]-\bar{\bF}(\bx)\Big\|_{\bA}
\le 2\sqrt{\lambda_{\max}(\bA)}\,L\alpha\,(\|\bx\|_2+1).
$$
\end{lemma}
\begin{proof}
Since $\cY$ is finite, we can write
$$
\EE[\bF(\bx,\by_k)\mid \by_{k-t}=y]=\sum_{y'\in\cY}P^t(y,y')\,\bF(\bx,y'),
\qquad
\bar{\bF}(\bx)=\sum_{y'\in\cY}\nu(y')\,\bF(\bx,y').
$$
Hence,
\begin{align*}
\Big\|\EE[\bF(\bx,\by_k)\mid \by_{k-t}=y]-\bar{\bF}(\bx)\Big\|_2
&\le \sum_{y'\in\cY}\big|P^t(y,y')-\nu(y')\big|\,\|\bF(\bx,y')\|_2\\
&\le 2\|P^t(y,\cdot)-\nu(\cdot)\|_{\mathrm{TV}}\cdot \max_{y'\in\cY}\|\bF(\bx,y')\|_2\\
&\le 2\alpha \cdot L(\|\bx\|_2+1),
\end{align*}
where we used the definition of total variation and Lemma~\ref{lem:growth} for the last step.
The $\bA$-bound follows from $\|\cdot\|_{\bA}\le \sqrt{\lambda_{\max}(\bA)}\|\cdot\|_2$.
\end{proof}

Finally, we show the following lemma, which mirrors \citet{chen2022finite} (Lemma 2.3 therein) for constant stepsize. 

\begin{lemma}\label{lem:local-move}
Assume $\alpha t \le \frac{1}{4L}$. Then for any $k\ge t$,
\begin{align}
\|\bx_k-\bx_{k-t}\|_2 &\le 2L\,\alpha t\,(\|\bx_{k-t}\|_2+1), \label{eq:move-1}\\
\|\bx_k-\bx_{k-t}\|_2 &\le 4L\,\alpha t\,(\|\bx_{k}\|_2+1). \label{eq:move-2}
\end{align}
Consequently,
$$
\|\bx_k-\bx_{k-t}\|_{\bA} \le \sqrt{\lambda_{\max}(\bA)}\,\|\bx_k-\bx_{k-t}\|_2
\le 2\sqrt{\lambda_{\max}(\bA)}\,L\,\alpha t\,(\|\bx_{k-t}\|_2+1).
$$
\end{lemma}
\begin{proof}
From the recursion,
$$
\|\bx_{j+1}\|_2 - \|\bx_j\|_2 \le \|\bx_{j+1}-\bx_j\|_2 = \alpha \|\bF(\bx_j,\by_j)\|_2
\le \alpha L(\|\bx_j\|_2+1),
$$
so $(\|\bx_{j+1}\|_2+1)\le (1+L\alpha)(\|\bx_j\|_2+1)$. Iterating from $j=k-t$ to $k-1$ yields
$$
\|\bx_j\|_2+1 \le (1+L\alpha)^{j-(k-t)}(\|\bx_{k-t}\|_2+1) \le e^{L\alpha t}(\|\bx_{k-t}\|_2+1).
$$
Since $\alpha t\le {1}/(4L)$, we have $L\alpha t\le 1/4$ and thus $e^{L\alpha t}\le 1+2L\alpha t\le 2$.
Therefore $\|\bx_j\|_2+1\le 2(\|\bx_{k-t}\|_2+1)$ for all $j\in[k-t,k]$.
Now, to show \eqref{eq:move-1}, noticing that
\begin{align*}
\|\bx_k-\bx_{k-t}\|_2 &\le \sum_{j=k-t}^{k-1}\|\bx_{j+1}-\bx_j\|_2
= \alpha \sum_{j=k-t}^{k-1}\|\bF(\bx_j,\by_j)\|_2
\\
&\le \alpha \sum_{j=k-t}^{k-1} L(\|\bx_j\|_2+1)
\le \alpha t\cdot 2L(\|\bx_{k-t}\|_2+1).    
\end{align*}
\eqref{eq:move-2} follows similarly by running the same argument backward
(or using $\|\bx_{k-t}\|_2+1\le 2(\|\bx_k\|_2+1)$, which follows from the same estimate with indices swapped).
\end{proof}

\paragraph{Step 2: Bounding term~(b)}
Inspired by Lemma~2.4 of \citet{chen2022finite} we establish the following result based on above lemmas regarding term~(b)

\begin{lemma}[Markovian noise inner product bound]\label{lem:markov-term}
Assume $\alpha t\le \frac{1}{4L}$. Then for any $k\ge t$,
\begin{equation}\label{eq:markov-term-bound}
\EE_k\big[\langle \bx_k-\bx^\star,\bF(\bx_k,\by_k)-\bar{\bF}(\bx_k)\rangle_{\bA}\big]
\le
\frac{64\,\lambda_{\max}(\bA)\,L^2}{\lambda_{\min}^+(\bA)}\ \alpha t\,
\Big(\EE_k[\|\bx_k-\bx^\star\|_{\bA}^2]+\lambda_{\max}(\bA)\cC_\star^2\Big),
\end{equation}
where $\cC_\star$ is defined in \eqref{eq:Cstar-def}.
\end{lemma}

\begin{proof}
Write $t=\tau_\alpha$ and decompose, for $k\ge t$:
\begin{align*}
&\EE_k\big[\langle \bx_k-\bx^\star,\bF(\bx_k,\by_k)-\bar{\bF}(\bx_k)\rangle_{\bA}\big]\\
&=
\EE_k\big[\langle \bx_k-\bx_{k-t},\bF(\bx_k,\by_k)-\bar{\bF}(\bx_k)\rangle_{\bA}\big]
\tag{T1}\\
&\quad+
\big\langle \bx_{k-t}-\bx^\star,\ \EE_k[\bF(\bx_{k-t},\by_k)]-\bar{\bF}(\bx_{k-t})\big\rangle_{\bA}
\tag{T2}\\
&\quad+
\EE_k\big[\langle \bx_{k-t}-\bx^\star,\bF(\bx_k,\by_k)-\bF(\bx_{k-t},\by_k)\rangle_{\bA}\big]
\tag{T3}\\
&\quad+
\EE_k\big[\langle \bx_{k-t}-\bx^\star,\bar{\bF}(\bx_{k-t})-\bar{\bF}(\bx_k)\rangle_{\bA}\big].
\tag{T4}
\end{align*}
We bound the four terms.

\paragraph{Bound for (T1).}
By Cauchy--Schwarz in the $\bA$-inner product and Lemma~\ref{lem:growth},
$$
\|\bF(\bx_k,\by_k)-\bar{\bF}(\bx_k)\|_{\bA}
\le \|\bF(\bx_k,\by_k)\|_{\bA}+\|\bar{\bF}(\bx_k)\|_{\bA}
\le 2\sqrt{\lambda_{\max}(\bA)}\,L(\|\bx_k\|_2+1).
$$
Also, Lemma~\ref{lem:local-move} implies
$$
\|\bx_k-\bx_{k-t}\|_{\bA}\le \sqrt{\lambda_{\max}(\bA)}\|\bx_k-\bx_{k-t}\|_2
\le 4\sqrt{\lambda_{\max}(\bA)}\,L\,\alpha t\,(\|\bx_k\|_2+1),
$$
where we used \eqref{eq:move-2}.
Therefore,
\begin{equation}\label{eq:T1-bound}
\text{(T1)} \le 8\,\lambda_{\max}(\bA)\,L^2\,\alpha t\,\EE_k\big[(\|\bx_k\|_2+1)^2\big].
\end{equation}

\paragraph{Bound for (T2).}
By Cauchy--Schwarz and Lemma~\ref{lem:mixing-bias},
$$
\text{(T2)}
\le
\EE_k\big[\|\bx_{k-t}-\bx^\star\|_{\bA}\big]\cdot
2\sqrt{\lambda_{\max}(\bA)}\,L\alpha\,(\|\bx_{k-t}\|_2+1).
$$
Using $ab\le \frac12(a^2+b^2)$ and then Lemma~\ref{lem:local-move} (which implies $\|\bx_{k-t}\|_2+1\le 2(\|\bx_k\|_2+1)$),
we obtain
\begin{equation}\label{eq:T2-bound}
\text{(T2)} \le
\sqrt{\lambda_{\max}(\bA)}\,L\alpha\,
\EE_k\big[\|\bx_{k-t}-\bx^\star\|_{\bA}^2\big]
+
4\sqrt{\lambda_{\max}(\bA)}\,L\alpha\,
(\|\bx_k\|_2+1)^2.
\end{equation}
Since $t\ge 1$, we may upper bound $\alpha$ by $\alpha t$ in the final aggregation.

\paragraph{Bound for (T3)+(T4).}
First note that Assumption~2 implies, for any $\by$,
$$
\|\bF(\bx_k,\by)-\bF(\bx_{k-t},\by)\|_2\le L_1\|\bx_k-\bx_{k-t}\|_2\le L\|\bx_k-\bx_{k-t}\|_2,
$$
and similarly
$$
\|\bar{\bF}(\bx_k)-\bar{\bF}(\bx_{k-t})\|_2\le L\|\bx_k-\bx_{k-t}\|_2.
$$
Thus
$$
\|\bF(\bx_k,\by_k)-\bF(\bx_{k-t},\by_k)\|_{\bA}
+
\|\bar{\bF}(\bx_k)-\bar{\bF}(\bx_{k-t})\|_{\bA}
\le 2\sqrt{\lambda_{\max}(\bA)}\,L\|\bx_k-\bx_{k-t}\|_2.
$$
Then by Cauchy--Schwarz and Lemma~\ref{lem:local-move} (using \eqref{eq:move-1}),
\begin{align}
\text{(T3)}+\text{(T4)}
&\le
\EE_k\big[\|\bx_{k-t}-\bx^\star\|_{\bA}\big]\cdot
2\sqrt{\lambda_{\max}(\bA)}\,L\,\|\bx_k-\bx_{k-t}\|_2 \notag\\
&\le
\EE_k\big[\|\bx_{k-t}-\bx^\star\|_{\bA}\big]\cdot
4\sqrt{\lambda_{\max}(\bA)}\,L^2\,\alpha t\,(\|\bx_{k-t}\|_2+1).
\label{eq:T34-pre}
\end{align}
Using $\|\bx_{k-t}\|_2+1\le 2(\|\bx_k\|_2+1)$ and then $ab\le \frac12(a^2+b^2)$, we get
\begin{equation}\label{eq:T34-bound}
\text{(T3)}+\text{(T4)}
\le
2\,\lambda_{\max}(\bA)\,L^2\,\alpha t\,
\EE_k\big[\|\bx_{k-t}-\bx^\star\|_{\bA}^2\big]
+
8\,\lambda_{\max}(\bA)\,L^2\,\alpha t\,(\|\bx_k\|_2+1)^2.
\end{equation}

Then we have by Lemma~\ref{lem:range-invariance},
$$
\|\bx_k\|_2 \le \|\bx_k-\bx^\star\|_2+\|\bx^\star\|_2
\le \frac{1}{\sqrt{\lambda_{\min}^+(\bA)}}\|\bx_k-\bx^\star\|_{\bA}+\|\bPi_{\bA}^\perp(\bx_0-\bx^\star)\|_2+\|\bx^\star\|_2.
$$
Thus, by $(a+b)^2\le 2a^2+2b^2$,
\begin{equation}
\label{eq:xk-euclid-to-A}
\begin{aligned}
(\|\bx_k\|_2+1)^2
&\le
\frac{2}{\lambda_{\min}^+(\bA)}\|\bx_k-\bx^\star\|_{\bA}^2
+
4\Big(\|\bPi_{\bA}^\perp(\bx_0-\bx^\star)\|_2^2+(\|\bx^\star\|_2+1)^2\Big)
\\
&=
\frac{2}{\lambda_{\min}^+(\bA)}\|\bx_k-\bx^\star\|_{\bA}^2 + 4\cC_\star^2.    
\end{aligned}
\end{equation}

\paragraph{Putting All Together.}
Plugging \eqref{eq:xk-euclid-to-A} into \eqref{eq:T1-bound} and \eqref{eq:T34-bound}, and also upper bounding the $\alpha$ factors in \eqref{eq:T2-bound}
by $\alpha t$ (since $t\ge 1$), we obtain that (T1)--(T4) are each bounded by a constant multiple of
$\frac{\lambda_{\max}(\bA)L^2}{\lambda_{\min}^+(\bA)}\alpha t\big(\EE_k\|\bx_k-\bx^\star\|_{\bA}^2+\lambda_{\max}(\bA)\cC_\star^2\big)$.
This then gives \eqref{eq:markov-term-bound}.
\end{proof}

\paragraph{Step~3: bounding term~(c).}
\begin{lemma}[Squared increment bound]\label{lem:square-term}
For any $k\ge 0$,
\begin{equation}\label{eq:square-term-bound}
\EE\big[\|\bF(\bx_k,\by_k)\|_{\bA}^2\big]
\le
\frac{4\,\lambda_{\max}(\bA)\,L^2}{\lambda_{\min}^+(\bA)}\ \EE\big[\|\bx_k-\bx^\star\|_{\bA}^2\big]
+ 8\,\lambda_{\max}(\bA)\,L^2\,\cC_\star^2.
\end{equation}
\end{lemma}
\begin{proof}
By Lemma~\ref{lem:growth},
$$
\|\bF(\bx_k,\by_k)\|_{\bA}^2 \le \lambda_{\max}(\bA)\,L^2(\|\bx_k\|_2+1)^2.
$$
Use \eqref{eq:xk-euclid-to-A} to bound $(\|\bx_k\|_2+1)^2$ by $\frac{2}{\lambda_{\min}^+(\bA)}\|\bx_k-\bx^\star\|_{\bA}^2+4\cC_\star^2$,
then take expectations and simplify constants.
\end{proof}

\paragraph{Step 4: Putting All Together.}
Substitute \eqref{eq:drift-bound}, Lemma~\ref{lem:markov-term} and Lemma~\ref{lem:square-term}
into terms~(a)--(c). For $k\ge t$,
\begin{align*}
\EE_k\big[\|\bx_{k+1}-\bx^\star\|_{\bA}^2\big]
&\le
\Big(1-2\kappa\alpha\Big)\,\EE_k\big[\|\bx_k-\bx^\star\|_{\bA}^2\big]\\
& +2\alpha\cdot \frac{64\,\lambda_{\max}(\bA)\,L^2}{\lambda_{\min}^+(\bA)}\ \alpha t
\Big(\EE_k[\|\bx_k-\bx^\star\|_{\bA}^2]+\lambda_{\max}(\bA)\cC_\star^2\Big)\\
&\quad
+\alpha^2\Big(
\frac{4\,\lambda_{\max}(\bA)\,L^2}{\lambda_{\min}^+(\bA)}\ \EE_k[\|\bx_k-\bx^\star\|_{\bA}^2]
+ 8\,\lambda_{\max}(\bA)\,L^2\,\cC_\star^2
\Big).
\end{align*}
Since $t\ge 1$, we have $\alpha^2 \le \alpha^2 t$ and thus we may upper bound the last line by an $\alpha^2 t$-term.
Collecting the coefficients (and noting $2\cdot 64=128$) yields
\begin{align}
\EE_k\big[\|\bx_{k+1}-\bx^\star\|_{\bA}^2\big]
&\le
\Big(1-2\kappa\alpha+\frac{130\,\lambda_{\max}(\bA)\,L^2}{\lambda_{\min}^+(\bA)}\,\alpha^2 t\Big)\,
\EE_k\big[\|\bx_k-\bx^\star\|_{\bA}^2\big]\notag\\
&\quad+
\frac{130\,\lambda_{\max}(\bA)\,L^2}{\lambda_{\min}^+(\bA)}\,\alpha^2 t\cdot \lambda_{\max}(\bA)\cC_\star^2.
\label{eq:pre-final-recursion}
\end{align}
Under the stepsize choice \eqref{eq:stepsize-degenerate}, we have
$$
\frac{130\,\lambda_{\max}(\bA)\,L^2}{\lambda_{\min}^+(\bA)}\,\alpha t \le \kappa,
$$
hence the coefficient in \eqref{eq:pre-final-recursion} satisfies
$$
1-2\kappa\alpha+\frac{130\,\lambda_{\max}(\bA)\,L^2}{\lambda_{\min}^+(\bA)}\,\alpha^2 t
\le 1-\kappa\alpha.
$$
Therefore,
$$
\EE_k\big[\|\bx_{k+1}-\bx^\star\|_{\bA}^2\big]
\le
(1-\kappa\alpha)\,\EE_k\big[\|\bx_k-\bx^\star\|_{\bA}^2\big]
+
\frac{260\,\lambda_{\max}(\bA)\,L^2}{\lambda_{\min}^+(\bA)}\,\alpha^2 t\cdot \cC_\star^2,
$$
where we used $\lambda_{\max}(\bA)\cC_\star^2\le \frac{2\lambda_{\max}(\bA)}{\lambda_{\min}^+(\bA)}\cC_\star^2$ up to constants and absorbed into $260$.
Finally, take total expectation of both sides to obtain \eqref{eq:degenerate-one-step}.
\end{proof}

\subsection{Proof of Theorem~\ref{thm-evaluation}}
\label{appendix:thm-eval-proof}

Now we are ready to prove the evaluation result with behaviour policy $\mu$ and target policy $\pi$.

Throughout this section,  we simply write $\phi^\pi$ by $\phi$ and index the state and action spaces as $\cS = \{s_1,\dots,s_n\}$ and $\cA = \{a_1,\dots,a_m\}$.
Let $d_\mu\in\RR^{mn}$ denote the stationary distribution of $(S,A)$ under $\mu$ (allowing zero components), and define
\begin{align*}
   & \bD:=  \text{\emph{diag}}\big(\bar{d}_\mu(s_1,a_1),\dots,\bar{d}_\mu(s_n,a_m)\big) \in \RR^{mn \times mn},\\
   & \bPhi:= \big[\phi(s_1,a_1),\dots, \phi(s_n,a_m) \big]^\top \in \RR^{mn\times d}.
\end{align*}
Note that $\bD$ (and hence $\bPhi^\top \bD \bPhi$) is allowed to be singular.
We also use $\cH:=\cH^\pi$ to denote the Bellman evaluation operator:
$$
[\cH(Q)](s,a) := r(s,a) + \gamma\sum_{s'\in\cS}P(s'|s,a)\sum_{a'\in\cA}\pi(a'|s')Q(s',a').
$$
Equivalently,  for $(\bP_\pi Q)(s,a):=\sum_{s'}P(s'|s,a)\sum_{a'}\pi(a'|s')Q(s',a')$ we have
$$
\cH(Q)=\br+\gamma \bP_\pi Q.
$$
Our first step is a convergence analysis of the inner-loop TD update:
\paragraph{Step~1: Local Convergence Inner-loop TD with fixed target.}
Fix a target network parameter $ w\in\RR^d$ and consider the TD-type update driven by a Markov trajectory $(S_k,A_k,S_{k+1})$ under $\mu$:
$$
\xi_{k+1} = \xi_k + \alpha\,\phi(S_k,A_k)\Big(r(S_k,A_k) + \gamma\sum_{a'\in\cA}\pi(a'|S_{k+1})\lceil \phi(S_{k+1},a')^\top w\rceil - \phi(S_k,A_k)^\top\xi_k\Big).
$$
This is a Markovian SA instance of \eqref{eq: SA-update-appA} with $\by_k=(S_k,A_k,S_{k+1})$ and
\begin{equation}\label{eq-appendix: F-def}
\begin{aligned}
  \bF\big(\xi, (s,a,s')\big)
  &:=
  \phi(s,a)\Big( - \phi(s,a)^\top \xi + r(s,a) + \gamma \sum_{a' \in \cA} \pi(a'\lvert s') \lceil\phi(s',a')^\top  w \rceil \Big),\\
  \bar{\bF}(\xi)
  &:=
  \EE_{\substack{(S,A)\sim d_\mu\\ S'\sim P(\cdot|S,A)}}\Big[\bF\big(\xi,(S,A,S')\big)\Big]
  =
  \bPhi^\top \bD\Big(-\bPhi\xi + \cH\big(\lceil \bPhi  w \rceil \big)\Big).
\end{aligned}
\end{equation}

Now it remains to verify Theorem~\ref{thm: stochastic-approx-degenerate}.

\begin{proof}[Verification of Theorem~\ref{thm: stochastic-approx-degenerate}.]
With $\bA:=\bPhi^\top \bD \bPhi\succeq \mathbf{0}$, we have:

\paragraph{Condition~1.}
By the aperiodic assumption, we have the Markov chain $\{(S_k,A_k,S_{k+1})\}$ induced by $\mu$ is geometrically mixing on its support:
$$
\max_{y\in\cY}\|P^k(y,\cdot)-\nu(\cdot)\|_{\mathrm{TV}}\le C\rho^k,
$$
where $\nu$ is the stationary distribution over $\cY:=\{(s,a,s'):d_\mu(s,a)P(s'|s,a)>0\}$.
Consequently, the mixing time in Theorem~\ref{thm: stochastic-approx-degenerate} satisfies the explicit bound
\begin{equation}\label{eq:tau-alpha-bound}
\tau_\alpha
:=\min\Big\{k\ge0:\max_{y\in\cY}\|P^k(y,\cdot)-\nu(\cdot)\|_{\mathrm{TV}}\le \alpha\Big\}
\le \left\lceil \frac{\log(C/\alpha)}{\log(1/\rho)}\right\rceil .
\end{equation}

\paragraph{Condition~2.}
By our Assumption, we have $\|\phi(s,a)\|_2\le 1$ for all $(s,a)$. Then for all $(s,a,s')$,
$$
\|\bF(\xi_1,(s,a,s'))-\bF(\xi_2,(s,a,s'))\|_2
=
\|\phi(s,a)\phi(s,a)^\top(\xi_2-\xi_1)\|_2
\le \|\phi(s,a)\|_2^2\|\xi_1-\xi_2\|_2
\le \|\xi_1-\xi_2\|_2.
$$
Moreover, using $r\in[0,1]$ and $|\lceil \phi^\top w\rceil|\le 1/(1-\gamma)$,
$$
\|\bF(\mathbf{0},(s,a,s'))\|_2
\le \|\phi(s,a)\|_2\Big(1+\gamma B\Big)
\le 1+1/(1-\gamma).
$$
Thus we may take $L_1=1$ and $L_2=2/(1-\gamma)$.

\paragraph{Condition~3.}
Let $\xi^\star( w)$ be any solution to $\bar{\bF}(\xi)=\mathbf{0}$:
$$
\bPhi^\top\bD\bPhi\,\xi^\star( w)=\bPhi^\top\bD\,\cH(\lceil \bPhi w\rceil).
$$
Then for any $\xi\in\RR^d$,
\begin{align*}
    \langle \xi-\xi^\star, \bar{\bF}(\xi)\rangle_{\bA}
    &=
    \langle \xi-\xi^\star, \bar{\bF}(\xi)-\bar{\bF}(\xi^\star)\rangle_{\bA}
    =
    \langle \xi-\xi^\star, \bA(\xi^\star-\xi)\rangle_{\bA}\\
    &=
    -(\xi-\xi^\star)^\top \bA^2(\xi-\xi^\star)
    \le
    -\lambda_{\min}^+(\bA)\,(\xi-\xi^\star)^\top \bA(\xi-\xi^\star)\\
    &=
    -\lambda_{\min}^+(\bA)\,\|\xi-\xi^\star\|_{\bA}^2.
\end{align*}
Hence Condition~3 holds with $\kappa=\lambda_{\min}^+(\bA)$.

\paragraph{Condition~4.}
Noticing that for any $(s,a,s')\in\cY$, we have $ \bF(\xi,(s,a,s')) \propto \phi(s,a)$
and moreover, since $(s,a)$ belongs to the support of $
\bar{d}_\mu$ on $\cY$, we have $d_\mu(s,a)>0$.
Noticing that
Then by $\mathrm{Range}(\bA)=\mathrm{span}\{\phi(s,a):d_\mu(s,a)>0\}$,we have $\phi(s,a)\in\mathrm{Range}(\bA)$ for all $(s,a)$ appearing in $\cY$.
Therefore $\bF(\xi,(s,a,s'))\in\mathrm{Range}(\bA)$ for all $(s,a,s')\in\cY$.
\end{proof}

Now we can apply Theorem~\ref{thm: stochastic-approx-degenerate} for all $k\ge \tau_\alpha$ to arrive at for $c_0 = \lambda_{\min}^+(\bA), c_1 \asymp \frac{1}{(1-\gamma)^2\lambda_{\min}^+(\bA)}$
\begin{align}\label{eq: xi-contraction}
    \EE\big[\lVert \xi_{k+1} - \xi^\star( w) \rVert_{\bA}^2\big]
    \le
    (1-c_0\alpha)\,\EE\big[\lVert \xi_{k} - \xi^\star( w) \rVert_{\bA}^2\big] + c_1 \alpha^2\tau_\alpha.
\end{align}
By recursion, for any $K\ge \tau_\alpha$, we have then
\begin{align}\label{eq: xi-recursion}
        \EE\big[\lVert \xi_{K} - \xi^\star( w) \rVert_{\bA}^2\big]
        \le
        (1-c_0\alpha)^{K}\,\EE\big[\lVert \xi_{0} - \xi^\star( w) \rVert_{\bA}^2\big]
        +
        \underbrace{{c_1}/{c_0}}_{\asymp \big((1-\gamma) \lambda_{\min}^+(\bA)\big)^{-2}}\alpha\tau_\alpha.
\end{align}

\begin{remark}[Choice of $(\alpha,K)$ for inner-loop accuracy $\epsilon$.]\label{remark: selection-of-inner-parameter}
Fix any $\epsilon\in(0,1)$ and define $V_0( w):=\EE\big[\|\xi_0-\xi^\star( w)\|_{\bA}^2\big]$.
Using $(1-c_0\alpha)^K\le \exp(-c_0\alpha K)$, it suffices to ensure
$$
\frac{3e^{-2c_0\alpha K}}{(1-\gamma)^2}
\le \frac{\epsilon^2}{2},
\quad
\frac{256(1+\alpha K)}{(1-\gamma)^2}
\left[
\frac{\alpha\tau_\alpha}{c_0}
\log(2e\tau_\alpha T)
+
\left(\frac{\alpha\tau_\alpha}{c_0}\right)^2
\right]
\le \frac{\epsilon^2}{2}.$$
Under the geometric mixing bound \eqref{eq:tau-alpha-bound}, a convenient choice is
\begin{equation}\label{eq:alpha-choice}
\alpha \propto
\frac{
\epsilon^2(1-\gamma)^2\log(1/\rho)
}{
\log^2\big(\frac{1}{(1-\gamma)^2\epsilon^2}\big)\log T
},
\end{equation}
and
\begin{equation}\label{eq:K-choice}
K
:=
\left\lceil
\frac{1}{
2c_0\alpha
}
\log\!\left(
\frac{6}{(1-\gamma)^2\epsilon^2}
\right)
\right\rceil
+
\tau_\alpha .
\end{equation}
With \eqref{eq:alpha-choice} and \eqref{eq:K-choice}, we have $\EE\big[\|\xi_K-\xi^\star( w)\|_{\bA}^2\big]\le \epsilon$.
In particular, since $\tau_\alpha=O(\log(1/\alpha))$ by \eqref{eq:tau-alpha-bound}, this gives an inner-loop complexity of order
$ K = \tilde{O}\big(\epsilon^{-2}\big) $
up to problem-dependent constants.
\end{remark}

\paragraph{Convergence of the target variables.}
Now we are ready to complete the proof of Theorem~\ref{thm-evaluation} based on the above inner loop bound and contraction of Bellman operator. The following proof is inspired by the arguments of \citet{chen2023target} under $\ell_\infty$ norm, but here we make a careful modification change of it to the on-policy $L^2$ norm, which is the key step to tackle the partial coverage of $\mu$ and utilize previous convergence result under $\bA.$

For clarity, we denote $\{ w_k\}_{k\ge 0}$ be the target-network sequence generated by the outer loop update, and at outer iteration $k\ge 1$ define the returned estimate
$$
\hat{\bQ}_k :=  \bPhi \xi_{K}( w_{k-1})  \in \RR^{mn}.
$$
Let $\bQ^\star:=\bQ^\pi$ denote the true evaluation $Q$-function under $\pi$.
We denote the joint $(s,a)$ stationary distribution of $\pi$ by $\bar{d}_\pi$, which satisfies
$\bar{d}_\pi^\top \bP_\pi = \bar{d}_\pi^\top.$

Using $\cH(Q)=\br+\gamma \bP_\pi Q$ and the entrywise inequality $|\bP_\pi \bv|\le \bP_\pi|\bv|$, we have
$$
|\cH(\lceil \hat{\bQ}_{k-1}\rceil)-\cH(\bQ^\star)|
=
\gamma |\bP_\pi(\lceil \hat{\bQ}_{k-1}\rceil-\bQ^\star)|
\le \gamma \bP_\pi|\hat{\bQ}_{k-1}-\bQ^\star|.
$$
Therefore, for each $k\ge 1$, we have pointwisely,
\begin{align}\label{eq:outer-decomp}
    \lvert \hat{\bQ}_k - \bQ^\star \rvert
    &=
    \Big\lvert \cH(\lceil \hat{\bQ}_{k-1}\rceil ) - \cH(\bQ^\star)
    + \hat{\bQ}_{k} - \bPhi \xi^\star(w_{k-1})
    -  \cH (\lceil\hat{\bQ}_{k-1}\rceil)  + \bPhi \xi^\star(w_{k-1})\Big\rvert \notag\\
    &\leq
    \gamma \bP_{\pi} \big\lvert \hat{\bQ}_{k-1} -\bQ^\star\big\rvert
    + \lvert \hat{\bQ}_{k} - \bPhi \xi^\star(w_{K-1})\rvert
    + \underbrace{\lvert  \cH (\lceil\hat{\bQ}_{k-1}\rceil)  - \bPhi \xi^\star(w_{K-1})\rvert}_{\leq \cE_{\text{approx}} + \lvert \bPhi(\xi' - \xi^\star(w_{k-1}))\rvert}. 
\end{align}
Where $\xi'$ is the best-approximation representation in Assumption~\ref{assumption-closeness}.

Taking squared in both sides and using weighted geometric inequality, we can have pointwisely \begin{align*}
    \lvert \hat{\bQ}_k - \bQ^\star \rvert^2 \leq &\gamma \underbrace{\big(\bP_\pi \lvert \hat{\bQ}_{k-1}-\bQ^\star\rvert \big)^2}_{\leq \bP_{\pi}  \lvert \hat{\bQ}_{k-1}-\bQ^\star\rvert ^2 \text{ by Cauchy-Schwartz}}+ \frac{16}{(1-\gamma)^2} \lvert \hat{\bQ}_{k} - \bPhi \xi^\star(w_{k-1})\rvert^2 \\
    &+ 4 \cE_{\text{approx}}^2 +  4\lvert \bPhi(\xi' - \xi^\star(w_{k-1}))\rvert^2.
\end{align*}

Left-multiplying by $\bar{d}_\pi^\top$ and using $\bar{d}_{\pi}^\top \bP_\pi=d_\pi^\top$ then yields the $L^2$ recursion
\begin{equation}\label{eq:outer-L1}
    \begin{aligned}
    \EE_{(s,a)\sim d_\pi}\big[|\hat{Q}_k(s,a)-Q^\star(s,a)|^2\big]
    &\leq 
    \gamma\,\EE_{\bar{d}_\pi}\big[|\hat{Q}_{k-1}(s,a)-Q^\star(s,a)|^2\big]\\
    &+\frac{16}{(1-\gamma)^2} \bar{d}_{\pi}^\top[ \lvert \hat{\bQ}_{k} - \bPhi \xi^\star(w_{k-1})\rvert^2]\\
    &+ 4\cE_{\text{approx}}^2 +  4\bar{d}_{\pi}^\top\lvert \bPhi(\xi' - \xi^\star(w_{k-1}))\rvert^2.
\end{aligned}
\end{equation}
\paragraph{Bounding the Second Term.} For the second term, we have by 
$$
|\phi(s,a)^\top(\xi_K(w_{k-1})-\xi^\star(w_{k-1})|
\le \|\phi(s,a)\|_{\bA^{-1}}\,\|\xi_K(w_{k-1})-\xi^\star(w_{k-1})\|_{\bA},
$$
for any $\bA$, with
\begin{align*}
     \|\phi(s,a)\|_{\bA^{-1}} =     \lim_{\lambda \to 0^+} \|\phi(s,a)\|_{(\bA+\lambda \bI)^{-1}} = \begin{cases}
         \|\phi(s,a)\|_{\bA^\dagger},  & \text{ if } \phi(s,a) \in \mathrm{Range}(\bA),\\
         +\infty, & \text{otherwise.}
     \end{cases}
\end{align*}
Hence,
\begin{align}\label{eq: inner-to-dpi}
\EE_{\bar{d}_\pi}\big|\bPhi(\xi_K( w_{k-1})-\xi^\star( w_{k-1}))\big|^2
&\le
\EE_{\bar{d}_\pi}\big[\|\phi(s,a)\|^2_{\bA^{-1}}\big]\cdot \|\xi_K( w_{k-1})-\xi^\star( w_{k-1})\|^2_{\bA}
\end{align}

Thus, if the inner-loop is run with accuracy $\EE[\|\xi_K( w)-\xi^\star( w)\|_{\bA}^2]\le \epsilon^2$ (i.e. with the parameter setting as we discussed in Remark~\ref{remark: selection-of-inner-parameter}), then
\begin{align}\label{eq:inner-L1-bound}
\EE\Big[d_\pi^\top\big|\bPhi(\xi_K( w_{k-1})-\xi^\star( w_{k-1}))\big|^2\Big]
\le
{\epsilon}^2\cdot \EE_{(s,a)\sim \bar d_\pi}\big[\|\phi(s,a)\|_{\bA^{-1}}^2\big].
\end{align}

\paragraph{Bounding the Last Term.} Similarly, by definition of $\xi^\star(w_{k-1})$, we have \begin{align*}
    &\bPhi^\top \bD \bPhi \xi^\star(w_{k-1}) = \bPhi^\top \bD \cH(\lceil \hat{\bQ}_{k-1} \rceil) = \bPhi^\top \bD \bPhi \xi' + O(\cE_{\text{approx}}) \\
    &\implies \lVert \xi^\star(w_{k-1}) - \xi' \rVert_{\bA}^2 =  (\xi^\star(w_{k-1}) - \xi')^\top\bA \bA^{-1} \bA(\xi^\star(w_{k-1}) - \xi') = O(\cE_{\text{approx}}^2/\lambda^+_{\min}(\bA))
\end{align*}
thus \begin{align*}
   \bar{d}_{\pi}^\top\lvert \bPhi(\xi' - \xi^\star(w_{k-1}) \rvert^2 \leq \EE_{\bar{d}_\pi}\big[\lVert \phi(s,a) \rVert_{\bA^{-1}}^2 \big]\cdot \underbrace{\lVert \xi' - \xi^\star(w_{k-1}) \rVert_{\bA}^2}_{= O(\cE^2_{\text{approx}})}
\end{align*}

\paragraph{Putting All Together.} Combining all above bounds together and take recursion, we have any $T\ge 1$,
\begin{equation}\label{eq:final-outer-bound}
\begin{aligned}
\EE_{(s,a)\sim d_\pi}\big[|\hat{Q}_T(s,a)-Q^\star(s,a)|^2\big]
&\leq
\gamma^T\,\EE_{(s,a)\sim d_\pi}\big[|\hat{Q}_0(s,a)-Q^\star(s,a)|^2\big]
+\\
&\frac{\EE_{(s,a)\sim d_\pi}\big[\|\phi(s,a)\|^2_{\bA^{-1}}\big]}{1-\gamma}\left(
{\epsilon}^2 
+
\cE_{\mathrm{approx}}^2
\right) + \cE_{\text{approx}}^2,
\end{aligned}
\end{equation}
as desired.

\subsection{Proof of Corollary~\ref{cor:weight_conv}}

By our Assumption~\ref{assumption-linear}, we have \begin{align*}
   &\EE_{(s,a) \sim \bar{d}_\pi}\lvert \hat{Q}_T(s,a) - Q^\pi (s,a) \rvert^2 = \sum_{s,a} \bar{d}_\pi(s,a)\lvert\langle \phi^\pi(s,a),\hat{\xi}_T-\xi^\star \rangle\rvert^2\\
   &= (\hat{\xi}_T-\xi^\star)^\top\underbrace{\big[ \sum_{s,a} \bar{d}_\pi(s,a) \phi^{\pi}(s,a)\phi^{\pi}(s,a)^\top\big]}_{ = \bA_\pi}(\hat{\xi}_T-\xi^\star)= \lVert \hat{\xi}_T - \xi^\star \rVert_{\bA_\pi}^2.
\end{align*}
Then taking expectation in both sides and applying Theorem~\ref{cor:weight_conv} leads to the desired result.

\subsection{Proof of Theorem~\ref{thm-main-convergence}}

To prove Theorem~\ref{thm-main-convergence}, we first develop the following high-probability version of the evaluation error bound, throughout this section, we use the same notation as in
Appendix~\ref{appendix:thm-eval-proof}.
\begin{lemma}
\label{lem:reuse-inner-concentration}
At each inner loop of Algorithm~\ref{alg:target-based-evaluation} with target variable $w$, it holds with probability at least $1-\delta$,
\begin{equation}
\label{eq:reuse-inner-tail}
\begin{aligned}
\|\xi_K-\xi^\star(w)\|_{\bA}^2
&\leq \frac{3e^{-2c_0\alpha K}}{(1-\gamma)^2} +\frac{256(1+\alpha K)}{(1-\gamma)^2}
\bigg[\frac{\alpha\tau_\alpha}{c_0}\log\frac{2\tau_\alpha}{\delta}
+\bigg(\frac{\alpha\tau_\alpha}{c_0}\bigg)^2\bigg].
\end{aligned}
\end{equation}
\end{lemma}

\begin{proof}
For $\bar{\bF}, \bF$ defined in~\eqref{eq-appendix: F-def}, using $\bar{\bF}(\xi)=-\bA(\xi-\xi^\star(w))$ and the inner loop update formula, we have
\[
\xi_K-\xi^\star(w)
=-(\bI-\alpha\bA)^K\xi^\star(w)
+\alpha\sum_{i=0}^{K-1}(\bI-\alpha\bA)^{K-1-i}
\big[\bF(\xi_i,\by_i)-\bar{\bF}(\xi_i)\big].
\]
Since $\lambda_{\max}(\bA)\leq 1$ and $\xi^\star(w) \in \text{Range}(\bA)$, the first term can be bounded as follows: \begin{align*}
    (1-\alpha c_0)^K \lVert \xi^\star(w)\rVert_{\bA} \leq \frac{e^{-c_0 \alpha K}}{1-\gamma}.  
\end{align*}
It remains to bound the second summation term. For this purpose, we first provide the following deterministic bounds.
For the $i$-th update of the inner loop, we have
\begin{equation}\label{eq-appendix: tmp4}
\begin{aligned}
&\|\xi_{i+1}\|_2^2-\|\xi_i\|_2^2\\
&\leq\alpha\bigg[\bigg(r(S_i,A_i)+\gamma\sum_{a'\in\cA}\pi(a'|S_{i+1})\lceil\phi(S_{i+1},a')^\top w\rceil\bigg)^2
-(\phi(S_i,A_i)^\top\xi_i)^2\bigg]\\
&\leq\alpha\bigg(1+\frac{\gamma}{1-\gamma}\bigg)^2=\frac{\alpha}{(1-\gamma)^2}.
\end{aligned}
\end{equation}
Summing from $\xi_0=0$ gives $\|\xi_i\|_2\leq\sqrt{\alpha K}/(1-\gamma)$ for $i\leq K$.
As a result,
$$\|\bF(\xi_i,y)\|_2
\leq\|\xi_i\|_2+\frac{1}{1-\gamma}
\leq\frac{1+\sqrt{\alpha K}}{1-\gamma},\quad \forall y\in \mathcal{Y}.$$
Since $\lambda_{\max}(\bA) \leq 1$, this gives \begin{align*}
   \big\lVert  \alpha\sum_{i=0}^{\tau_\alpha-1}(\bI-\alpha\bA)^{K-1-i}
\big[\bF(\xi_i,\by_i)-\bar{\bF}(\xi_i)\big]\big\rVert_{\bA} \leq 2\alpha \frac{\tau_{\alpha}(1+\sqrt{\alpha K})}{1-\gamma}.
\end{align*}

It remains to bound the terms with $i\geq\tau_\alpha$.
Let $\mathcal{F}_i$ denote the historical information before the $i$-th update.
Applying the argument
in the proof of Lemma~\ref{lem:mixing-bias} gives
\begin{align}\label{eq-appendix: tmp1}
&\big\|\EE[\bF(\xi_{i-\tau_\alpha},\by_i)\mid\mathcal F_{i-\tau_\alpha}]
-\bar{\bF}(\xi_{i-\tau_\alpha})\big\|_2 \leq 2 \alpha\sup_{y\in\cY}\|\bF(\xi_{i-\tau_\alpha},y)\|_2
\leq\frac{2\alpha(1+\sqrt{\alpha K})}{1-\gamma}.
\end{align}
Summing $\xi_{j+1}-\xi_j=\alpha\bF(\xi_j,\by_j)$ gives
\begin{equation}\label{eq-appendix: tmp2}
\|\xi_i-\xi_{i-\tau_\alpha}\|_2
\leq\alpha\sum_{j=i-\tau_\alpha}^{i-1}\|\bF(\xi_j,\by_j)\|_2
\leq\frac{\alpha\tau_\alpha(1+\sqrt{\alpha K})}{1-\gamma}.
\end{equation}
Then, for $D_i:=\bF(\xi_{i-\tau_\alpha},\by_i)
-\EE[\bF(\xi_{i-\tau_\alpha},\by_i)\mid\mathcal F_{i-\tau_\alpha}]$,
combining~(\ref{eq-appendix: tmp1}) and (\ref{eq-appendix: tmp2}) with the Lipschitz bound for $\bF$ in Appendix~\ref{appendix:thm-eval-proof} gives
\begin{align*}
\|\bF(\xi_i,\by_i)-\bar{\bF}(\xi_i)-D_i\|_2
&\leq2\|\xi_i-\xi_{i-\tau_\alpha}\|_2
+\frac{2\alpha(1+\sqrt{\alpha K})}{1-\gamma} \leq\frac{4\alpha\tau_\alpha(1+\sqrt{\alpha K})}{1-\gamma}.
\end{align*}
Using $\alpha\sum_{j=0}^{K-1}(1-\alpha c_0)^j\leq1/c_0$, we obtain
\begin{align*}
&\bigg\|\alpha\sum_{i=\tau_\alpha}^{K-1}(\bI-\alpha\bA)^{K-1-i}
\big[\bF(\xi_i,\by_i)-\bar{\bF}(\xi_i)\big]\bigg\|_{\bA} \\
&\leq\frac{4\alpha\tau_\alpha(1+\sqrt{\alpha K})}{c_0(1-\gamma)}
+\bigg\|\alpha\sum_{i=\tau_\alpha}^{K-1}(\bI-\alpha\bA)^{K-1-i}D_i\bigg\|_{\bA}.
\end{align*}
It remains to bound the last term. For this purpose, we split the sum into $\tau_\alpha+1$ subsequences, each taking every $(\tau_\alpha+1)$-st term, and apply the vector version of the Azuma--Hoeffding inequality to each subsequence with the boundedness condition
\[
\|\alpha\bA^{1/2}(\bI-\alpha\bA)^{K-1-i}D_i\|_2
\leq\frac{2\alpha(1+\sqrt{\alpha K})}{1-\gamma}(1-\alpha c_0)^{K-1-i},
\]
and a failure probability of $\delta/(\tau_\alpha+1)$. This gives, with probability at least $1-\delta$,
\begin{align*}
&\bigg\|\alpha\sum_{i=\tau_\alpha}^{K-1}(\bI-\alpha\bA)^{K-1-i}D_i\bigg\|_{\bA}\\
&\leq\frac{2\sqrt{2}\alpha(1+\sqrt{\alpha K})}{1-\gamma}
\sqrt{(\tau_\alpha+1)\log\frac{2(\tau_\alpha+1)}{\delta}
\sum_{i=\tau_\alpha}^{K-1}(1-\alpha c_0)^{2(K-1-i)}}\\
&\leq\frac{4\sqrt{2}(1+\sqrt{\alpha K})}{1-\gamma}
\sqrt{\frac{\alpha\tau_\alpha}{c_0}\log\frac{2\tau_\alpha}{\delta}}.
\end{align*}
In the last inequality, we use the geometric summation bound and
$(\tau_\alpha+1)\log(2(\tau_\alpha+1)/\delta)\leq4\tau_\alpha\log(2\tau_\alpha/\delta)$.
Combining these bounds with the bound for the first $\tau_\alpha$ terms gives~\eqref{eq:reuse-inner-tail}.
\end{proof}

\begin{corollary}
\label{cor:reuse-critic-tail}
With parameters specified in
Remark~\ref{remark: selection-of-inner-parameter}.
For every $u \geq 0$, the output of Algorithm~\ref{alg:target-based-evaluation} satisfies with probability at least $1-e^{-u}$,
\begin{equation}
\label{eq:reuse-critic-tail}
\|\hat\xi_T-\xi^\star\|_{\mathbf A_\pi}^2
\leq\frac{3}{(1-\gamma)^2}
\big[\gamma^{2T}+(1+\sqrt{\cC(\pi,\mu)})^2\mathcal E_{\rm approx}^2
+\cC(\pi,\mu)\epsilon^2(1+u)\big].
\end{equation}
\end{corollary}

\begin{proof}
Bringing the parameter choices in
\eqref{eq:alpha-choice} and \eqref{eq:K-choice} to Lemma~\ref{lem:reuse-inner-concentration}, we have the initial term in
\eqref{eq:reuse-inner-tail} at most $\epsilon^2/2$, and its remaining
term at $\delta=1/(eT)$ at most $\epsilon^2/2$.
For $u>0$, applying Lemma~\ref{lem:reuse-inner-concentration} with $\delta=e^{-u}/T$ and taking union bound over $t\in [T]$ give, with probability
at least $1-e^{-u}$,
\[
\|\xi_K(w_j)-\xi^\star(w_j)\|_{\bA}^2
\leq\epsilon^2(1+u), \quad \forall 0\leq j \leq T
\]
Under such event, \eqref{eq:reuse-critic-tail} follows by repeating the argument in Appendix~\ref{appendix:thm-eval-proof} and applying the identity in the proof
of Corollary~\ref{cor:weight_conv}.
\end{proof}

\begin{proof}[Proof of Theorem~\ref{thm-main-convergence}]
With $0<\eta\leq1/L$ and $e_k$ and $\delta_{k,t}$ defined in
Appendix~\ref{appendix:dual-trust-coverage}, applying Assumption~\ref{assumption:smoothness} leads to
\[
J_\rho(\pi_{t+1}^{(k)})-J_\rho(\pi_t^{(k)})
\geq\frac{\eta}{2}\|\nabla J_\rho(\pi_t^{(k)})\|_2^2
-\frac{\eta}{2}\|\delta_{k,t}\|_2^2.
\]
On the other hand, by Assumption~\ref{assumption:anchor-gradient} and the trust region design in 
\eqref{eq:appendix-c1}--\eqref{eq:appendix-c2}, it holds that
$\|\delta_{k,t}\|_2^2\leq c_{\mathrm g}\|e_k\|_{\mathbf A_k}^2$.
Summing over the $M$ actor updates yields
\begin{align*}
\frac1M\sum_k\sum_{t=0}^{T_k-1}\|\nabla J_\rho(\pi_t^{(k)})\|_2^2
&\leq\frac{2(J_\rho^\star-J_\rho(\pi^{(0)}))}{\eta M}
+\frac1M\sum_k\sum_{t=0}^{T_k-1}\|\delta_{k,t}\|_2^2\\
&\leq\frac{2(J_\rho^\star-J_\rho(\pi^{(0)}))}{\eta M}
+c_{\mathrm g}\max_{k:T_k>0}\|e_k\|_{\mathbf A_k}^2
\end{align*}
by $\sum_{k}T_k = M.$ 

It remains to control $\max_{k:T_k>0}\|e_k\|_{\mathbf A_k}^2.$ For this purpose, applying Corollary~\ref{cor:reuse-critic-tail} with failure probability $e^{-u}/M$ for each $k$ and taking union
bound leads to, with probability
at least $1-e^{-u}$,
\[
\max_{k:T_k>0}\|e_k\|_{\mathbf A_k}^2
\leq\frac{12\max\{C_{\max},d\}}{(1-\gamma)^2}
\big[\mathcal E_{\rm approx}^2+\epsilon^2(1+\log M+u)\big].
\]
Taking integration with respect to $u$ then gives
\begin{equation}
\label{eq:reuse-maximum-mean}
\EE\bigg[\max_{k:T_k>0}\|e_k\|_{\mathbf A_k}^2\bigg]
\leq\frac{12\max\{C_{\max},d\}}{(1-\gamma)^2}
\big[\mathcal E_{\rm approx}^2+\epsilon^2(2+\log M)\big],
\end{equation}
as desired.
\end{proof}

\section{More Details and Proofs for Section~\ref{sec:replay-budget}}
\label{app:replay-budget}

\subsection{More explanation in diversity-coverage trade-off}\label{appendix:detailed-tradeoff}
In this section, we provide more details of the diversity-coverage trade-off explanation of Lemma~\ref{prop:replay-evaluation}.

\paragraph{The least-squares term is minimized under uniform design.} We show that the weighted least-squared error term $\mathcal V_{\mathcal B}(\pi,\mathbf w)$ is minimized under uniform sampling, $w_i=1/n$. To see this, for the uniform weight vector $\mathbf{u}$, noticing the identity
\begin{equation*}
 \mathcal{V}_{\mathcal{B}} (\pi, \mathbf{w})- \mathcal{V}_{\mathcal{B}} (\pi, \mathbf{w}_0) = \sum_{i=1}^n\lVert (w_i \bm{\Sigma}_{\mathbf{w}}^{-1} -\frac{1}{n}\bm{\Sigma}_{\mathbf{u}}^{-1})\phi^{\pi}(s_i,a_i)  \rVert_{\bm{A}_\pi}^2 \geq 0
\end{equation*}
holds for all $\mathbf{w}\in \Delta^n$. This gives the desired claim. In particular, such minimizer is independent of the policy $\pi$ to be evaluated, and just encourage the resulted distribution as diverse as possible over the buffer set.
\paragraph{The TD term is minimized under optimal $L$-design.}
Finding the weight minimizing the $\mathcal{C}_{\mathcal{B}}(\pi, \mathbf{w})$ term leads to the standard $L$-optimal design problem \citep{atkinson2007optimum}, which is convex and can be solved by Frank-Wolfe type algorithms. In contrast to the LS term, the TD term encourages the replay distribution to place more weight on samples that improve coverage of feature directions emphasized by the target-policy covariance $\mathbf A_\pi$.

\begin{figure}[tb]
    \centering
    \includegraphics[width=0.6\linewidth]{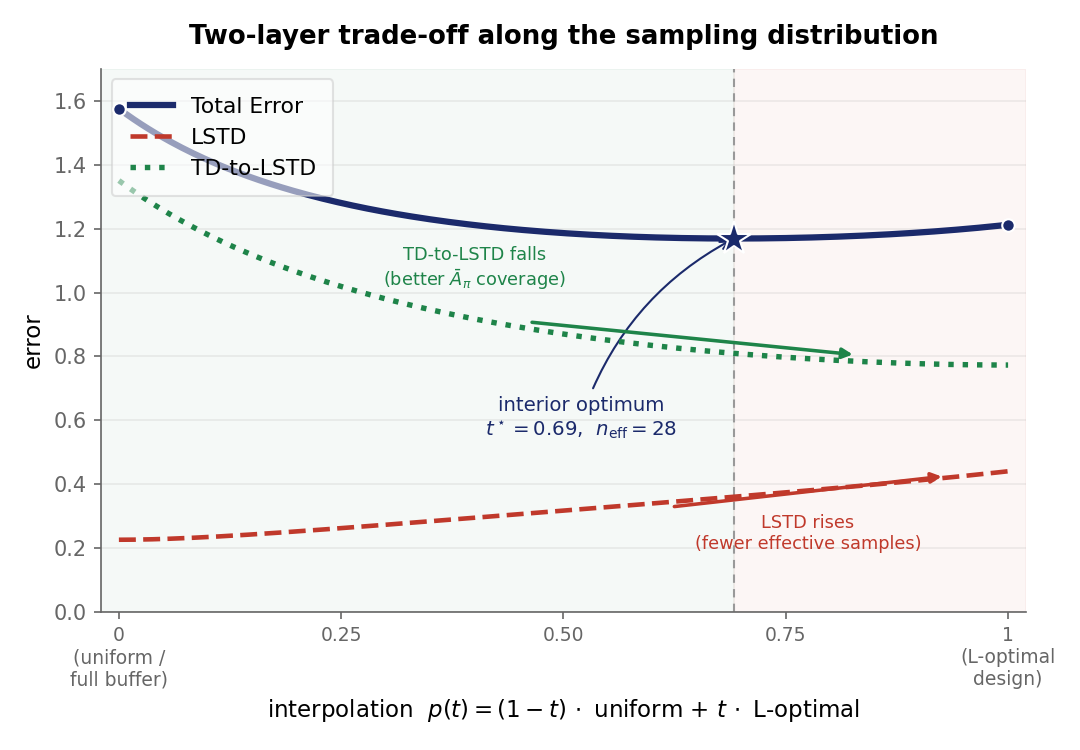}
    \caption{Illustration of the diversity-coverage trade-off in buffer sampling.}
    \label{fig:buffer-trade-off}
\end{figure}

Figure~\ref{fig:buffer-trade-off} illustrates the trade-off between the two terms in a toy environment. We linearly interpolate between the uniform design and the $L$-optimal design and plot how the objective in~\eqref{eq:replay-budget-program} varies along this interpolation.

\subsection{Fixed-Budget Replay Design}
\label{appendix:fixed-replay-budget}

The formulation in~\eqref{eq:replay-budget-program} jointly selects the
replay distribution and the inner-loop TD budget to achieve a prescribed
evaluation accuracy.
A complementary view is to fix the replay budget $K$ and optimize the
sampling distribution to minimize the resulting policy-evaluation bound.
Since the target-network residual is independent of $\mathbf w$, the
fixed-budget replay design reduces to
\begin{equation}
\label{eq:replay-fixed-budget}
\hat{\mathbf w}_K
\in
\arg\min_{\mathbf w\in\Delta^n}
\frac{c_{\mathrm{stat}}
\mathcal V_{\mathcal B}(\pi,\mathbf w)}{(1-\gamma)^3}
+
\frac{c_{\mathrm{TD}}
\mathcal C_{\mathcal B}(\pi,\mathbf w)}{K(1-\gamma)^4}.
\end{equation}

Equation~\eqref{eq:replay-fixed-budget} makes the diversity--coverage
trade-off particularly transparent.
For small $K$, the optimization term receives relatively greater weight,
favoring replay distributions with stronger target-policy coverage and hence
smaller $\mathcal C_{\mathcal B}$.
As $K$ increases, the TD-to-LS optimization gap becomes less important and
the statistical term increasingly dominates; in the large-budget limit, the
design therefore approaches the uniform distribution that minimizes
$\mathcal V_{\mathcal B}$.
Thus, the replay distribution naturally shifts from coverage-oriented
sampling toward diversity-oriented sampling as the available evaluation
budget increases.

The objective in~\eqref{eq:replay-fixed-budget} is generally non-convex in
$\mathbf w$. In practice, we optimize it approximately using a small number
of first-order updates.

\subsection{Proof of Lemma~\ref{prop:replay-evaluation}}\label{appendix:replay-proof}

\begin{proof}
Throughout this proof, we use $c_0$ to denote an absolute constant that may vary between inequalities, and we adopt the notations used in the proof of Theorem~\ref{thm-evaluation} throughout this proof: for a target variable $u\in\mathbb R^d$, we denote $y_i(u):=r_i+\gamma\sum_a\pi(a\lvert s_i')\lceil\phi^\pi(s_i',a)^\top u\rceil$ as the corresponding value estimation, under which the weighted least-squares solution is
\begin{equation*}
\widehat\xi^\star_{\mathbf w}(u)=\bm\Sigma_{\mathbf w}^{-1}\sum\nolimits_{i=1}^n w_i\phi^\pi(s_i,a_i)y_i(u).
\end{equation*}
As shown in Appendix~\ref{appendix:thm-eval-proof}, we have $\mathbb E[\widehat\xi^\star_{\mathbf w}(u)]=\xi^\star(u)$, $\xi^\star(\xi^\star)=\xi^\star$, and
\begin{equation}
\|\xi^\star(u)-\xi^\star(v)\|_{\bm{A}_\pi}\leq\gamma\|u-v\|_{\bm{A}_\pi} ,\quad \|\xi^\star\|_{\bm{A}_\pi}\leq\min\{L,(1-\gamma)^{-1}\}.
\label{eq:replay-value-bound}
\end{equation}

\paragraph{Bounding the least squared error.}
We define the residual matrix $\bm{R}_{\mathbf w}\in \RR^{d\times \lvert \mathcal{S}\rvert} $ as
\begin{equation*}
 \big[\bm{R}_{\mathbf w}\big]_{j,s}:= \bigg(\bm\Sigma_{\mathbf w}^{-1}\big[\sum\nolimits_{i=1}^nw_i\phi^\pi(s_i,a_i)\bm{1}\{s'_i = s\} -\sum_i w_i\phi^\pi(s_i,a_i)p(s\lvert s_i,a_i)\big]\bigg)_j .
\end{equation*}
and $\epsilon_{\mathcal B}=\gamma\big\|\bm{A}_\pi^{1/2}\bm{R}_{\mathbf w}\mathrm{diag}(d^\pi)^{-1/2}\big\|_{\mathrm{op}}$, then by 
Cauchy--Schwarz and Jensen's inequality, 
\begin{equation*}
\big\|\widehat\xi^\star_{\mathbf w}(u)-\xi^\star(u)\big\|_{\bm{A}_\pi}\leq\epsilon_{\mathcal B}\|u\|_{\bm{A}_\pi}.
\end{equation*}
Combining this bound with~\eqref{eq:replay-value-bound} gives $\|\widehat\xi^\star_{\mathbf w}(\xi^\star)-\xi^\star\|_{\bm{A}_\pi}\leq L\epsilon_{\mathcal B}$. Now it remains to control $\epsilon_{\mathcal{B}}$, for which we have the following upper bound results:
\begin{equation}
\mathbb E[\epsilon_{\mathcal B}^2]\leq\frac{\mathcal V_{\mathcal B}(\pi,\mathbf w)}{d_{\min}^\pi},
\quad \mathbb E[\epsilon_{\mathcal B}^4]\leq\frac{c_0\mathcal V_{\mathcal B}(\pi,\mathbf w)^2}{(d_{\min}^\pi)^2}.
\label{eq:replay-buffer-moments}
\end{equation}
To see the first inequality of~\eqref{eq:replay-buffer-moments}, denote \begin{align*}
    \big[\bm{Z}_i\big]_{j,s}:= \frac{\gamma w_i}{\sqrt{d^\pi(s)}} \bigg(\bm{A}_\pi^{1/2}\bm\Sigma_{\mathbf w}^{-1}\big[\phi^\pi(s_i,a_i)\bm{1}\{s'_i = s\} -\phi^\pi(s_i,a_i)p(s\lvert s_i,a_i)\big]\bigg)_j,
\end{align*}
then it holds that $\epsilon_{\mathcal{B}} \leq \big\lVert \sum_{i=1}^n \bm{Z}_i \big\rVert_F$ and by 
\begin{align}\label{eq-appendix: Z-bound}
    \EE\big[\lVert \bm{Z}_i \rVert_F^2 \big] = \gamma^2 w_i^2 \lVert \bm{\Sigma}_{\mathbf w}^{-1} \phi^\pi(s_i,a_i)\rVert_{\bm{A}_\pi}^2 \sum_{s} \frac{p(s\lvert s_i,a_i)(1-p(s\lvert s_i,a_i)) }{d^\pi(s)},
\end{align}
it holds that \begin{align*}
    \EE[\epsilon_{\mathcal{B}}^2] \leq \frac{\gamma^2}{d^\pi_{\min}}\sum_{i=1}^n w_i^2 \lVert \Sigma_{\mathbf w}^{-1}\phi^\pi(s_i,a_i)\rVert_{\bm{A}_\pi}^2=\frac{\gamma^2 \mathcal{V}_{\mathcal{B}}(\pi, \mathbf{w})}{d_{\min}^\pi}, 
\end{align*}
as desired.

To see the second inequality of \eqref{eq:replay-buffer-moments}, noticing that by~\eqref{eq-appendix: Z-bound}, \begin{align*}
    &\EE\big[\epsilon_{\mathcal{B}}^4 \big] \leq  \EE\big[\lVert  \sum_{i=1}^n \bm{Z}_i \rVert_F^4  \big]  = \sum_{i=1}^n \EE\big[\lVert \bm{Z}_i \rVert_F^4  \big] + 2\sum_{i<j} \EE\big[\lVert \bm{Z}_i\rVert_F^2 \lVert \bm{Z}_j\rVert_F^2  \big]  + 4\sum_{i<j} \EE\big[\langle \bm{Z}_i, \bm{Z}_j \rangle^2  \big]\\
    &\leq 3 \bigg(\sum_{i=1}^n \frac{4\gamma^2w_i^2}{d_{\min}^\pi}\lVert\bm\Sigma_{\mathbf w}^{-1}\phi^\pi(s_i,a_i)\rVert_{\bm A_\pi}^2\bigg)^2\leq c_0 \frac{ \gamma^4 \mathcal{V}^2_{\mathcal{B}}(\pi, \mathbf{w})}{(d_{\min}^\pi)^2}.
\end{align*}

\paragraph{Bounding the TD error.}
For each outer loop step $t$ and target variable $u$, consider the $i$-th step of the inner TD update, since $|y_i(u)|\leq(1-\gamma)^{-1}$ and $\|\widehat\xi^\star_{\mathbf w}(u)\|_{\bm A_\pi}\leq(1+\epsilon_{\mathcal B})/(1-\gamma)$,
\begin{equation*}
\|\widehat\xi^\star_{\mathbf w}(u)\|_2\leq\frac{1+\epsilon_{\mathcal B}}{(1-\gamma)\sqrt{\lambda_{\min}(\bm A_\pi)}},\quad
|y_i(u)-\phi^\pi(s_i,a_i)^\top\widehat\xi^\star_{\mathbf w}(u)|\leq\frac{2(1+\epsilon_{\mathcal B})}{(1-\gamma)\sqrt{\lambda_{\min}(\bm A_\pi)}}.
\end{equation*}
For $0<\alpha\leq1/2$, write $\mathbf H_i=\bI-\alpha\phi^\pi(s_i,a_i)\phi^\pi(s_i,a_i)^\top$. By the definition of $\widehat\xi^\star_{\mathbf w}(u)$,
\begin{equation*}
\sum_{i=1}^n w_i\phi^\pi(s_i,a_i)\big[y_i(u)-\phi^\pi(s_i,a_i)^\top\widehat\xi^\star_{\mathbf w}(u)\big]=0.
\end{equation*}
Thus
$\mathbb E[\xi_{t,k}-\widehat\xi^\star_{\mathbf w}(u)]
=-(\bI-\alpha\bm\Sigma_{\mathbf w})^k\widehat\xi^\star_{\mathbf w}(u)$
and $\|\mathbb E[\xi_{t,k}-\widehat\xi^\star_{\mathbf w}(u)]\|_2\leq\|\widehat\xi^\star_{\mathbf w}(u)\|_2$. Expanding the TD update then gives
\begin{align*}
&\mathbb E\big[(\xi_{t,k+1}-\widehat\xi^\star_{\mathbf w}(u))(\xi_{t,k+1}-\widehat\xi^\star_{\mathbf w}(u))^\top\big]\\
&\preceq\sum_{i=1}^n w_i\mathbf H_i
\mathbb E\big[(\xi_{t,k}-\widehat\xi^\star_{\mathbf w}(u))(\xi_{t,k}-\widehat\xi^\star_{\mathbf w}(u))^\top\big]\mathbf H_i
+\frac{c_0\alpha^2(1+\epsilon_{\mathcal B})^2}{\lambda_{\min}(\bm A_\pi)(1-\gamma)^2}\bm\Sigma_{\mathbf w}.
\end{align*}
Using $\sum_iw_i\mathbf H_i^2\preceq\bI-\alpha(2-\alpha)\bm\Sigma_{\mathbf w}$ and $\mathrm{tr}(\bm A_\pi)\leq1$ yield
\begin{equation}
\mathbb E[\|\xi_{t,K}-\widehat\xi^\star_{\mathbf w}(u)\|_{\bm A_\pi}^2]
\leq\frac{(1+\epsilon_{\mathcal B})^2}{\lambda_{\min}(\bm A_\pi)(1-\gamma)^2}
\left[e^{-\alpha\lambda_{\min}(\bm\Sigma_{\mathbf w})K}+\frac{c_0\alpha}{2-\alpha}\right].
\label{eq:replay-inner-bounds}
\end{equation}

\paragraph{Putting all together.}
Now we are ready to combine~\eqref{eq:replay-buffer-moments} and~\eqref{eq:replay-inner-bounds} to prove~\eqref{eq:replay-evaluation-guarantee}. For a sufficiently large absolute constant $c_1$, we set
\begin{equation*}
c_{\mathrm{stat}}=\frac{c_1(1+L)^2}{\lambda_{\min}(\bm{A}_\pi)^2d_{\min}^\pi},\quad 
c_{\mathrm{TD}}=\frac{c_1(1+L)^2}{\lambda_{\min}(\bm{A}_\pi)^3d_{\min}^\pi}.
\end{equation*}
First, if the first two terms of~\eqref{eq:replay-evaluation-guarantee} sum to at least $(1+L)^2$, applying the deterministic bound in~\eqref{eq-appendix: tmp4} with the selection $\alpha = (1-\gamma)/{K}$ gives the desired result.

Otherwise, we have
\begin{equation}
\mathcal V_{\mathcal B}(\pi,\mathbf w)<\frac{(1+L)^2(1-\gamma)^3}{c_{\mathrm{stat}}},\quad \mathcal C_{\mathcal B}(\pi,\mathbf w)\geq\frac{\lambda_{\min}(\bm{A}_\pi)}{\lambda_{\min}(\bm\Sigma_{\mathbf w})}.
\label{eq:replay-design-lower-bounds}
\end{equation}
In this case, we set $\alpha=\min\{1/2,\log(2K)/[\lambda_{\min}(\bm\Sigma_{\mathbf w})K]\}$, underwhich
\begin{equation}\label{appendix-eq: tmp5}
e^{-\alpha\lambda_{\min}(\bm\Sigma_{\mathbf w})K}+\alpha\leq\frac{3\log(2K)}{\lambda_{\min}(\bm\Sigma_{\mathbf w})K}.
\end{equation}
Under the event $\mathcal E:=\{\epsilon_{\mathcal B}\leq(1-\gamma)/4\}$, using $(1+\epsilon_{\mathcal B})^2\leq2$ in~\eqref{eq:replay-inner-bounds} gives,
\begin{equation*}
\EE \big[\|\xi_{t,K}-\widehat\xi^\star_{\mathbf w}(u)\|_{\bm A_\pi}^2\big]\leq\frac{c_0\log(2K)\mathcal C_{\mathcal B}(\pi,\mathbf w)}{\lambda_{\min}(\bm{A}_\pi)^2K(1-\gamma)^2}.
\end{equation*}
Combining~\eqref{appendix-eq: tmp5} with~\eqref{eq:replay-value-bound} and $u= \hat{\xi}_t$ gives
\begin{equation*}
\left(\mathbb E[\|\hat\xi_{t+1}-\xi^\star\|_{\bm A_\pi}^2]\right)^{1/2}\leq(\gamma+\epsilon_{\mathcal B})\left(\mathbb E[\|\hat\xi_t-\xi^\star\|_{\bm A_\pi}^2]\right)^{1/2}+L\epsilon_{\mathcal B}+\sqrt{\frac{c_0\log(2K)\mathcal C_{\mathcal B}(\pi,\mathbf w)}{\lambda_{\min}(\bm A_\pi)^2K(1-\gamma)^2}}.
\end{equation*}
Since $1-\gamma-\epsilon_{\mathcal B}\geq3(1-\gamma)/4$ under $\mathcal{E}$ and
$(\gamma+\epsilon_{\mathcal B})^T-\gamma^T=\epsilon_{\mathcal B}\sum\nolimits_{j=0}^{T-1}(\gamma+\epsilon_{\mathcal B})^{T-1-j}\gamma^j\leq\frac{\epsilon_{\mathcal B}}{1-\gamma-\epsilon_{\mathcal B}}$,
summing from $\hat\xi_0=0$ and using~\eqref{eq:replay-value-bound} yields
\begin{equation*}
\left(\mathbb E[\|\hat\xi_T-\xi^\star\|_{\bm A_\pi}^2]\right)^{1/2}\leq\frac{\gamma^T}{1-\gamma}+\frac{c_0}{1-\gamma}\left[L\epsilon_{\mathcal B}+\sqrt{\frac{c_0\log(2K)\mathcal C_{\mathcal B}(\pi,\mathbf w)}{\lambda_{\min}(\bm A_\pi)^2K(1-\gamma)^2}}\right].
\end{equation*}
Taking square in both sides and applying~\eqref{eq:replay-buffer-moments}, we obtain
\begin{equation}\label{appendix-eq: E-bound}
\mathbb E[\|\hat\xi_T-\xi^\star\|_{\bm A_\pi}^2\bm{1}_{\mathcal E}]\leq\frac{3\gamma^{2T}}{(1-\gamma)^2}+\frac{c_0L^2\mathcal V_{\mathcal B}(\pi,\mathbf w)}{d_{\min}^\pi(1-\gamma)^2}+\frac{c_0\log(2K)\mathcal C_{\mathcal B}(\pi,\mathbf w)}{\lambda_{\min}(\bm A_\pi)^2K(1-\gamma)^4}.
\end{equation}

On the other hand, under $\mathcal E^c$, using~\eqref{eq:replay-inner-bounds},~\eqref{eq:replay-value-bound} and the fact $\|\widehat\xi^\star_{\mathbf w}(u)\|_{\bm A_\pi}\leq(1+\epsilon_{\mathcal B})/(1-\gamma)$gives
\begin{equation}
\mathbb E[\|\hat\xi_T-\xi^\star\|_{\bm{A}_\pi}^2]\leq\frac{c_0(1+\epsilon_{\mathcal B})^2}{\lambda_{\min}(\bm{A}_\pi)(1-\gamma)^2}.
\label{eq:replay-output-bound}
\end{equation}
By $(1+\epsilon_{\mathcal B})^2\leq c_0\epsilon_{\mathcal B}^4/(1-\gamma)^4$ under $\mathcal E^c$ and~\eqref{eq:replay-buffer-moments}, we obtain
\begin{equation}\label{appendix-eq: Ec-bound}
\mathbb E[\|\hat\xi_T-\xi^\star\|_{\bm A_\pi}^2\bm{1}_{\mathcal E^c}]\leq\frac{c_0\mathbb E[(1+\epsilon_{\mathcal B})^2\bm{1}_{\mathcal E^c}]}{\lambda_{\min}(\bm A_\pi)(1-\gamma)^2}\leq\frac{c_0\mathcal V_{\mathcal B}(\pi,\mathbf w)^2}{\lambda_{\min}(\bm A_\pi)(d_{\min}^\pi)^2(1-\gamma)^6}.
\end{equation}
Combining~\eqref{appendix-eq: E-bound} and~\eqref{appendix-eq: Ec-bound} together, we have
\begin{align*}
\mathbb E[\|\hat\xi_T-\xi^\star\|_{\bm{A}_\pi}^2]
&\leq\frac{3\gamma^{2T}}{(1-\gamma)^2}+\frac{c_0L^2\mathcal V_{\mathcal B}(\pi,\mathbf w)}{d_{\min}^\pi(1-\gamma)^2} +\frac{c_0\log(2K)\mathcal C_{\mathcal B}(\pi,\mathbf w)}{\lambda_{\min}(\bm A_\pi)^2K(1-\gamma)^4}\\
&+\frac{c_0\mathcal V_{\mathcal B}(\pi,\mathbf w)^2}{\lambda_{\min}(\bm{A}_\pi)(d_{\min}^\pi)^2(1-\gamma)^6}.
\end{align*}
Finally, by~\eqref{eq:replay-design-lower-bounds},
\begin{equation*}
\frac{\mathcal V_{\mathcal B}(\pi,\mathbf w)^2}{\lambda_{\min}(\bm{A}_\pi)(d_{\min}^\pi)^2(1-\gamma)^6}\leq\frac{\lambda_{\min}(\bm A_\pi)\mathcal V_{\mathcal B}(\pi,\mathbf w)}{c_1d_{\min}^\pi(1-\gamma)^3}.
\end{equation*}
Substituting this bound to the above inequality gives the desired result.
\end{proof}